\documentclass[preprint,12pt]{elsarticle}

\usepackage{lineno}

\usepackage{graphicx}
\usepackage[utf8]{inputenc}
\usepackage{lmodern}
\usepackage[T1]{fontenc}
\usepackage{lscape}
\usepackage{amsmath}	
\usepackage{amssymb}
\usepackage{cases}
\usepackage{graphicx}
\usepackage{caption}
\usepackage{comment}

\usepackage{makecell}
\usepackage{multirow}
\usepackage{amsfonts}
\usepackage{float} 
\usepackage{diagbox}
\usepackage{fullpage}
\usepackage{url}
\usepackage{rotating}
\usepackage{eurosym}
\usepackage{wrapfig}
\usepackage[final]{pdfpages} 
\usepackage{epstopdf} 
\usepackage[a4paper]{geometry}
\usepackage[nottoc]{tocbibind}
\usepackage{placeins}
\usepackage{float}
\usepackage{listings}
\usepackage{color, colortbl}
\usepackage{hyperref}
\usepackage{xcolor}
\usepackage{graphicx, caption, subcaption}
\usepackage[nohyperlinks]{acronym}
\usepackage{sidecap}
\usepackage{algorithm,algpseudocode}
\usepackage{multirow}
\usepackage{booktabs}
\hypersetup{
colorlinks,
citecolor=black,
filecolor=black,
linkcolor=black,
urlcolor=black
} 

\begin{document}

\begin{frontmatter}

\title{Variational Parameter Calibration with Physics-Aware Latent-Space Surrogates}

\author{Qiyao Zhou$^{1}$, Xujia Zhu$^{2}$, Pierre Joli$^{1}$, Yu Cong$^{1}$, Sibo Cheng$^{3,*}$}

\address{\small $^{1}$ LMEE, UnivEvry, Université Paris-Saclay, Evry, France\\
         \small $^{2}$ L2S, Université Paris-Saclay, CNRS, CentraleSupéléc, Gif-sur-Yvette, France\\
         \small $^{3}$ CEREA, ENPC, EDF R\&D, Institut Polytechnique de Paris, Île-de-France, France\\
         \texorpdfstring{$^{*}$}{*} corresponding: sibo.cheng@enpc.fr \\

}

\begin{abstract} 
Forward and inverse modeling of parametric dynamical systems requires surrogate models that are not only accurate for state prediction, but also informative for parameter calibration. However, a systematic end-to-end differentiable formulation for coupling deep-learning-based reduced-order surrogates with variational parameter estimation remains underdeveloped. In this work, we introduce a physics-aware neural-network-based latent-space framework for reduced-order forward modeling and variational parameter estimation. The proposed autoencoder-based approach yields a differentiable surrogate that maps physical parameters to predicted flow fields through a latent representation. The observable supervision is used during offline training to encourage the latent variables to retain information correlated with system parameters, while the online inverse problem is solved in the parameter space through the surrogate-induced observation operator. The method is evaluated on two computational-fluid-dynamics benchmarks. The results show that reconstruction accuracy alone is insufficient for inverse modeling, owing to the lack of end-to-end differentiability or physics awareness for variational parameter calibration. Quantitative latent-space analysis further shows that observable supervision improves case-level separability and temporal organization of latent representations. Experiments with realistic measurement settings, including noisy, low-resolution, randomly masked, and block-wise partial observations, demonstrate the robustness of the proposed framework and show that it generally reduces calibration error and variability compared with the standard surrogate models. 
\end{abstract}

\end{frontmatter}

\newpage
\section*{Main Notations}
\begin{table*}[ht!]
    \centering
    \begin{tabular}{ p{3.5cm} p{15cm}}
$\mathcal{X}$ & Snapshot matrix\\
$\mathbf{x}_t$ & Flattened state vector at time $t$ \\
$\mathbf{C}_{\mathcal{X}}$ & Empirical covariance matrix of snapshots \\
$\Psi_j$ & $j$-th POD basis\\
$q$ & Dimension of latent space \\
$\mathbf{L}_{\mathcal{X},q}$ & POD projection matrix with $q$ retained modes \\
$\hat{\mathbf{x}}_t$ & POD coefficient vector at time $t$\\
$k_{j,t}$ & $j$-th POD coefficient at time $t$ \\
$\boldsymbol{\theta}$ & Parameter vector\\
$\Omega_{\boldsymbol{\theta}}$ & Parameter space\\
$k_j(\boldsymbol{\theta})$ & $j$-th POD coefficient with parameter $\boldsymbol{\theta}$ \\
$\mathbf{z}$ & Latent vector in CAE \\
$\mathbf{z}^r$ & Latent vector in OACAE \\
$\mathcal{F}_e, \mathcal{F}_d$ & Encoder and decoder of CAE \\
$\mathcal{F}_e^r, \mathcal{F}_d^r$ & Encoder and decoder of OACAE \\
$\mathcal{E}_a$ & Observable regression MLP branch \\
$\mathcal{E}_b$ & Parameter-to-latent regressor MLP in OACAE-MLP \\
$\mathcal{E}_c$ & Parameter-to-latent regressor MLP in CAE-MLP \\
$\phi_e, \phi_d, \phi_a, \phi_b, \phi_c$ & Trainable weights of encoders/decoders/regressors \\
$\mathcal{L}_1, \mathcal{L}_2$ & CAE-MLP training loss functions (reconstruction/latent regression) \\
$\mathcal{L}_1^r, \mathcal{L}_2^r$ & OACAE-MLP training loss functions (reconstruction/latent regression) \\
$\alpha, \beta$ & Weighting factors balancing loss terms \\
$N_e$ & Ensemble size in EnKF \\
$\boldsymbol{\theta}^b, \boldsymbol{\theta}^a$ & Background and analysis parameter vectors \\
$\mathbf{B}$ & Background error covariance matrix \\
$\mathbf{R}$ & Observation error covariance matrix \\
$\mathbf{K}^*$ & Kalman gain in EnKF \\
$\mathcal{H}_{\text{POD-GPR}}$ & Observation operator with POD-GPR surrogate \\
$\mathcal{H}_{\text{CAE-MLP}}$ & Observation operator with CAE-MLP surrogate \\
$J_{\textbf{3D-Var}}, J_{\textbf{4D-Var}}$ & 3D-Var/4D-Var cost functions \\
    \end{tabular}
\end{table*}
\clearpage

\section{Introduction}
\subsection{Parametric systems}

Many physical and engineering problems can be described by Partial Differential Equations (PDEs) whose solutions depend on a set of parameters, such as material properties, boundary or initial conditions, and geometrical configurations. Such parametric systems provide a flexible and realistic representation of physical processes, allowing researchers and engineers to explore system behavior across a wide range of configurations. At the same time, the resulting solution manifolds are often high-dimensional and may exhibit strong nonlinear variability with respect to the parameters, making their analysis computationally demanding.

A solution field of physical quantities, such as temperature, displacement, velocity, or pressure, is typically expressed by 
\[
\mathbf{x} : \mathcal{T} \times \Omega \times \mathcal{P} \;\to\; \mathbb{R}^m ,
\]
where they depend on time $t \in \mathcal{T}\subset \mathbb{R}$, spatial coordinates $s \in \Omega \subset \mathbb{R}^d$, and system parameters $ \mathbf{p} \in \mathcal{P} \subset \mathbb{R}^{d_{\mathbf{p}}}$. The latter may encode initial or boundary conditions, material properties, or other system characteristics, and the solution field is constrained by the governing PDE.

The study of parametric systems is motivated by a wide range of applications. In materials engineering, parametric descriptions of microstructure or elastic coefficients are essential for predicting macroscopic responses under load \cite{Najian2025}.  In wildfire propagation models, wind conditions and vegetation parameters strongly affect the fire front dynamics \cite{Burela2025}. In fluid dynamics, different Reynolds numbers or inflow profiles generate distinct vortex shedding and turbulence structures \cite{Fresca2021, Vlachas2025}. These examples illustrate the complexity and sensitivity of parametric PDE solutions, motivating the development of efficient modeling and analysis tools.

Among parametric PDE systems, Computational Fluid Dynamics (CFD) represents one of the most prominent and challenging application areas, with a wide range of applications in engineering and geophysical sciences. The governing equations of fluid dynamics are the Navier-Stokes equations, which express conservation of mass and momentum. For a Newtonian fluid with time-independent density, and neglecting body forces, a simplified form of these equations can be written as
\begin{equation}\label{eq:NS}
\begin{aligned}
\nabla \cdot (\rho \mathbf{u}) & = 0, \\
\frac{\partial}{\partial t}(\rho \mathbf{u}) + \nabla \cdot (\rho \mathbf{u} \mathbf{u}) 
& = - \nabla p + \nabla \cdot \mu\big[\nabla \mathbf{u} + (\nabla \mathbf{u})^\top \big],
\end{aligned}
\end{equation}
where $\rho$ denotes the fluid density, $\mathbf{u}=(u,v)^\top$ the velocity field, $p$ the pressure, and $\mu$ the viscosity. Under incompressibility and Newtonian assumptions, these equations form the classical foundation of fluid mechanics \cite{Batchelor1967, LandauLifshitz1987}. Boundary conditions, material properties, and geometrical configurations define a high-dimensional parameter space that strongly shapes the resulting flow dynamics, leading to transitions between laminar and turbulent regimes, vortex shedding, and highly unsteady behavior. The numerical solution of the Navier-Stokes equations relies on high-fidelity discretization techniques such as finite difference, finite volume, finite element, or spectral methods. While these approaches can deliver accurate solutions, they are computationally expensive, particularly for complex geometries, long-time integrations, or high-dimensional parameter studies. 

\medskip
\subsection{Reduced order modeling} 
\label{Sec:ROM}
Compared with full-order modeling approaches \cite{Fresca2021} for high-fidelity computational fluid dynamics simulations, reduced-order modeling aims to reduce the computational cost of repeated simulations while preserving the dominant structures of the underlying dynamics. Most reduced-order models are constructed from a set of high-fidelity solutions, usually referred to as snapshots, and seek a low-dimensional representation of the solution manifold.

Projection-based methods constitute a classical class of reduced-order modeling techniques. Proper orthogonal decomposition (POD)\cite{chuqiao2025} and dynamic mode decomposition (DMD)\cite{Jonathan2013,liu2025hybrid} are representative examples that construct linear reduced spaces from snapshot data. These methods are efficient, interpretable, and often effective when the dominant flow variability can be captured by a limited number of linear modes. However, their reduced coordinates are mainly optimized according to reconstruction criteria. As a result, they are not necessarily organized with respect to the physical parameters that would typically be inferred in an inverse problem. To enable non-intrusive prediction for new parameter values, projection-based models are often combined with regression techniques, e.g., Gaussian process regression (GPR) and kernel ridge regression (KRR) \cite{weiji2025}, to learn mappings from the parameter space to reduced coordinates. This would result in hybrid reduced-order surrogates. Typicaly examples include POD-GPR \cite{lumet2025POD-GPR} and POD-LSTM \cite{Arvind2018}. These approaches can provide efficient parameter-to-state prediction without repeatedly solving the full-order model. Nevertheless, when the reduced representation is inherited from a reconstruction-oriented projection, the learned surrogate may still fail to preserve parameter-sensitive directions that are important for calibration. 

Deep-learning-based reduced-order models provide a nonlinear alternative to projection-based approaches. Convolutional autoencoders, multilayer perceptrons, and recurrent neural-network architectures have been widely used to construct nonlinear latent representations and parametric surrogates for time-dependent physical systems \cite{Federico2024, SimoneBrivio2025, jiayangXu2020}. In particular, autoencoder (AE)-based models can approximate nonlinear solution manifolds more flexibly than linear reduced bases. However, a standard autoencoder is usually trained only by minimizing the reconstruction error. Consequently, same as projection-based approaches, the latent variables are not explicitly constrained to be correlated with physical parameters, time, or other informations. This can lead to latent representations that are accurate for reconstruction but less informative for parameter identification.

Recent developments have explored the incorporation of physical information or observable constraints into latent representations \cite{Fukami2023, yang2022physics, fukami2025observable}. By embedding physical parameters, or other system information into the training process, physics-aware autoencoder architectures can produce more structured latent spaces. This idea is particularly relevant for inverse modeling, because parameter calibration requires a surrogate that is not only accurate in the physical space, but also informative to the parameters being estimated.  
This motivates the observable-augmented autoencoder framework introduced in this work.

\subsection{Inverse problem: Parameter calibration using data assimilation} 

While reduced-order models are often developed for forward prediction, many practical applications require solving the inverse problem of estimating unknown system parameters from observations of the system dynamics. This parameter-calibration task is essential for improving model fidelity and predictive capability, but it is generally ill-posed because different parameter values may generate similar observed responses \cite{BENAISSA2021101451, amoura2022deep}, especially when the available observations are noisy, sparse, or incomplete. Data assimilation provides a principled framework for addressing this problem by combining prior parameter information with observational data in a statistically consistent manner.

Among existing data-assimilation methodologies, sequential filtering approaches, such as the ensemble Kalman filter, and variational formulations (including three- and four-dimensional variational assimilation), have been widely used for parameter estimation while accounting for uncertainties in the background and observations \cite{Cheng_2023,Bocquet_2020,Asch2016DataAssimilation,Carrassi2018DAoverview,egusphere-2025-dainparam,liu2022enkf}. In surrogate-based inverse modeling, a reduced-order model is embedded into the assimilation procedure as a fast forward or observation operator \cite{cheng2023generalised,cheng2024multi}. This strategy avoids repeated full-order simulations and can substantially reduce the online computational cost of parameter calibration. Existing reduced-order data-assimilation frameworks can be broadly grouped by the surrogate type and the assimilation strategy. Projection-based models, such as POD--Galerkin models, have been coupled with variational assimilation for parameter estimation in fluid and heat-transfer problems \cite{chuqiao2025,liu2024application}. However, their reduced coordinates are usually obtained from projection and reconstruction criteria and are not explicitly organized by the physical information. Regression-based non-intrusive surrogates, such as POD-GPR, have also been used with ensemble-based assimilation for large-scale flow and pollutant-dispersion problems \cite{lumet2025POD-GPR,lumet2025ENKF-ESMDA}. Although such surrogates are compatible with ensemble filtering through repeated forward evaluations, their online cost can grow with the ensemble size and observation window, and their reduced representations are not necessarily parameter-identifiable and physics-consistent. Ensemble-based assimilation can rely on repeated surrogate evaluations, whereas variational assimilation requires gradients of the cost function with respect to the control parameters. Therefore, the present framework uses an end-to-end differentiable parameter-to-observation surrogate, enabling efficient optimization of the three- and four-dimensional variational objectives by automatic differentiation.

Therefore, two requirements must be considered simultaneously for deep-learning-based surrogate data assimilation. First, the surrogate should be end-to-end differentiable with respect to the physical control parameters, allowing efficient gradient-based optimization in variational assimilation. Second, its latent representation should preserve information that is relevant to the parameters being calibrated, rather than being optimized only for field reconstruction. Existing projection-based, regression-based, and standard autoencoder-based reduced-order data-assimilation frameworks do not fully address these two requirements at the same time. This motivates the physics-aware neural-network based latent-space variational framework developed in this work, in which observable supervision is used to organize the latent representation and a differentiable parameter-to-latent-to-prediction-to-observation surrogate is used as the observation operator for parameter calibration.

\subsection{Main contributions}
Motivated by the above limitations, this work develops a physics-aware latent-space framework for reduced-order forward modeling and variational parameter calibration. The main contributions of this study are summarized as follows:

\begin{itemize}
    \item We formulate an end-to-end differentiable reduced-order data-assimilation framework for inverse modeling, in which a deep-learning-based surrogate serves as the observation operator for three- and four-dimensional variational parameter estimation.
    
    \item We introduce a physics-aware latent-space surrogate into variational data assimilation by leveraging observable-augmented autoencoding, which improves online calibration accuracy and reduces variability in the physical control-parameter space.
    
    \item We validate the proposed framework on CFD benchmarks by assessing its robustness under degraded-observation settings and by conducting systematic sensitivity analyses of the forward-modeling and inverse-calibration settings, in comparison with standard AE-variational DA and POD-GPR-ensemble DA baselines.
\end{itemize}

\section{Methodology}
\label{sec:Metho}
\subsection{Problem setup}

In this work, we consider parametrized, time-dependent flow systems whose dynamics are represented by high-dimensional state variables evolving on a spatial grid. The primary objective of this work is to construct efficient reduced-order surrogate models capable of approximating the full-order system dynamics and to embed these surrogates into DA frameworks for parameter calibration.

Let $\mathbf{x}_t \in \mathbb{R}^{N_c \times N_x \times N_y}$ denote the discrete solution fields of the system at time $t$, where $N_c$ is the number of physical variables (two in this work, corresponding to the horizontal and vertical velocity components) and $(N_x, N_y)$ define the spatial grid dimensions. The datasets considered in this work consist of multiple simulation cases, each corresponding to a distinct configuration of physical parameters and boundary conditions. For each case, a sequence of time-dependent snapshots $\{\mathbf{x}_t\}_{t=0}^{n_{\text{state}}-1}$ is collected together with the associated parameter values and temporal indices, where $n_{\text{state}}$ denotes the number of snapshots. Each snapshot can therefore be viewed as a data pair
\[
\left( \mathbf{x}_t,\, \boldsymbol{\theta} \right),
\]
where $\boldsymbol{\theta} \in \mathbb{R}^{d_{\theta}}$ gathers the system parameters and the temporal index. From these data, we consider three closely related modeling tasks. First, in the forward modeling and reconstruction setting, the surrogate model takes high-dimensional states $\mathbf{x}_t$ as input and outputs their reconstructed counterparts, with the objective of learning a compact latent representation that preserves the original flow fields. Second, in the forward-prediction setting, the surrogate model aims to approximate the parametric mapping from system parameters to flow states, taking $\boldsymbol{\theta}$ as input and producing the corresponding predicted states $\mathbf{x}_t$. Finally, in the inverse problem addressed by DA, the objective is to infer calibrated system parameters $\boldsymbol{\theta}$ from observed flow states $\mathbf{x}_t$, which may be available at a single time instant or at multiple time instants. In this setting, the forward surrogate models efficiently provide the state predictions required in the assimilation process.

\subsection{Reduced-order modeling approaches}
\label{Sec:ROMs}

\subsubsection{POD-GPR}
\label{sec:POD-GPR}
Due to the high-dimensional nature of $\mathbf{x}_t$, constructing Gaussian process surrogate models directly in the physical space is hindered by the \emph{curse of dimensionality}, stemming from the non-parametric nature of this method. As a result, it is typically combined with dimensionality reduction techniques, such as POD \citep{lumet2025POD-GPR}. Following this idea, we first employ POD to construct a reduced linear subspace from a collection of snapshots.   
We assemble the snapshot matrix
\begin{equation}
\mathcal{X} = \big[\, \mathbf{x}_0 , \mathbf{x}_1 ,\ldots, \mathbf{x}_{n_{\text{state}}-1} \,\big] 
\;\in\; \mathbb{R}^{(N_c \times N_x \times N_y)\times n_{\text{state}}},
\end{equation}
where each column represents a flattened state vector at a given time step. The empirical covariance matrix is then given by
\begin{equation}
\mathbf{C}_{\mathcal{X}} = \frac{1}{n_{\text{state}}-1}\, \mathcal{X}\, \mathcal{X}^T,
\end{equation}
which admits the eigen-decomposition
\begin{equation}
\mathbf{C}_{\mathcal{X}} = \mathbf{L}_{\mathcal{X}}\, \mathbf{D}_{\mathcal{X}}\, \mathbf{L}_{\mathcal{X}}^T,
\end{equation}
where the columns of $\mathbf{L}_{\mathcal{X}} = [\, \Psi_0, \ldots, \Psi_{n_{\text{state}}-1} \,]\in \mathbb{R}^{(N_c \times N_x \times N_y)\times n_{state}}$ are the POD modes, and 
$\mathbf{D}_{\mathcal{X}} = \mathrm{diag}(\lambda_0, \ldots, \lambda_{n_{\text{state}}-1})$ 
contains the eigenvalues in decreasing order.

To obtain a reduced subspace of dimension $q$ ($q\in\mathbb{N}^+,q \ll n_{\text{state}}$), we retain the first $q$ POD modes, forming the projection matrix 
\[
\mathbf{L}_{\mathcal{X},q} = [\, \Psi_0, \ldots, \Psi_{q-1} \,] \;\in\; \mathbb{R}^{(N_c \times N_x \times N_y)\times q}.
\] 
For a given state vector $\mathbf{x}_t$, the reduced latent representation (POD coefficients) is
\begin{equation}
\mathbf{z}_t = \mathbf{L}_{\mathcal{X},q}^T \mathbf{x}_t 
= \big[\, k_{0,t},\, k_{1,t},\, \ldots,\, k_{q-1,t} \,\big]^\top \in\mathbb{R}^q,
\end{equation}
where $k_{j,t}$ denotes the $j$-th POD coefficient at time $t$ and the reconstruction in the original space is
\begin{equation}
\mathbf{x}_t^r = \mathbf{L}_{\mathcal{X},q}\, \mathbf{z}_t.
\end{equation}

To enable forward prediction in the reduced space, the reduced coordinates must be approximated by an additional model. In this work, the regression task is performed using Gaussian process regression (GPR) applied to the POD coefficients \cite{lumet2025POD-GPR}. For completeness, the regression formulation of the POD-GPR surrogate is provided in~\ref{appendix: GPR}.

\subsubsection{OACAE-MLP}
\label{Sec:OACAE-MLP ROM}
While the POD-GPR framework provides an effective surrogate modeling strategy for reduced-order representations, it relies on the assumption that the system dynamics can be captured by a linear subspace and on the regression of reduced POD coefficients that are not explicitly constrained by physical structures. In complex nonlinear flow regimes, however, these assumptions may become restrictive, leading to limited expressiveness and reduced robustness when extrapolating across parameter ranges or over long temporal horizons. 
To overcome these limitations, we present a deep learning based reduced-order model that combines Convolutional AutoEncoder (CAE) with Multilayer Perceptron (MLP), further enhanced by the application of Observable-Augmented Convolutional AutoEncoder (OACAE) \cite{fukami2025observable}. 

\begin{figure}[htbp]
    \centering
    \includegraphics[width=0.8\textwidth]{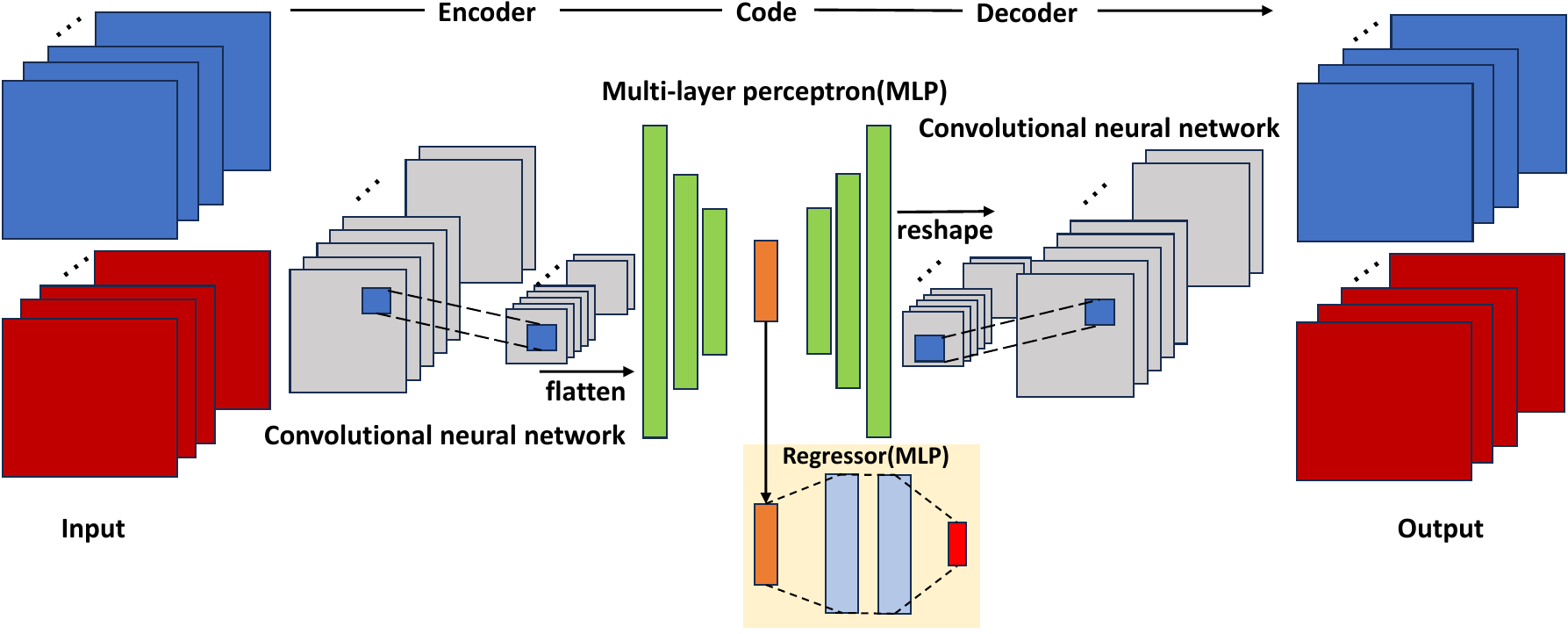}
    \caption{Convolutional autoencoder (CAE) for two-channel velocity fields. The module highlighted in light yellow is activated when embedding parameters into the latent space, yielding the Observable-Augmented CAE (OACAE).}
    \label{fig:cae-mlp-twochannel}
\end{figure}

A CAE is a neural network architecture commonly employed for nonlinear dimensionality reduction and reconstruction of high-dimensional data.  
As illustrated in Figure~\ref{fig:cae-mlp-twochannel}, a CAE model typically consists of an encoder $\mathcal{F}_{e}$ that maps the input field to a low-dimensional latent representation, and a decoder $\mathcal{F}_{d}$ that reconstructs the field from the latent variables.  
Formally, we write
\begin{equation}
    \mathbf{z} \;=\; \mathcal{F}_{e}(\mathbf{x}; \phi_{e}), 
    \qquad 
    \hat{\mathbf{x}} \;=\; \mathcal{F}_{d}(\mathbf{z}; \phi_{d}),
\end{equation}
where $\mathbf{z} \in \mathbb{R}^{q}$ denotes the latent variables and $\phi_{\cdot}$ denotes the weights of the corresponding neural network.
The encoder is composed of successive convolutional layers that extract local spatial features and progressively reduce dimensionality. 
The resulting feature maps are flattened and passed through an MLP to obtain the latent representation. The decoder follows a symmetric architecture based on transposed convolutions to reconstruct the high-dimensional field. By leveraging the locality and translation invariance of convolutional kernels, CAEs are well-suited to capture multiscale flow structures and nonlinear manifolds beyond the representational capability of linear techniques such as POD.

A standard CAE is purely data-driven: it learns a latent representation by minimizing only the reconstruction error.  
As a consequence, the latent variables are not explicitly constrained to be organized with respect to physical parameters or the time index.  Nearby parameter configurations may therefore correspond to weakly correlated latent codes, and different parameter-induced dynamical regimes may partially overlap in the latent space. Such a representation can be sufficient for reconstruction, but it is not necessarily favorable for inverse modeling, where the surrogate must preserve information that is sensitive to the
parameters to be calibrated.

To address this issue, we augment the CAE by incorporating physical parameters into the training process, thereby obtaining a more informative encoder–decoder architecture and guiding the latent space toward physically consistent structures.  
The key idea is to supplement the CAE with an additional MLP branch that links the latent variables to measurable physical quantities, such as fluid density, viscosity, and time index.  
This observable-augmented structure is illustrated by the highlighted branch in light yellow in Figure~\ref{fig:cae-mlp-twochannel}.  

Formally, in the OACAE architecture (notations with superscript $r$ are used to distinguish from the standard CAE), in addition to the usual encoding–decoding relations
\begin{equation}
    \mathbf{z}^r \;=\; \mathcal{F}_{e}^r(\mathbf{x}; \phi_{e}^r), 
    \qquad 
    \hat{\mathbf{x}}^r \;=\; \mathcal{F}_{d}^r(\mathbf{z}^r; \phi_{d}^r),
\end{equation}
we introduce a regression task $\mathcal{E}_a$ from the latent space to the parameter space:
\begin{equation}
    \hat{\boldsymbol{\mathbf{\theta}}} \;=\; \mathcal{E}_a(\mathbf{z}^r; \phi_{a}),
\end{equation}
where $\boldsymbol{\mathbf{\theta}}$ denotes the vector of physical observables, and $\phi_{a}$ are the trainable parameters of the observable regression network.  
The overall training loss combines the field reconstruction error with the regression error of the observables:
\begin{equation}
    \mathcal{L}_1^r 
    \;=\; \|\mathbf{x} - \hat{\mathbf{x}}^r\|_2^2
    \;+\; \alpha \, \|\boldsymbol{\mathbf{\theta}} - \hat{\boldsymbol{\mathbf{\theta}}}\|_2^2,
    \label{eq:L1r}
\end{equation}
where $\alpha>0$ balances the contribution of observable supervision against the reconstruction objective.  
 
\begin{figure}[htbp]
    \centering
    \begin{subfigure}[t]{0.48\textwidth}
        \centering
        \includegraphics[width=\textwidth]{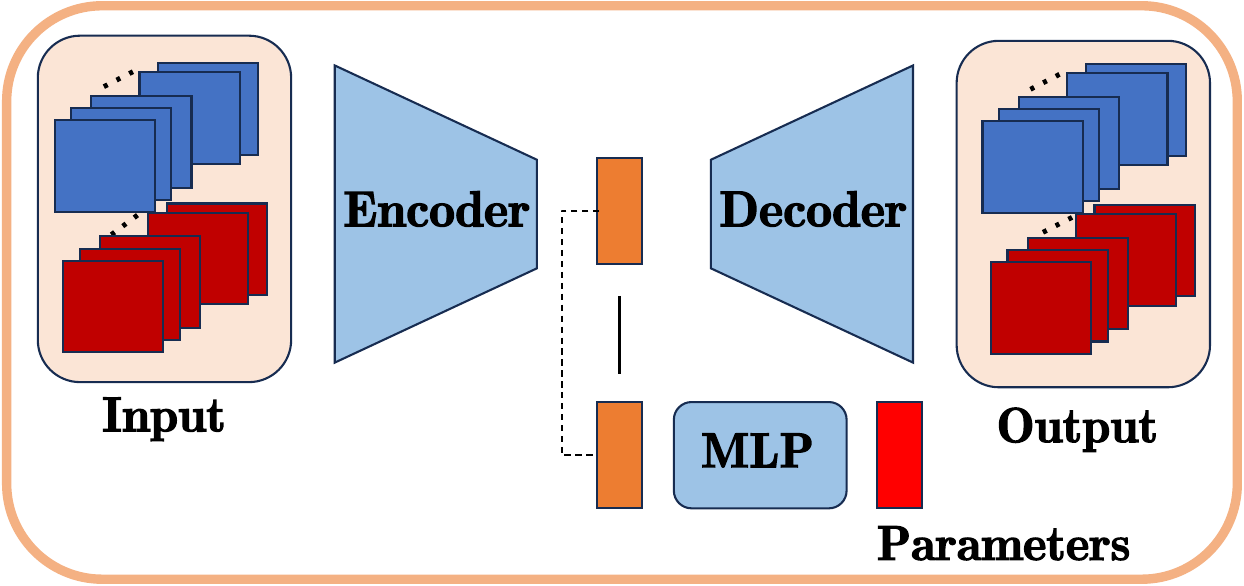}
        \subcaption{Offline training phase 1}
        \label{fig:oacae-training}
    \end{subfigure}
    \hfill
    \begin{subfigure}[t]{0.48\textwidth}
        \centering
        \includegraphics[width=\textwidth]{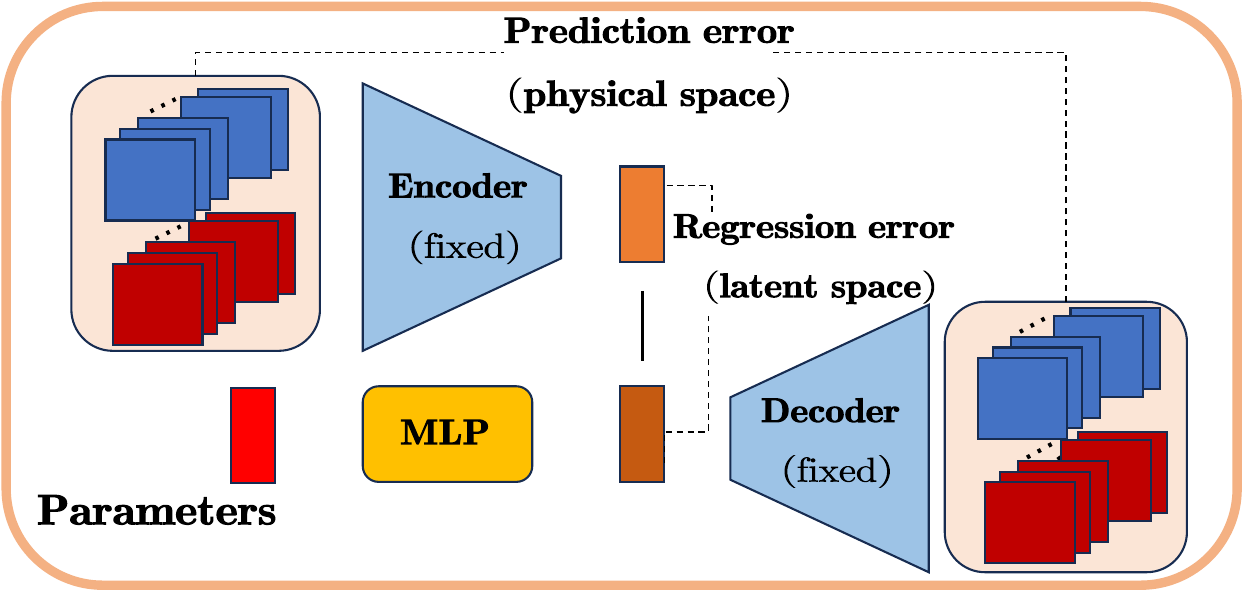}
        \subcaption{Offline training phase 2}
        \label{fig:oacae-mlp-training}
    \end{subfigure}

    \vspace{0.8em}

    \begin{subfigure}[t]{0.6\textwidth}
        \centering
        \includegraphics[width=\textwidth]{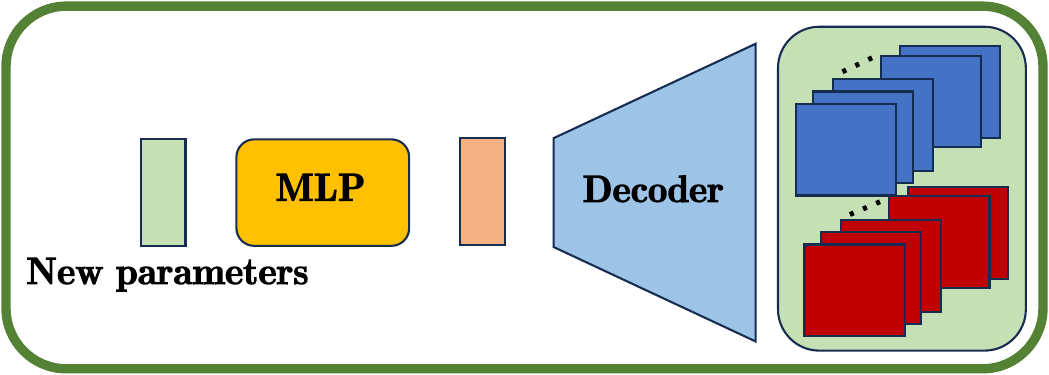}
        \subcaption{Online prediction}
        \label{fig:oacae-mlp-online}
    \end{subfigure}

    \caption{Workflow of the OACAE-MLP surrogate model.
    (a) Offline training of the OACAE with an auxiliary MLP branch $\mathcal{E}_a$ to incorporate physical parameters.
    (b) Offline training of the parameter-to-latent regressor $\mathcal{E}_b$ using the frozen OACAE obtained in phase~(a).
    (c) Online prediction stage, where a new parameter vector $\boldsymbol{\theta}$ is mapped to the latent space by $\mathcal{E}_b$ and decoded by the OACAE decoder to produce the corresponding physical field.}
    \label{fig:oacae-mlp-workflow}
\end{figure}

The observable-supervision term in \(\mathcal{L}_1^r\) encourages the latent variables to retain the
information in the physical fields that is most correlated with the prescribed physical observables, including the system
parameters and the time index. More precisely, different combinations of physical parameters guide the autoencoder to learn distinguishable latent representations for different dynamical scenarios. Furthermore, for a fixed set of physical parameters, the additional guidance from the time index encourages the latent variables to evolve in a temporally consistent and smoother manner. This temporal regularity is particularly important for multi-time-step data assimilation, such as 4D-Var, where the optimization relies on the consistency of the predicted states over a sequence of observation times. The resulting latent-space organization is quantitatively examined in~\ref{app:latent_analysis}, where case-level separability and temporal smoothness metrics are reported for both CAE and OACAE.
  
Once the OACAE model is available, its encoder $\mathcal{F}_{e}^r(\cdot;\phi_e^r)$ and decoder $\mathcal{F}_{d}^r(\cdot;\phi_d^r)$ are frozen and used as fixed functions.  
To construct the forward reduced-order model, we introduce an MLP, denoted by $\mathcal{E}_b$, which maps the system parameters to the observable-augmented latent space (see Figure~\ref{fig:oacae-mlp-training}).  
The predicted latent vector is then decoded by the OACAE decoder to recover the physical field:  
\begin{equation}
    \mathbf{z}^{r,*} \;=\; \mathcal{E}_b(\boldsymbol{\theta};\phi_b^r), 
    \qquad 
    \hat{\mathbf{x}}^{r,*} \;=\; \mathcal{F}_{d}^r(\mathbf{z}^{r,*};\phi_d^r),
\end{equation}
where $\boldsymbol{\theta}\in\mathbb{R}^{d_\theta}$ denotes the normalized input system parameters, 
$\phi_b^r$ is the trainable weights of the regressor, $\mathbf{z}^{r,*}$ is the latent prediction, and $\hat{\mathbf{x}}^{r,*}$ is the final predicted field.  

The training of $\mathcal{E}_b$ is performed in a supervised fashion using pairs $\{(\boldsymbol{\theta}_i,\mathbf{x}_i)\}_{i=1}^N$, as illustrated in Figure~\ref{fig:oacae-mlp-training}.  
For each snapshot, the corresponding reference latent code is precomputed via the frozen encoder, $\mathbf{z}^r_i=\mathcal{F}_{e}^r(\mathbf{x}_i;\phi_e^r)$.  
The objective loss function for $\mathcal{E}_b$ combines a prediction error in the physical space with a regression error in the latent space:  
\begin{equation}
    \mathcal{L}_2^r 
    \;=\; \|\mathbf{x} - \hat{\mathbf{x}}^{r,*}\|_2^2
    + \beta \,\|\mathbf{z}^r - \mathbf{z}^{r,*}\|_2^2 ,
    \label{eq:L2r}
\end{equation}
where $\beta>0$ balances the contribution of latent supervision against field prediction.  
During this training stage, the OACAE encoder and decoder remain fixed, so that only the parameters $\phi_b^r$ of the MLP $\mathcal{E}_b$ are updated.  
This workflow provides a non-intrusive parametric ROM that directly maps physical parameters to physical fields via the augmented latent representation, while leveraging the structure learned by the OACAE. 
 In the online stage (see Figure~\ref{fig:oacae-mlp-online}), this allows fast and repeated evaluations: once trained, the model can efficiently generate large numbers of flow-field predictions for arbitrary parameter inputs without resorting to expensive numerical simulations.
  
The OACAE and its observable branch are implemented using a standard convolutional encoder-decoder architecture with MaxPooling layers and fully connected layers, complemented by additional MLP components.  
The detailed layer-by-layer configurations of the OACAE encoder/decoder, the observable regression branch MLP ($\mathcal{E}_a$), and the parameter-to-latent regressor MLP ($\mathcal{E}_b$) are provided in~\ref{appendix:nn-structures}.  

As illustrated in Figure~\ref{fig:oacae-mlp-workflow}, the training procedure of the OACAE-MLP ROM consists of two successive phases.  
In the first phase (Figure~\ref{fig:oacae-training}), the parameters of the encoder, decoder, and observable branch ($\phi_e^r, \phi_d^r, \phi_a^r$) are optimized by minimizing the loss $\mathcal{L}_1^r$ defined in Equation~\eqref{eq:L1rphi}.  
In the second phase (Figure~\ref{fig:oacae-mlp-training}), the encoder and decoder are kept frozen, and the parameter-to-latent regressor $\mathcal{E}_b$ with parameters $\phi_b^r$ is trained by minimizing $\mathcal{L}_2^r$ in Equation~\eqref{eq:L2rphi}.  
For clarity, we rewrite the two objective functions below to explicitly highlight their dependence on the trainable parameters:  
\begin{equation}
    \mathcal{L}_1^r(\phi_a^r, \phi_e^r, \phi_d^r) 
    \;=\; \|\mathbf{x} - \mathcal{F}_{d}^r( \mathcal{F}_{e}^r(\mathbf{x}; \phi_{e}^r); \phi_{d}^r)\|_2^2
    \;+\; \alpha \, \|\boldsymbol{\theta}-\mathcal{E}_a(\mathcal{F}_{e}^r(\mathbf{x}; \phi_{e}^r); \phi_{a}^r)\|_2^2,
    \label{eq:L1rphi}
\end{equation}
\begin{equation}
    \mathcal{L}_2^r(\phi_b^r) 
    \;=\; \|\mathbf{x} - \mathcal{F}_{d}^r(\mathcal{E}_b(\boldsymbol{\theta};\phi_b^r);\phi_d^r)\|_2^2
    + \beta \,\|\mathcal{F}_{e}^r(\mathbf{x}; \phi_{e}^r) - \mathcal{E}_b(\boldsymbol{\theta};\phi_b^r)\|_2^2 ,
    \label{eq:L2rphi}
\end{equation}
with the corresponding optimization problems solved by the Adam optimizer:  
\begin{equation}
    (\phi_a^r, \phi_e^r, \phi_d^r)^* \;=\; \arg\min_{(\phi_a^r, \phi_e^r, \phi_d^r)} \,\mathcal{L}_1^r(\phi_a^r, \phi_e^r, \phi_d^r),
\end{equation}
\begin{equation}
    (\phi_b^r)^* \;=\; \arg\min_{(\phi_b^r)} \,\mathcal{L}_2^r(\phi_b^r).
\end{equation}

In this study, the weighting coefficients are set to $\alpha=0.05$ and $\beta=1$ based on an empirical assessment of the trained model performance (see~\ref{app:loss_weight_sensitivity} for a detailed analysis of the choice of these coefficients).  

In addition, we denote by CAE-MLP the baseline model obtained by replacing the OACAE in the OACAE-MLP framework with a standard CAE, that is, by setting $\alpha=0$ and removing the observable regression branch. Accordingly, the CAE-MLP can be regarded as a special case of the proposed OACAE-MLP model, and is introduced to assess the impact of physics-aware latent supervision. In this case, the model consists of a standard CAE with an encoder $\mathcal{F}_e$ and a decoder $\mathcal{F}_d$ together with a parameter-to-latent regressor MLP, denoted by $\mathcal{E}_c$.  

Formally, the encoder-decoder relations of the standard CAE read
\begin{equation}
    \mathbf{z} \;=\; \mathcal{F}_{e}(\mathbf{x}; \phi_{e}), 
    \qquad 
    \hat{\mathbf{x}} \;=\; \mathcal{F}_{d}(\mathbf{z}; \phi_{d}),
\end{equation}
where $\mathbf{z}$ is the latent code, and $\phi_{e},\phi_{d}$ are the trainable weights of the encoder and decoder.  

The training loss for the CAE stage is thus given by
\begin{equation}
    \mathcal{L}_1(\phi_e,\phi_d) 
    \;=\; \|\mathbf{x} - \mathcal{F}_{d}(\mathcal{F}_{e}(\mathbf{x}; \phi_{e}); \phi_{d})\|_2^2.
    \label{eq:L1_cae}
\end{equation}

In the second stage, a parameter-to-latent regressor $\mathcal{E}_c$ with weights $\phi_c$ is trained to map input parameters $\boldsymbol{\theta}$ to the latent space, while the decoder $\mathcal{F}_d$ remains frozen.  
The corresponding loss function is
\begin{equation}
    \mathcal{L}_2(\phi_c) 
    \;=\; \|\mathbf{x} - \mathcal{F}_{d}(\mathcal{E}_c(\boldsymbol{\theta};\phi_c);\phi_d)\|_2^2
    + \beta \,\|\mathcal{F}_{e}(\mathbf{x}; \phi_{e}) - \mathcal{E}_c(\boldsymbol{\theta};\phi_c)\|_2^2,
    \label{eq:L2_cae}
\end{equation}
with optimization problems
\begin{equation}
    (\phi_e,\phi_d)^* \;=\; \arg\min_{(\phi_e,\phi_d)} \mathcal{L}_1(\phi_e,\phi_d), 
    \qquad 
    (\phi_c)^* \;=\; \arg\min_{\phi_c} \mathcal{L}_2(\phi_c).
\end{equation}

\subsection{Data assimilation frameworks for parameter calibration}

In this section, we present three complementary DA frameworks that leverage the previously 
introduced ROMs. First, we discuss the use of the Ensemble Kalman Filter (EnKF) combined with the POD-GPR surrogate for parameter estimations. Second, we formulate variational assimilation strategies (3D-Var/4D-Var) combined with CAE-MLP and OACAE-MLP, allowing for optimization strategies of parameter estimates.

\subsubsection{EnKF combined with POD-GPR}
\label{sec: ENKF}
Among the available data assimilation techniques, the EnKF is adopted in combination with the POD-GPR surrogate due to its derivative-free and ensemble-based formulation. In contrast to variational approaches, which require gradients of the reduced-order model with respect to the parameters, the POD-GPR surrogate does not provide analytic derivatives and is therefore not directly compatible with gradient-based optimization methods. The EnKF addresses the calibration problem within a statistical framework by propagating an ensemble of parameter realizations, making it particularly well suited to handle the variability and nonlinearity inherent in POD-GPR predictions.
Moreover, the use of ensembles naturally represents uncertainty while maintaining computational efficiency, making EnKF a natural benchmark for parameter calibration with POD-GPR.
  
The Ensemble Kalman Filter (EnKF) is a sequential Monte Carlo method that extends the 
classical Kalman filter to high-dimensional nonlinear problems.  
It represents the uncertainty of the control vector by an ensemble of realizations, 
which are propagated forward by the model and used to compute the Kalman gain in a 
statistical manner.  
Given a background ensemble 
$\mathbf{E}^b = (\boldsymbol{\theta}_1^b,\ldots,\boldsymbol{\theta}_{N_e}^b)^\top 
\in \mathbb{R}^{d\times N_e}$ 
sampled from a prior distribution $\mathcal{N}(\bar{\boldsymbol{\theta}}^b,\mathbf{B})$, where $\bar{\boldsymbol{\theta}}^b$ denotes the background mean parameter vector (typically chosen as the empirical mean of the training dataset) and $\mathbf{B}$ represents the background error covariance matrix characterizing the prior uncertainty in the parameters, estimated from the empirical covariance of the training parameter ensemble, computed as the sample covariance of the parameter anomalies with respect to the ensemble mean. The EnKF proceeds as follows \cite{lumet2025ENKF-ESMDA}:

\emph{(i) Forecast step.}  
Each ensemble member is directly evaluated through the observation operator $\mathcal{H}$, yielding
\begin{equation}
\mathbf{x}_i^f = \mathcal{H}(\boldsymbol{\theta}_i^b), \qquad i=1,\ldots,N_e.
\end{equation}

\emph{(ii) Analysis step.}  
The forecast ensemble $\{\mathbf{x}_i^f\}$ is compared with the observations $\mathbf{y}$.  
The error cross-covariances are estimated from the ensemble as
\begin{equation}
\mathbf{B}\mathbf{H}^\top \;\approx\; 
\frac{1}{N_e-1}\sum_{i=1}^{N_e} 
(\boldsymbol{\theta}_i^b - \bar{\boldsymbol{\theta}}^b)
\big(\mathbf{x}_i^f-\overline{\mathbf{x}^f}\big)^\top ,
\end{equation}
\begin{equation}
\mathbf{H}\mathbf{B}\mathbf{H}^\top \;\approx\; 
\frac{1}{N_e-1}\sum_{i=1}^{N_e} 
\big(\mathbf{x}_i^f-\overline{\mathbf{x}^f}\big)
\big(\mathbf{x}_i^f-\overline{\mathbf{x}^f}\big)^\top .
\end{equation}
The Kalman gain is then
\begin{equation}
\mathbf{K}^* \;=\; 
\mathbf{B}\mathbf{H}^\top\big(\mathbf{H}\mathbf{B}\mathbf{H}^\top+\mathbf{R}\big)^{-1},
\end{equation}
where $\mathbf{R}$ denotes the observation error covariance matrix that 
quantifies measurement uncertainties.  

Finally, each parameter ensemble member is updated according to
\begin{equation}
\boldsymbol{\theta}_i^a = \boldsymbol{\theta}_i^b 
+ \mathbf{K}^*\big(\mathbf{y}_i - \mathbf{x}_i^f\big), 
\qquad i=1,\ldots,N_e,
\end{equation}
where $\mathbf{y}_i$ are perturbed observations sampled from 
$\mathcal{N}(\mathbf{y},\mathbf{R})$.  

The optimal parameter estimate is obtained as the ensemble mean
\begin{equation}
\bar{\boldsymbol{\theta}}^a = \frac{1}{N_e}\sum_{i=1}^{N_e}\boldsymbol{\theta}_i^a.
\end{equation}
 
In our framework, the EnKF is applied directly to the system parameters 
$\boldsymbol{\theta}$.  
The complete parameter vector must be provided as input to the POD-GPR surrogate, 
which serves as the observation operator in the data assimilation process.  
This vector includes both the physical parameters of the system and the time index 
associated with each snapshot.  
Accordingly, a single $\boldsymbol{\theta}$ corresponds to one snapshot, 
whereas the combination of fixed physical parameters with a sequence of time indices 
describes the entire flow evolution.  
For parameter calibration, however, we restrict the update to the most sensitive 
physical parameters, denoted by $\boldsymbol{\theta}^{\mathrm{control}}$, while the remaining 
physical parameters and the snapshot time index (e.g., $t_k$) are considered as known, 
denoted by $\boldsymbol{\theta}^{\mathrm{fixed}}_{t_k}$.  
Thus, each parameter vector at time $t_k$ can be represented as  
\begin{equation}
    \boldsymbol{\theta}_{t_k} = \big(\boldsymbol{\theta}^{\mathrm{control}},\; \boldsymbol{\theta}^{\mathrm{fixed}}_{t_k}\big).
\end{equation}
The observation operator $\mathcal{H}$ is identified with the POD-GPR surrogate model, 
denoted $\mathcal{H}_{\text{POD-GPR}}$, which maps a complete parameter vector 
$\boldsymbol{\theta}_{t_k}$ to a prediction of the physical field.  
The corresponding observation $\mathbf{y}$ is taken from the true physical field 
(e.g., flattened snapshots of high-fidelity simulations).  
This leads to the following assimilation stages at time $t_k$:  

\begin{enumerate}
    \item \emph{Prediction step:} each ensemble member $\boldsymbol{\theta}_i^{\text{control},b}$ 
    is combined with the fixed parameters and the time index $t_k$ to form 
    \begin{equation}
        \boldsymbol{\theta}_{t_k,i} = \big(\theta^{\mathrm{control},b}_i ,\; \theta^{\mathrm{fixed}}_{t_k}\big),
    \end{equation}
    and propagated through the surrogate to obtain the forecast field  
    $\mathbf{x}_i^f = \mathcal{H}_{\text{POD-GPR}}(\boldsymbol{\theta}_{t_k,i})$.  
    \item \emph{Analysis step:} the discrepancy between the ensemble forecasts 
    $\mathbf{x}_i^f$ and the perturbed observations is used to update the parameter ensemble 
    according to the EnKF equations.  
    \item \emph{Ensemble mean:} the calibrated parameter estimate is given by the mean 
    of the analysis ensemble,  
    $\bar{\boldsymbol{\theta}}^{\text{control},a} = \tfrac{1}{N_e}\sum_{i=1}^{N_e}\boldsymbol{\theta}_i^{\text{control},a}$.
\end{enumerate}

This procedure corresponds to a single-time-step EnKF strategy, often applied in the 
calibration of stationary or time-averaged fields, where the update of the control parameters relies on observations at a single time instant.  
In our study, however, the inclusion of the time index in the parameter vector allows 
us to extend the assimilation to a multi-time-step EnKF.  
Consider a sequence of snapshots from one evolution case,  
\begin{equation}
    \mathbf{x}_{t_0}, \mathbf{x}_{t_1}, \mathbf{x}_{t_2}, \ldots, \mathbf{x}_{t_k},
\end{equation}
which serve as observations at successive assimilation steps.  
Starting from a prior control vector, the EnKF update with $\mathbf{x}_{t_0}$ yields 
an analysis control vector, which is then taken as the background for the update 
with $\mathbf{x}_{t_1}$, and so on.  
In this way, observations from multiple time steps are successively assimilated, 
and the final analysis control vector incorporates observational information from 
the entire sequence of snapshots. 

\subsubsection{3D-Var and 4D-Var combined with CAE-MLP and OACAE-MLP}
A key challenge in variational data assimilation lies in defining the forward and observation operators, which often involve nonlinear and high-dimensional mappings. In particular, integrating deep learning models into DA workflows remains difficult, as most existing software packages do not natively support neural networks within variational schemes.  To bypass this limitation, we rely on the recently proposed Python package TorchDA~\cite{cheng2025torchda}, which provides a unified framework for coupling deep learning surrogates with variational DA algorithms.  Built on the PyTorch ecosystem, TorchDA streamlines 3D-Var/4D-Var workflows by enabling neural networks to serve directly as forward or observation operators, with automatic differentiation for tangent-linear and adjoint computations, GPU acceleration, and user-friendly configuration interfaces.  
This capability allows us to seamlessly integrate DL-based ROMs, such as the CAE-MLP and OACAE-MLP introduced in Section~3.2, into variational assimilation pipelines.

Variational DA seeks to estimate the optimal state or parameter vector $\boldsymbol{\theta}$ by minimizing a cost function that balances background information with observational constraints. 
Here, we focus on parameter calibration using variational DA with DL-based observation operators. In the 3D-Var case, where assimilation is performed at a single time step, the cost function reads
\begin{equation}
    J_{\textbf{3D-Var}}(\boldsymbol{\theta}) \;=\; 
    \tfrac{1}{2}\big(\boldsymbol{\theta} - \boldsymbol{\theta}^b\big)^\top 
    \mathbf{B}^{-1}
    \big(\boldsymbol{\theta} - \boldsymbol{\theta}^b\big) \;+\;
    \tfrac{1}{2}\big(\mathbf{y}_0 - \mathcal{H}(\boldsymbol{\theta})\big)^\top 
    \mathbf{R}^{-1}
    \big(\mathbf{y}_0 - \mathcal{H}(\boldsymbol{\theta})\big),
\end{equation}
where $\boldsymbol{\theta}^b$ denotes the background parameter vector, 
$\mathbf{B}$ the background error covariance matrix, $\mathbf{y}_0$ the observation, 
$\mathcal{H}$ the DL-based observation operator, and $\mathbf{R}$ the observation error 
covariance matrix.  

The 4D-Var framework extends this principle to a temporal window 
$[t_0, \ldots, t_K]$, by assimilating all observations within this interval:  
\begin{equation}
    J_{\textbf{4D-Var}}(\boldsymbol{\theta}) \;=\; 
    \tfrac{1}{2}\big(\boldsymbol{\theta} - \boldsymbol{\theta}^b\big)^\top 
    \mathbf{B}^{-1}
    \big(\boldsymbol{\theta} - \boldsymbol{\theta}^b\big) \;+\;
    \tfrac{1}{2}\sum_{k=0}^K
    \big(\mathbf{y}_{t_k} - \mathcal{H}_{t_k}(\boldsymbol{\theta})\big)^\top 
    \mathbf{R}^{-1}
    \big(\mathbf{y}_{t_k} - \mathcal{H}_{t_k}(\boldsymbol{\theta})\big),
\end{equation}
where $\mathcal{H}_{t_k}(\boldsymbol{\theta})$ denotes the surrogate-predicted field at time $t_k$ corresponding to the parameter vector $\boldsymbol{\theta}$.  Unlike classical 4D-Var formulations, where the control variable evolves in time according to the governing dynamics, the present 4D-Var treats the target parameter vector $\boldsymbol{\theta}$ as a fixed control variable, which is mapped to multiple observation times through distinct observation operators $\mathcal{H}_{t_k}$. The minimization of $J_{\textbf{3D-Var}}$ and $J_{\textbf{4D-Var}}$ is carried out iteratively, often using gradient-based optimizers such as Adam or SGD, with gradients obtained by automatic differentiation of the DL-based surrogates enabled by TorchDA. A learning-rate sensitivity sweep is reported in~\ref{app:learning_rate} to examine the stability of this gradient-based optimization and to identify a suitable range of learning rates.

We illustrate the application of the framework using the CAE-MLP model, 
noting that the procedure for the OACAE-MLP surrogate is entirely analogous.  
In this setting, the reduced-order surrogate CAE-MLP serves as the observation 
operator $\mathcal{H}$.  
As defined in Section~\ref{sec: ENKF}, the complete parameter vector is 
$\boldsymbol{\theta}_{t_k} = \big(\boldsymbol{\theta}^{\mathrm{control}},\; \boldsymbol{\theta}^{\mathrm{fixed}}_{t_k}\big)$, 
which includes the time-index parameter $t_k$ and is used as input to the surrogate.  
The control variables $\boldsymbol{\theta}^{\mathrm{control}}$ act as the independent variables 
in the objective cost functions.  
Accordingly, the 3D-Var and 4D-Var cost functions can be written as
\begin{align}
    \label{eq:3dvar costfunc}
    J_{\textbf{3D-Var}}(\boldsymbol{\theta}^{\mathrm{control}}) \;=\; 
    & \tfrac{1}{2}\big(\boldsymbol{\theta}^{\mathrm{control}} - \boldsymbol{\theta}^{\mathrm{control},b}\big)^\top 
    \mathbf{B}^{-1}
    \big(\boldsymbol{\theta}^{\mathrm{control}} - \boldsymbol{\theta}^{\mathrm{control},b}\big) \;+\nonumber\\
    & \tfrac{1}{2}\big(\mathbf{x}_{t_k} - \mathcal{H}_{t_k}(\boldsymbol{\theta}^{\mathrm{control}})\big)^\top 
    \mathbf{R}^{-1}
    \big(\mathbf{x}_{t_k} - \mathcal{H}_{t_k}(\boldsymbol{\theta}^{\mathrm{control}})\big),
\end{align}
\begin{align}
    \label{eq:4dvar costfunc}
    J_{\textbf{4D-Var}}(\boldsymbol{\theta}^{\mathrm{control}}) \;=\; 
    & \tfrac{1}{2}\big(\boldsymbol{\theta}^{\mathrm{control}} - \boldsymbol{\theta}^{\mathrm{control},b}\big)^\top 
    \mathbf{B}^{-1}
    \big(\boldsymbol{\theta}^{\mathrm{control}} - \boldsymbol{\theta}^{\mathrm{control},b}\big) \;+\nonumber\\
    & \tfrac{1}{2} \sum_{k=0}^K
    \big(\mathbf{x}_{t_k} - \mathcal{H}_{t_k}(\boldsymbol{\theta}^{\mathrm{control}})\big)^\top 
    \mathbf{R}^{-1}
    \big(\mathbf{x}_{t_k} - \mathcal{H}_{t_k}(\boldsymbol{\theta}^{\mathrm{control}})\big),
\end{align}
where, for a given control parameter vector $\boldsymbol{\theta}^{\mathrm{control}}$, 
the surrogate prediction of the physical field is expressed as
\begin{equation}
    \mathcal{H}_{t_k}(\boldsymbol{\theta}^{\mathrm{control}}) 
= \mathcal{H}_{\text{CAE-MLP}}(\boldsymbol{\theta}^{\mathrm{control}},\boldsymbol{\theta}^{\mathrm{fixed}}_{t_k})=\mathcal{F}_{d}(\mathcal{E}_c(\boldsymbol{\theta}^{\mathrm{control}},\boldsymbol{\theta}^{\mathrm{fixed}}_{t_k})).
\end{equation}
where $\mathcal{F}_{d}$ denotes the decoder and $\mathcal{E}_c$ the parameter-to-latent MLP regressor, as introduced in Sec~\ref{Sec:OACAE-MLP ROM}.
The variational assimilation proceeds as follows.  
First, the background control vector $\boldsymbol{\theta}^{\mathrm{control},b}$ is defined as the empirical mean of training parameters, and the associated background covariance $\mathbf{B}$ is estimated from the training dataset as well.  At each observation time $t_k$, the CAE-MLP surrogate predicts the physical field corresponding to the current complete parameter vector.  The discrepancy between the predicted field and the flattened true snapshot $\mathbf{x}_{t_k}$ (used here as the observation) is then evaluated in the 3D-Var/4D-Var cost function. Gradients of the objective function with respect to $\boldsymbol{\theta}^{\mathrm{control}}$ are computed via backpropagation through the CAE-MLP network using TorchDA, and the control vector is iteratively updated with a gradient-based optimizer until convergence, yielding the analysis estimate $\boldsymbol{\theta}^{\mathrm{control},a}$. Through this procedure, observational information distributed over multiple time steps can be effectively assimilated into the control parameters, while leveraging the differentiable structure of the CAE-MLP surrogate. 
Since the balance between the background and observation terms depends on the relative scaling of $\mathbf{B}$ and $\mathbf{R}$, we also report a covariance-scaling sensitivity study in~\ref{app:covariance_scaling}.

\section{Numerical Experiments}
\label{sec: Experiment}
We evaluate the presented and proposed ROM-DA frameworks on two representative CFDBench problems: the dam-break flow and the lid-driven cavity flow. To emphasize the contribution of the proposed physics-aware variational DA-DL-ROM approach, we adopt the EnKF-POD-GPR framework as a baseline, following the methodology introduced in Sec~\ref{sec:POD-GPR} and Sec~\ref{sec: ENKF}, and inspired by the idea from~\cite{lumet2025POD-GPR,lumet2025ENKF-ESMDA}. In addition to the EnKF-POD-GPR baseline, we compare CAE-MLP and OACAE-MLP to isolate the effect of observable supervision. These two models share the same encoder-decoder backbone, latent dimension, parameter-to-latent regression strategy, and variational assimilation setting. The main difference is that OACAE-MLP uses observable supervision during the autoencoder training stage, whereas CAE-MLP relies only on the reconstruction loss. Therefore, the comparison between CAE-MLP and OACAE-MLP serves as a controlled ablation for assessing the effect of observable supervision on latent-space organization, forward prediction, and inverse calibration.

To rigorously assess the generalization capability of the surrogate models, data splitting is performed at the case level rather than at the snapshot level. All snapshots associated with a given simulation case are assigned exclusively to either the training or the test set. This strategy prevents information leakage across different parameter realizations and provides a realistic evaluation of model performance on unseen configurations. Both the solution fields and the associated parameter data are normalized using feature-wise min--max scaling to the range $[0,1]$. This normalization ensures consistent scaling across different state variables and parameters, avoids numerical imbalances, and facilitates the convergence of learning algorithms. The same normalization procedure is applied consistently throughout the reduced-order forward modeling and DA inverse modeling stages.
Before fixing the final inverse-modeling setting, we performed preliminary multi-parameter assimilation tests in which all available candidate parameters were included as control variables. These tests were used to assess the practical identifiability of each parameter from the available observations. Parameters that remained weakly identifiable, led to unstable assimilation updates, or produced calibrated errors comparable to or larger than the background errors were fixed in the subsequent experiments. This reduced control setting avoids over-parameterizing the inverse problem and enables a clearer assessment of the achievable inverse modeling performance from the available observations.

\subsection{The CFDBench dataset}
\label{sec:CFDBench}
To evaluate data-driven reduced-order models and their integration with data assimilation frameworks, we rely on the recently released CFDBench dataset~\cite{luo2024cfdb}. CFDBench provides a large-scale benchmark for assessing the generalization ability of machine learning surrogates in computational fluid dynamics (CFD). Among the problems included in CFDBench, the two-dimensional dam-break flow and the lid-driven cavity flow are selected as the primary test cases in this study.

The operating parameters in CFDBench are systematically varied along three main axes: (i) boundary conditions (e.g., inlet or lid velocity), (ii) fluid properties (density $\rho$ and viscosity $\mu$), and (iii) geometry (e.g., cavity dimensions, tube diameter, barrier height, or cylinder radius). For each flow configuration, cases are organized into subsets corresponding to these parameter variations and their combinations, enabling rigorous evaluation of model extrapolation to unseen operating conditions.

\subsection{Dam-break flow problem}
\label{sec:dam}
Dam-break flows constitute a canonical class of gravity-driven free-surface problems and are widely used as simplified models for coastal inundation and nearshore wave dynamics in ocean and coastal engineering. The sudden release of an initially retained water column generates rapidly propagating free-surface waves characterized by strong nonlinearity, sharp gradients, and transient flow structures. Owing to these features, dam-break configurations serve as challenging yet physically meaningful benchmarks for assessing numerical solvers, reduced-order models, and inverse methodologies in regimes relevant to oceanic and coastal modeling. In this work, the dam-break flow is adopted to evaluate the performance of the proposed ROM--DA frameworks under strongly nonlinear dynamics.

In CFDBench, the dam flow is parameterized by the height $h$ and width $w$ of the dam obstacle, the inlet velocity $u_{in}$, the fluid density $\rho$, and the fluid viscosity $\mu$. By systematically varying these parameters, the dataset spans a wide range of Reynolds numbers (see Table~\ref{tab:dam_params}).

\begin{table}[htbp]
\centering
\caption{Parameter ranges for the dam flow problem in CFDBench.}
\label{tab:dam_params}
\begin{tabular}{ll}
\hline
Category & Values \\
\hline
Boundary condition 
& $u_{in} \in \{0.05, 0.1, \ldots, 1\} \cup \{1.02, 1.04, \ldots, 2\}$ m/s \\

Fluid density 
& $\rho \in \{0.1, 0.5, 1, 2, 3, \ldots, 10\}$ kg/m$^3$ \\

Fluid viscosity 
& $\mu \in \{10^{-5},\, 5\times10^{-5},\, \ldots,\, 5\times10^{-3},\, 10^{-2}\}$ Pa$\cdot$s \\

Barrier height 
& $h \in \{0.11, 0.12, 0.13, 0.14, 0.15\}$ m \\

Barrier width 
& $w \in \{0.01, 0.02, \ldots, 0.08, 0.09\}$ m \\
\hline
\end{tabular}
\end{table}

For each simulation case, the velocity components $(u,v)$ are stored at successive time steps on a uniform $64 \times 64$ grid, yielding a total of $21{,}700$ snapshots across $217$ evolution cases. Each snapshot is associated with a set of physical parameters and a temporal index, and can be represented as
\[
\left( \mathbf{x}_t,\, \boldsymbol{\theta} \right),
\qquad
\mathbf{x}_t \in \mathbb{R}^{2 \times 64 \times 64}, \quad
\boldsymbol{\theta} \in \mathbb{R}^6,
\]
where $\mathbf{x}_t$ denotes the velocity field at time $t$ and $\boldsymbol{\theta}$ collects the five physical parameters together with the time index.

The dataset is organized into three groups of cases: $100$ cases varying fluid properties $(\rho,\mu)$, $50$ cases varying the cavity geometry $(h,w)$, and $67$ cases varying the boundary condition $u_{in}$. For model evaluation, data splitting is performed at the case level, ensuring that each simulation case is assigned to either the training or the test set. This results in $173$ training cases ($17{,}300$ snapshots) and $44$ test cases ($4{,}400$ snapshots), providing a strict assessment of generalization to unseen parameter configurations.

The first row of Figure~\ref{fig:prediction-dam-cavity} illustrates representative snapshots of the horizontal and vertical velocity components in the dam-break flow problem. The presence of gravity-driven jets, strong velocity gradients, flow separation around the obstacle, and free-surface-induced recirculation zones leads to highly transient and nonlinear flow structures. These features make the dam-break flow a particularly challenging benchmark for evaluating the reconstruction, prediction, and parameter calibration capabilities of the proposed ROM-DA frameworks.

In our numerical experiment, the inlet velocity $u_{in}$ and the barrier height $h$ are identified as the primary calibration parameters, while the remaining parameters are set to their true values. Unless otherwise stated, all reported calibration results correspond to this reduced control setting.

\subsection{Lid-driven cavity flow problem}
\label{sec:cavity}
The lid-driven cavity flow is a classical benchmark in CFD and provides a controlled setting to investigate both flow reconstruction/prediction and parameter calibration. In contrast to the dam-break flow, which is dominated by strongly nonlinear free-surface dynamics, the cavity flow represents a confined, wall-bounded shear-driven system, thereby offering a complementary test case to assess the generality and robustness of the proposed ROM--DA frameworks. The flow is confined within a square cavity, where the top lid moves with velocity $u_{lid}$, while the remaining walls are stationary and impose no-slip conditions. In CFDBench, the cavity flow is parameterized by the cavity geometry (length $l$ and width $w$), the lid velocity $u_{lid}$, the fluid density $\rho$, and the fluid viscosity $\mu$. By systematically varying these parameters, the dataset spans a wide range of Reynolds numbers (see Table~\ref{tab:cavity_params}).

\begin{table}[htbp]
\centering
\caption{Parameter ranges for the cavity flow problem in CFDBench.}
\label{tab:cavity_params}
\begin{tabular}{ll}
\hline
Category & Values \\
\hline
Boundary condition & $u_{lid} \in \{1,2,3,\ldots,50\}$ m/s \\
Fluid density & $\rho \in \{0.1,0.5,1,\ldots,10\}$ kg/m$^3$ \\
Fluid viscosity & $\mu \in \{10^{-5},\, 5\times10^{-5},\, 10^{-4},\, \ldots,\, 5\times10^{-3},\, 10^{-2}\}$ Pa$\cdot$s \\
Geometry & $l,w \in \{0.01,0.02,0.03,0.04,0.05\}$ m \\
\hline
\end{tabular}
\end{table}

For each simulation case, the velocity components $(u,v)$ are stored at successive time steps on a uniform $64 \times 64$ grid, yielding a total of $12{,}681$ snapshots across $158$ cases. Each snapshot is associated with a set of physical parameters and a temporal index, and can be represented as
\[
\left( \mathbf{x}_t,\, \boldsymbol{\theta} \right),
\qquad
\mathbf{x}_t \in \mathbb{R}^{2 \times 64 \times 64}, \quad
\boldsymbol{\theta} \in \mathbb{R}^6,
\]
where $\mathbf{x}_t$ denotes the velocity field at time $t$ and $\boldsymbol{\theta}$ collects the five physical parameters together with the time index.

The dataset is organized into three groups of cases: $84$ cases varying fluid properties $(\rho,\mu)$, $24$ cases varying the cavity geometry $(l,w)$, and $50$ cases varying the boundary condition $u_{lid}$. For model evaluation, data splitting is performed at the case level, ensuring that each simulation case is assigned to either the training or the test set. This results in $144$ training cases ($8{,}036$ snapshots) and $14$ test cases ($4{,}645$ snapshots), providing a strict assessment of generalization to unseen parameter configurations.

The fifth row of Figure~\ref{fig:prediction-dam-cavity} illustrates representative snapshots of the horizontal and vertical velocity components on the cavity domain. The emergence of secondary vortices and increasingly complex flow structures at higher Reynolds numbers makes this benchmark particularly suitable for evaluating the reconstruction, prediction, and parameter calibration capabilities of ROM-DA frameworks.

In our numerical experiment, the lid velocity $u_{lid}$ and the fluid density $\rho$ are identified as the primary control parameters for calibration, while the remaining parameters are fixed at their true values. Unless otherwise stated, all reported calibration results correspond to this reduced control setting.

\subsection{Reconstruction}
\label{sec:reconstruction}
We first evaluate the reconstruction capabilities of the three reduced-order models introduced in 
Section~\ref{Sec:ROMs}, namely POD, standard CAE, and OACAE.  

To provide a quantitative comparison, we evaluate the average reconstruction errors over the entire test dataset. For fairness, the latent dimension is fixed to 32 for all three ROMs. This value is selected as a robust and balanced choice according to the latent-dimension sensitivity study reported in~\ref{app:latent_dimension_sensitivity}. The results, summarized in Table~\ref{tab:reconstruction-errors}, are reported in terms of relative $\ell_2$ error (REL), mean square error (MSE), and structural similarity index (SSIM). These metrics highlight the contrast between the linear POD coefficients and the nonlinear latent manifolds constructed by CAE and OACAE.

\begin{table}[htbp]
\centering
\small
\caption{Average reconstruction metrics of POD, CAE, and OACAE on the test datasets for the cavity flow and dam flow problems.}
\label{tab:reconstruction-errors}
\begin{tabular}{lccc|ccc}
\toprule
 & \multicolumn{3}{c}{Cavity flow} 
 & \multicolumn{3}{c}{Dam flow} \\
\cmidrule(lr){2-4} \cmidrule(lr){5-7}
Model 
& REL~$\downarrow$ & MSE~$\downarrow$ & SSIM~$\uparrow$ 
& REL~$\downarrow$ & MSE~$\downarrow$ & SSIM~$\uparrow$ \\
\midrule
POD   & \textbf{0.002424} & \textbf{0.000006} & \textbf{0.987987} & \textbf{0.005586} & \textbf{0.000022} & \textbf{0.976817} \\
CAE   & 0.004584 & 0.000011 & 0.982691 & 0.007704 & 0.000054 & 0.957641 \\
OACAE & 0.008111 & 0.000031 & 0.969903 & 0.009966 & 0.000069 & 0.942009 \\
\bottomrule
\end{tabular}
\end{table}

Two main observations can be drawn from Table~\ref{tab:reconstruction-errors}.
First, the OACAE exhibits slightly lower reconstruction accuracy than the standard CAE. This behavior is expected, since the OACAE is not entirely trained for reconstruction but also incorporates an additional regression loss to align the latent space with physical parameters. As a result, part of the latent representation is allocated to parameter regression rather than purely minimizing reconstruction error, making the OACAE suboptimal for the reconstruction task alone. Second, unexpectedly, POD achieves the best reconstruction accuracy. This contrasts with existing studies on more complex flows, where CAE models usually outperform linear POD in capturing nonlinear structures. However, the cavity flow and dam flow dataset considered here is 
relatively less complex, as they are represented on a uniform $64\times64$ grid, which reduces the spatial resolution compared with higher-resolution datasets (e.g., $128\times128$ or $256\times256$).

Figure~\ref{fig:reconstruction-dam} in~\ref{appendix: Reconstruction snapshots} shows one test sample snapshot of the dam flow problem and cavity flow problem with the original field, the reconstructions from the three ROMs, and their error maps.

\subsection{Prediction}
\label{sec:prediction}

We now turn to the prediction task, where the goal is to directly infer velocity fields from physical parameters without relying on reference simulations. In this setting, the three ROMs are employed as surrogate models: POD-GPR, CAE-MLP, and OACAE-MLP.  

To quantitatively evaluate their predictive accuracy, we compute the average 
errors on the test set, summarized in Table~\ref{tab:prediction-errors}. 

\begin{table}[htbp]
\centering
\small
\caption{Average prediction metrics of different predictive models on the test dataset for the cavity flow and dam flow problems.}
\label{tab:prediction-errors}
\begin{tabular}{lccc|ccc}
\toprule
 & \multicolumn{3}{c}{Cavity flow} 
 & \multicolumn{3}{c}{Dam flow} \\
\cmidrule(lr){2-4} \cmidrule(lr){5-7}
Model 
& REL~$\downarrow$ & MSE~$\downarrow$ & SSIM~$\uparrow$ 
& REL~$\downarrow$ & MSE~$\downarrow$ & SSIM~$\uparrow$ \\
\midrule
POD-GPR   & 0.061342 & 0.002382 & 0.899857 & 0.020275 & 0.000109 & 0.944738 \\
CAE-MLP   & \textbf{0.014033} & \textbf{0.000079} & \textbf{0.952016} & \textbf{0.012304} & \textbf{0.000038} & \textbf{0.968740} \\
OACAE-MLP & 0.019615 & 0.000104 & 0.941041 & 0.014075 & 0.000042 & 0.952114 \\
\bottomrule
\end{tabular}
\end{table}

The same metrics as in the reconstruction study are reported: relative $\ell_2$ error (REL), 
mean square error (MSE), and structural similarity index (SSIM). 
While both reconstruction and prediction are evaluated on unseen test samples, the prediction task is inherently more difficult, as the models must infer the flow field entirely from physical parameters rather than from a latent encoding of the field itself.

Two main conclusions can be drawn from Table~\ref{tab:prediction-errors}.
First, although POD exhibits strong reconstruction capabilities in Section~\ref{sec:reconstruction}, its predictive extension (POD-GPR) performs markedly worse than the DL-based surrogates, CAE-MLP and OACAE-MLP, across all evaluation metrics, including the REL, MSE, and SSIM. This result highlights the superior expressiveness of deep learning ROMs for parameter-to-field prediction tasks and provides clear justification for their applications in more complex flow configurations. A second observation concerns the comparison between CAE-MLP and OACAE-MLP. CAE-MLP achieves superior reconstruction accuracy, as it is solely optimized for field reconstruction. In contrast, OACAE-MLP introduces an additional parameter regression constraint that promotes a more structured and physically informed latent representation. As a result, the performance gap between the two models is substantially reduced in the prediction task, indicating that the physics-aware latent space learned by OACAE-MLP enhances parameter-to-field generalization despite a slight reduction in pure reconstruction accuracy. 

 Figure~\ref{fig:prediction-dam-cavity} presents representative test cases from the dam flow and cavity flow datasets. The results indicate that DL-based surrogates not only outperform POD-GPR in terms of average performance but also exhibit robust accuracy on individual samples where POD-GPR fails to produce reliable predictions.

\begin{figure}[H]
\centering
\subfloat[truth u]{\includegraphics[width = 1in]{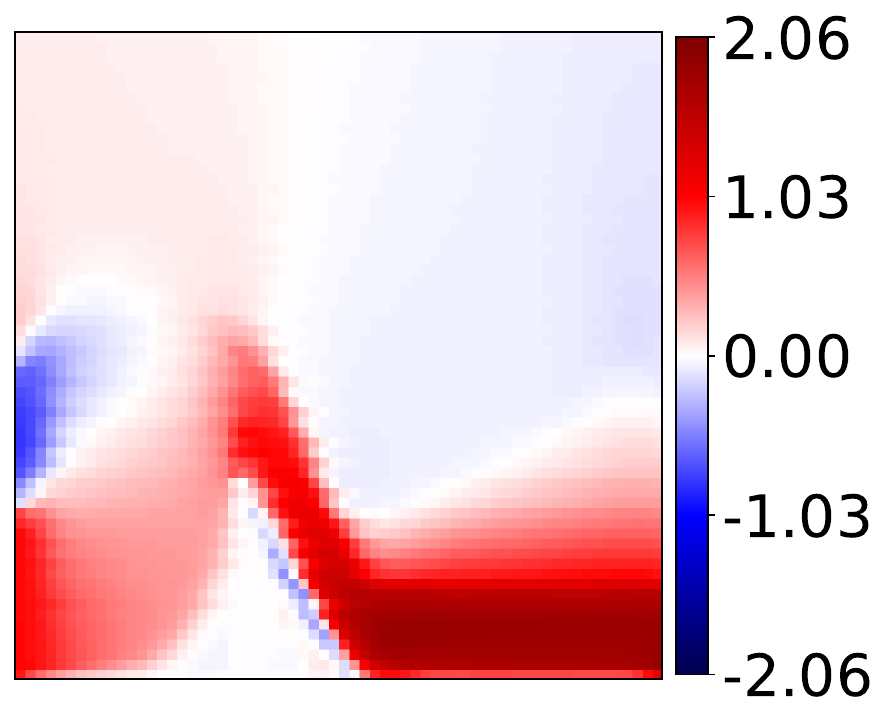}}
\qquad
\subfloat[no error]{\includegraphics[width = 1in]{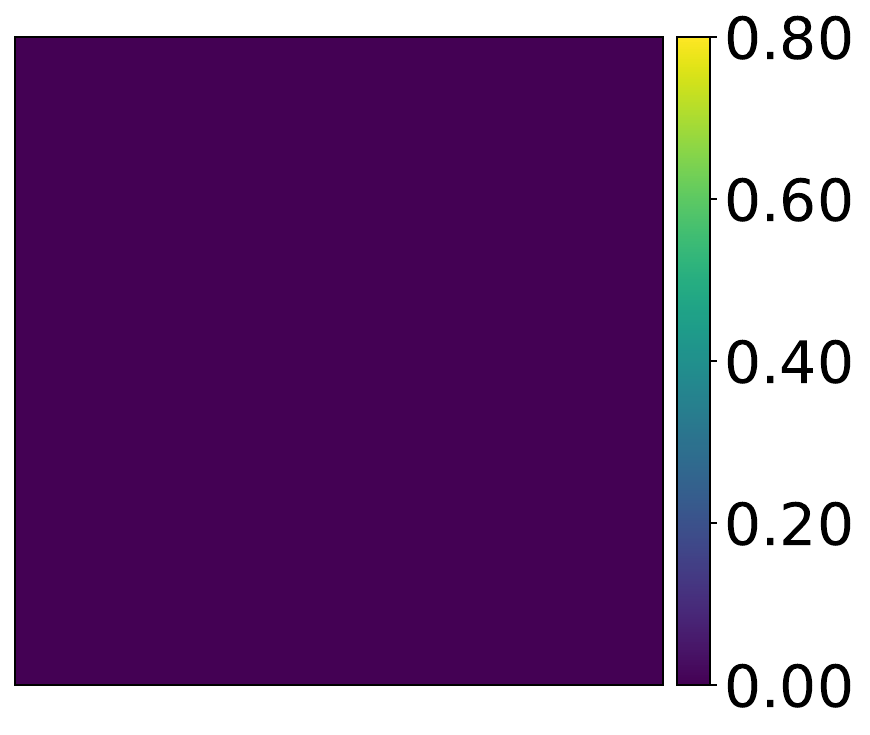}}
\qquad
\subfloat[truth v]{\includegraphics[width = 1in]{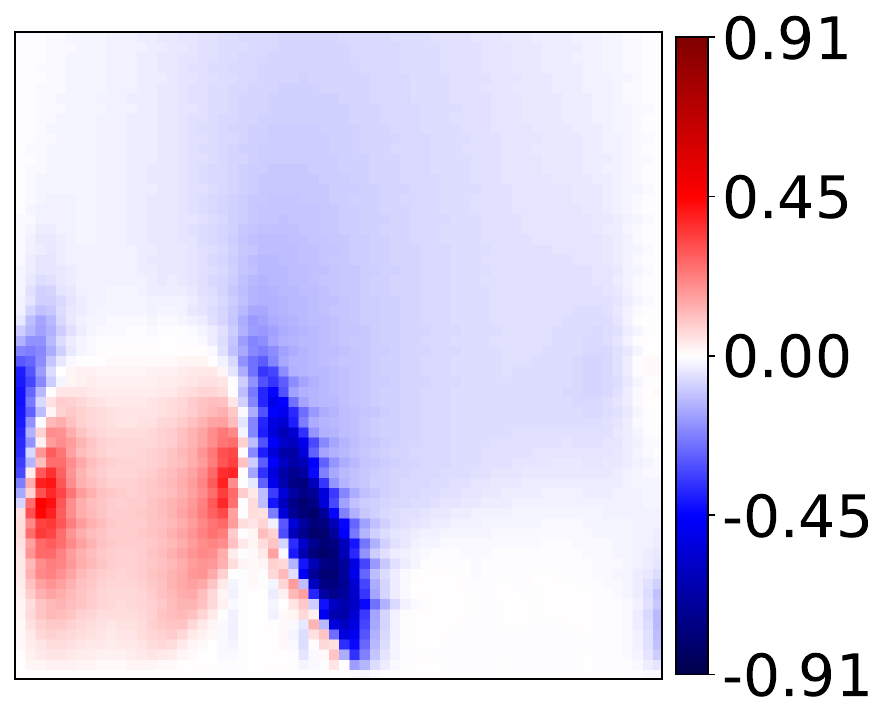}}
\qquad
\subfloat[no error]{\includegraphics[width = 1in]{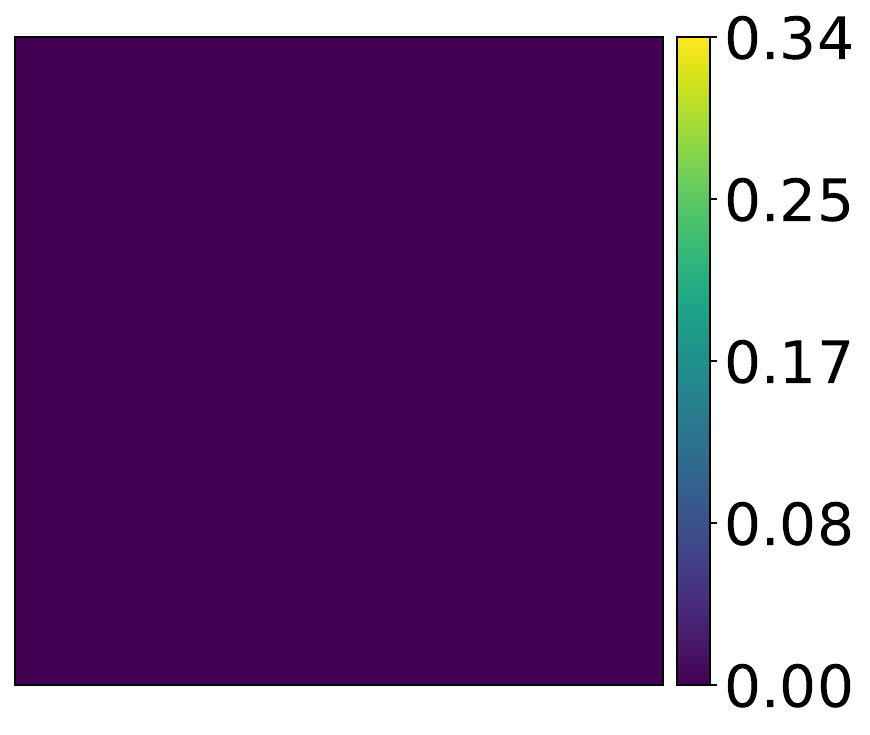}}\\
\subfloat[POD-GPR prediction]{\includegraphics[width = 1in]{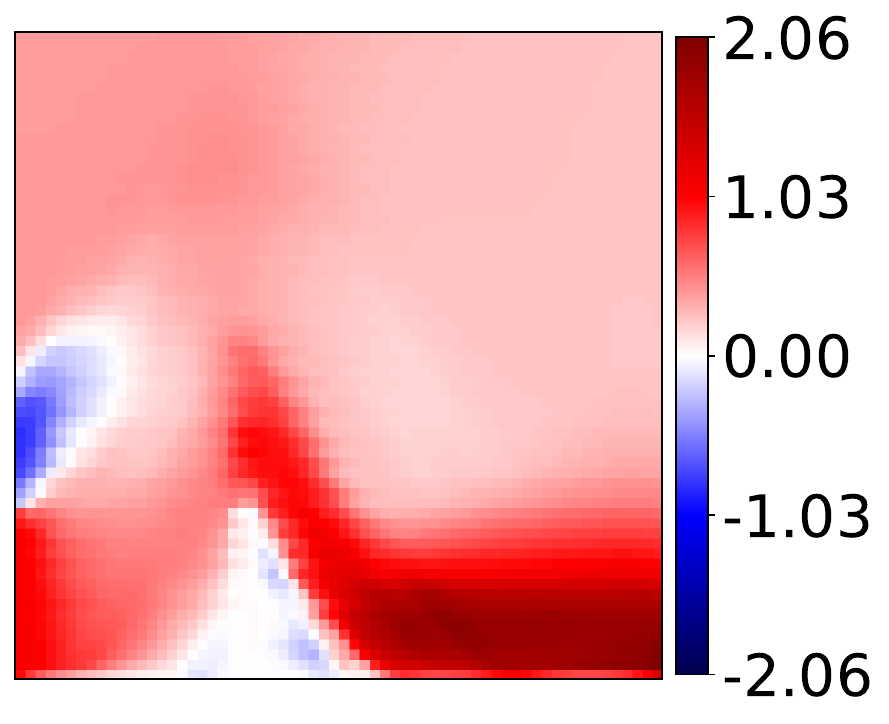}}
\qquad
\subfloat[error-u POD-GPR]{\includegraphics[width = 1in]{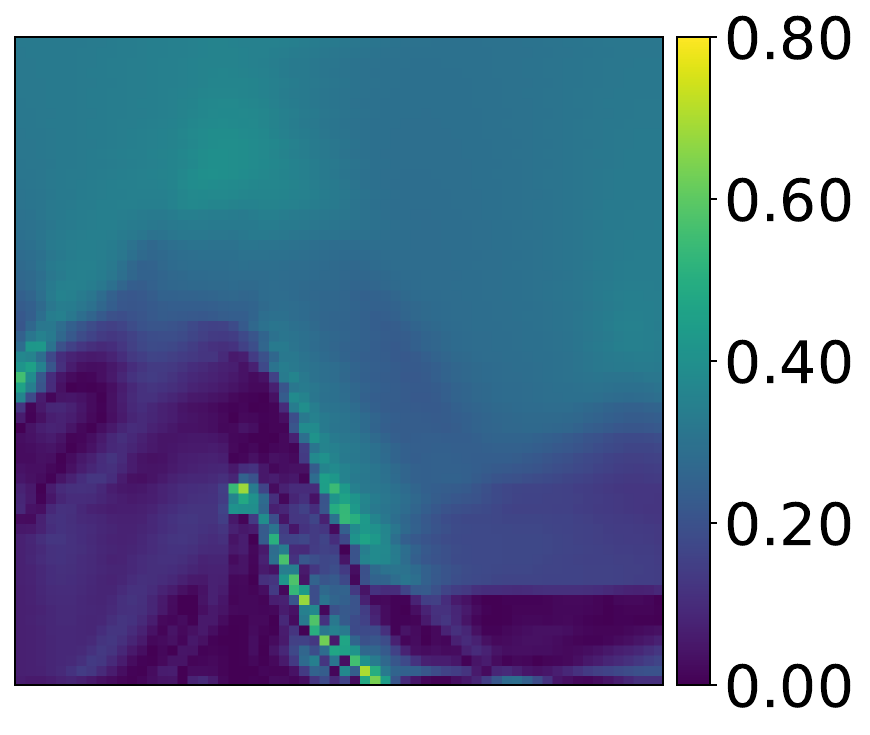}}
\qquad
\subfloat[POD-GPR prediction]{\includegraphics[width = 1in]{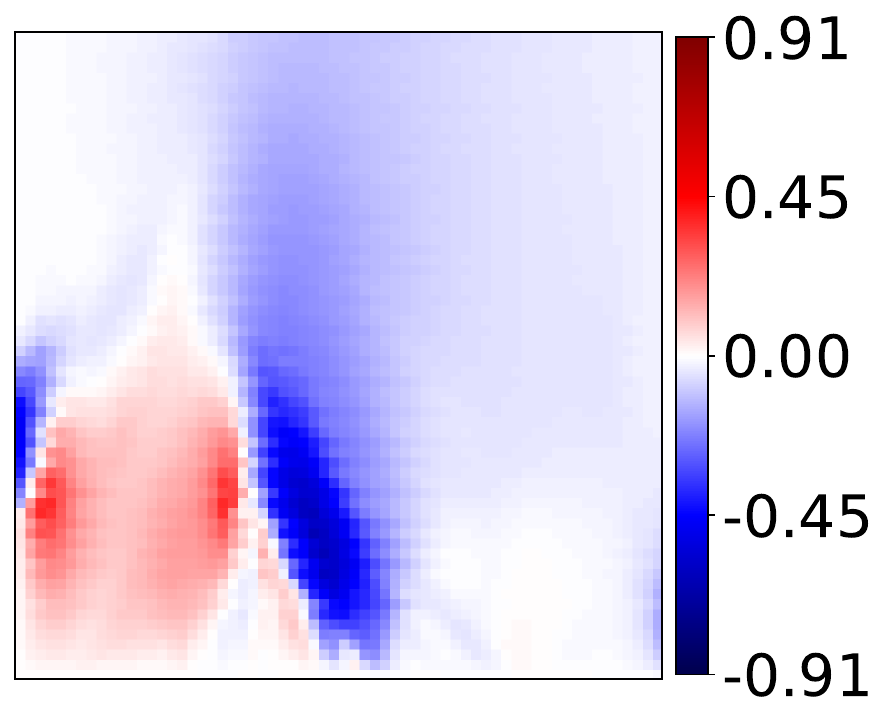}}
\qquad
\subfloat[error-v POD-GPR]{\includegraphics[width = 1in]{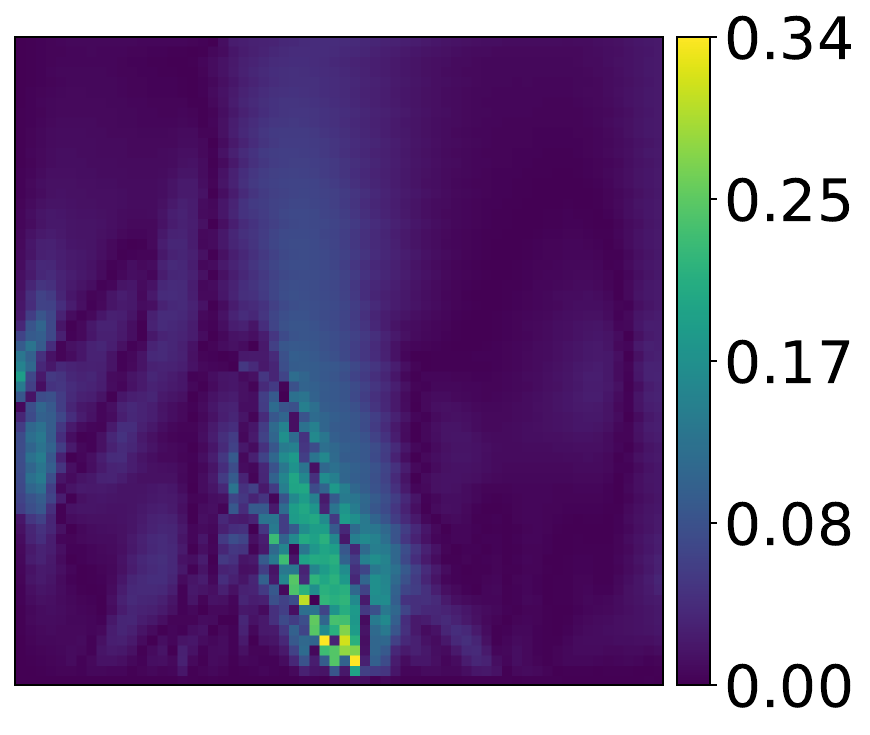}}\\
\subfloat[CAE-MLP prediction]{\includegraphics[width = 1in]{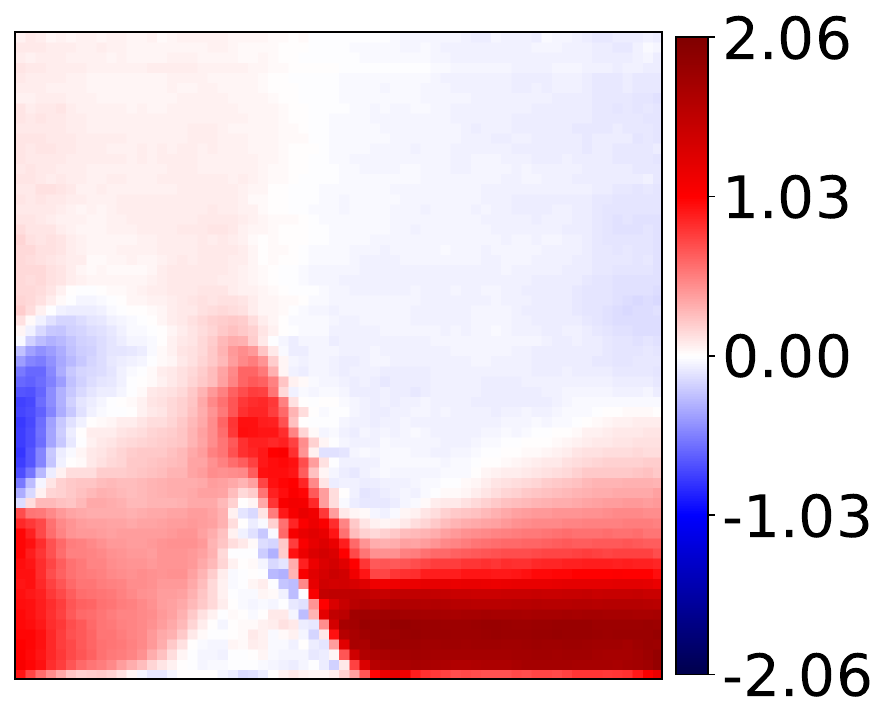}}
\qquad
\subfloat[error-u CAE-MLP]{\includegraphics[width = 1in]{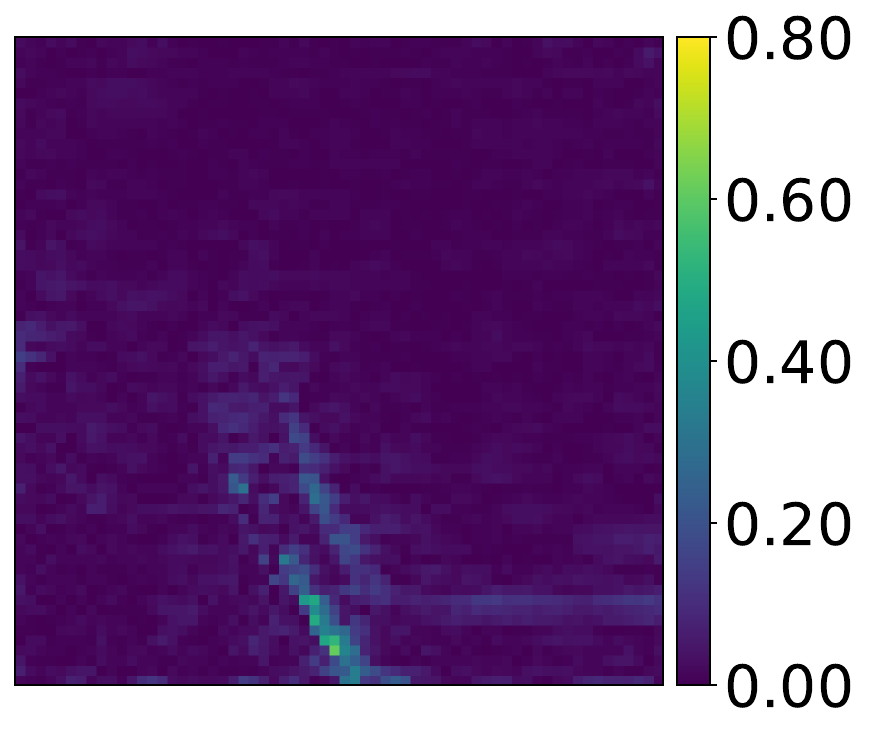}}
\qquad
\subfloat[CAE-MLP prediction]{\includegraphics[width = 1in]{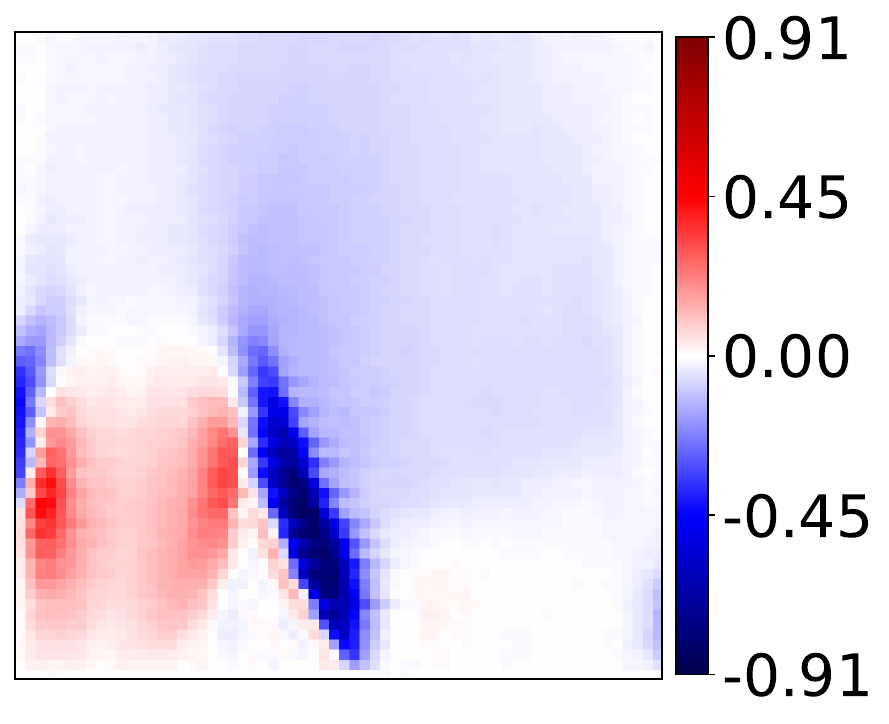}}
\qquad
\subfloat[error-v CAE-MLP]{\includegraphics[width = 1in]{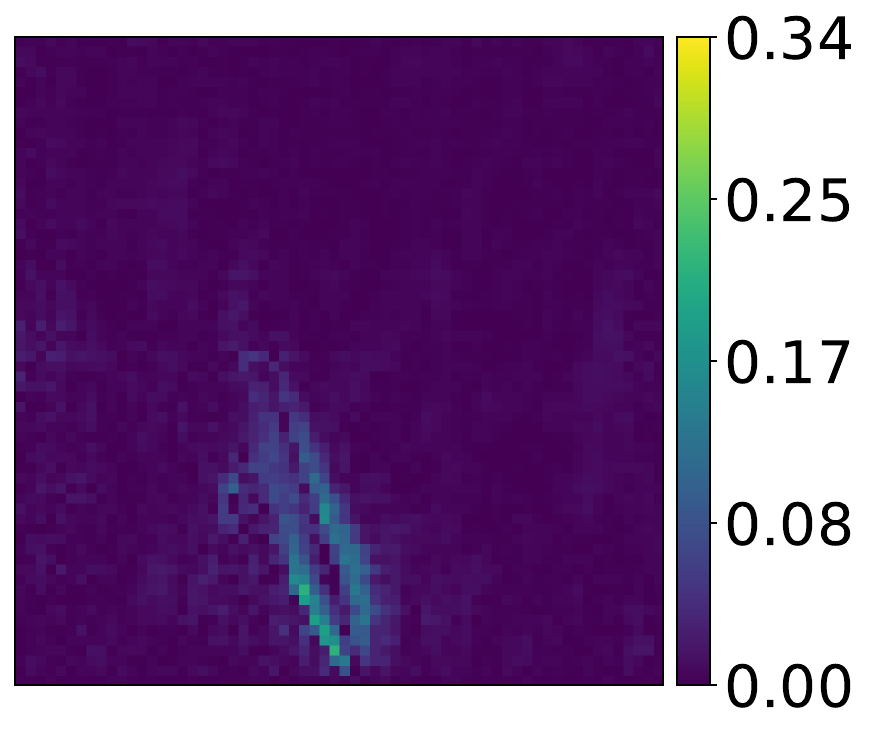}}\\
\subfloat[OACAE-MLP prediction]{\includegraphics[width = 1in]{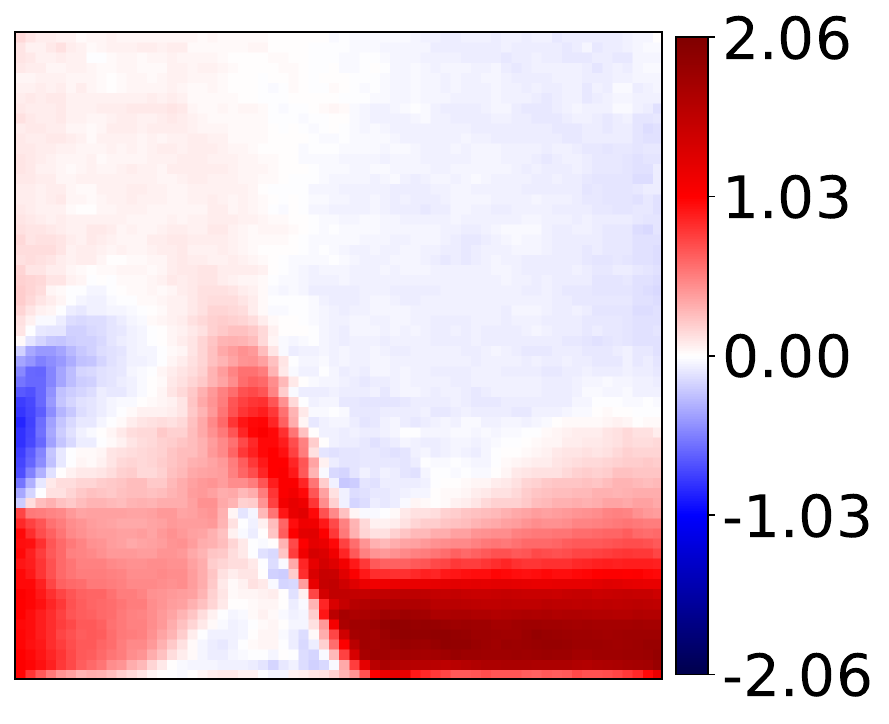}}
\qquad
\subfloat[error-u OACAE-MLP]{\includegraphics[width = 1in]{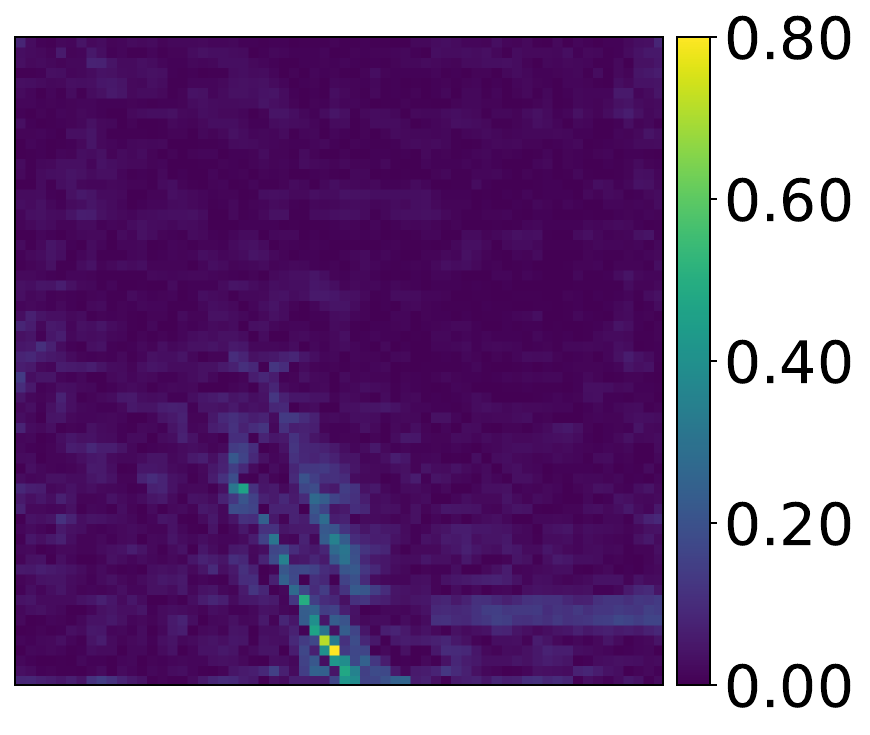}}
\qquad
\subfloat[OACAE-MLP prediction]{\includegraphics[width = 1in]{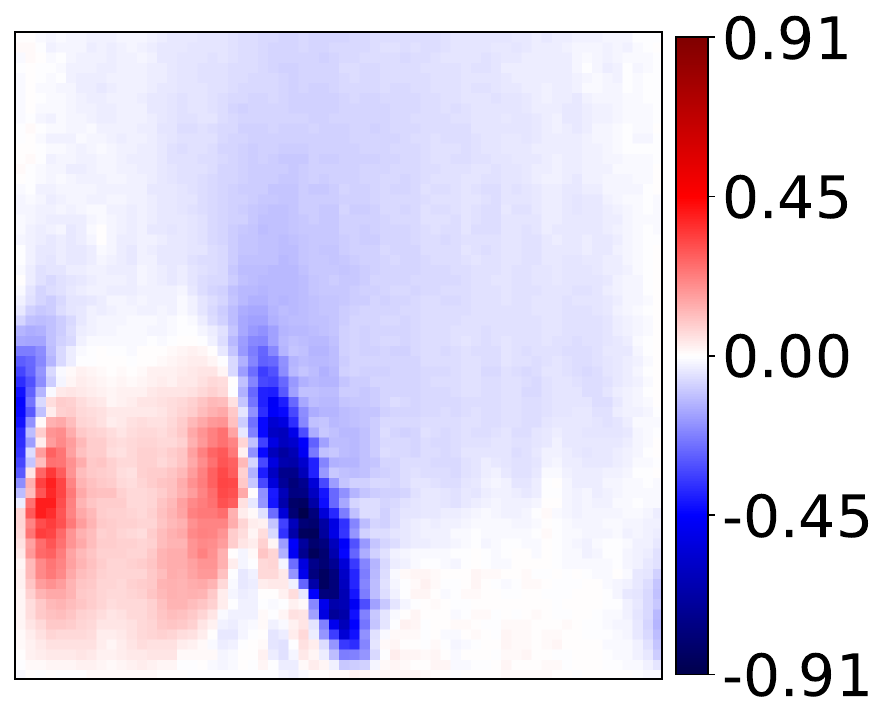}}
\qquad
\subfloat[error-v OACAE-MLP]{\includegraphics[width = 1in]{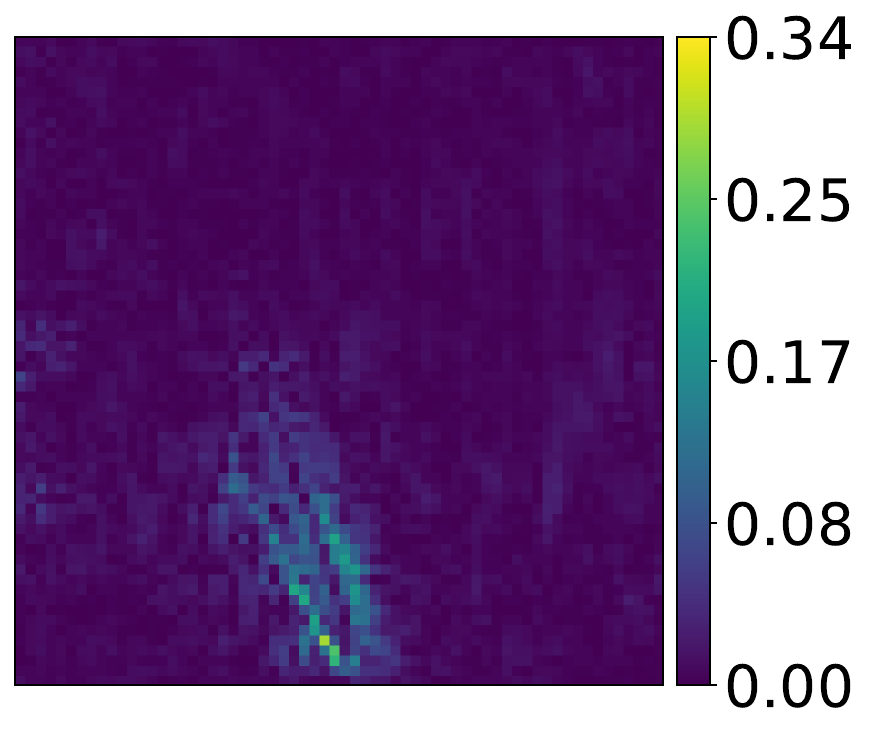}}\\
\end{figure}

\addtocounter{figure}{-1}

\begin{figure}[H]
\centering
\subfloat[truth u]{\includegraphics[width = 1in]{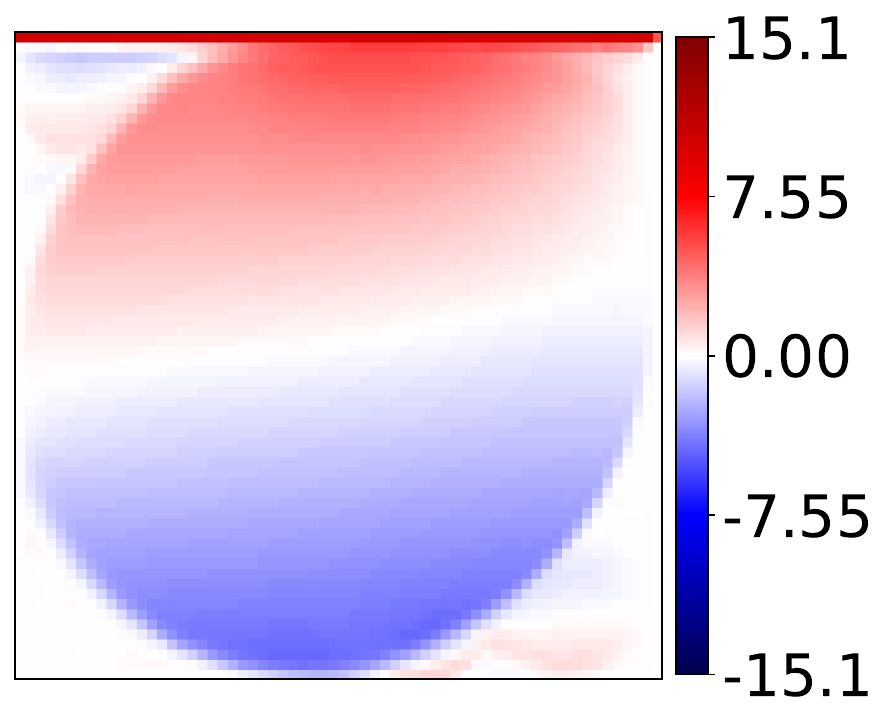}}
\qquad
\subfloat[no error]{\includegraphics[width = 1in]{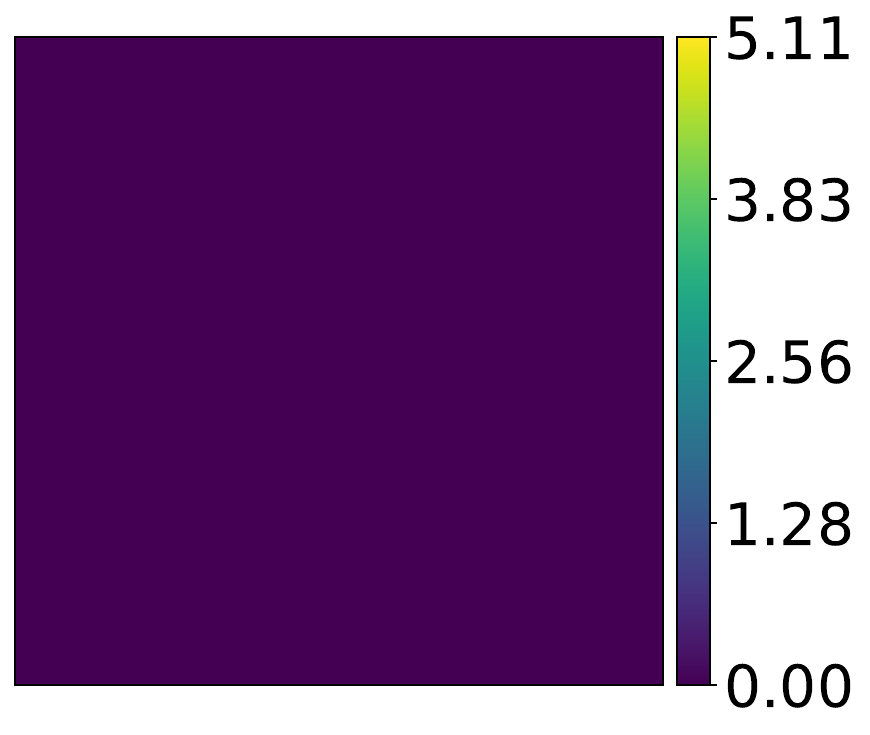}}
\qquad
\subfloat[truth v]{\includegraphics[width = 1in]{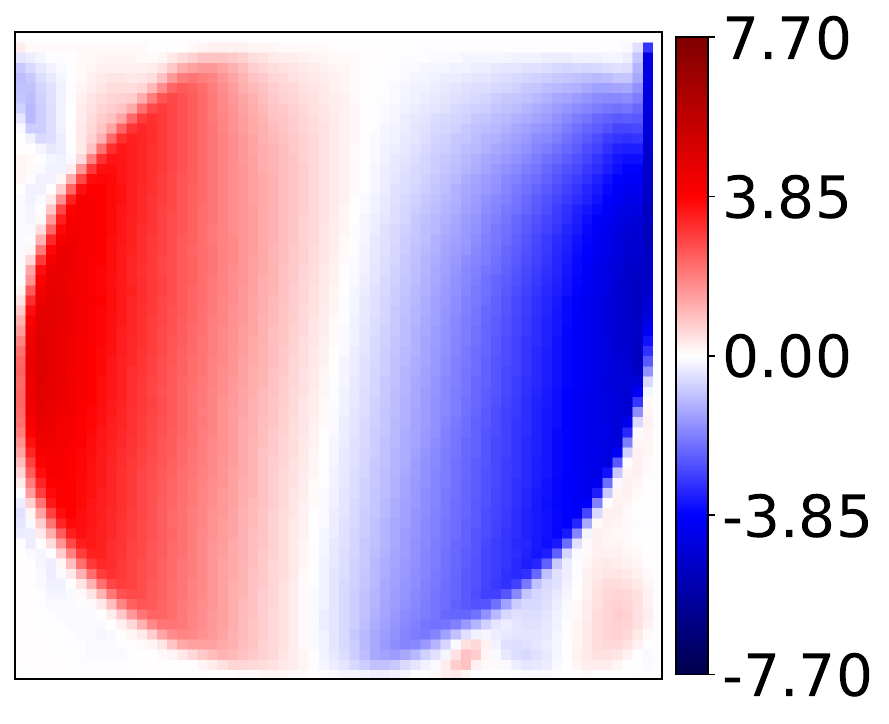}}
\qquad
\subfloat[no error]{\includegraphics[width = 1in]{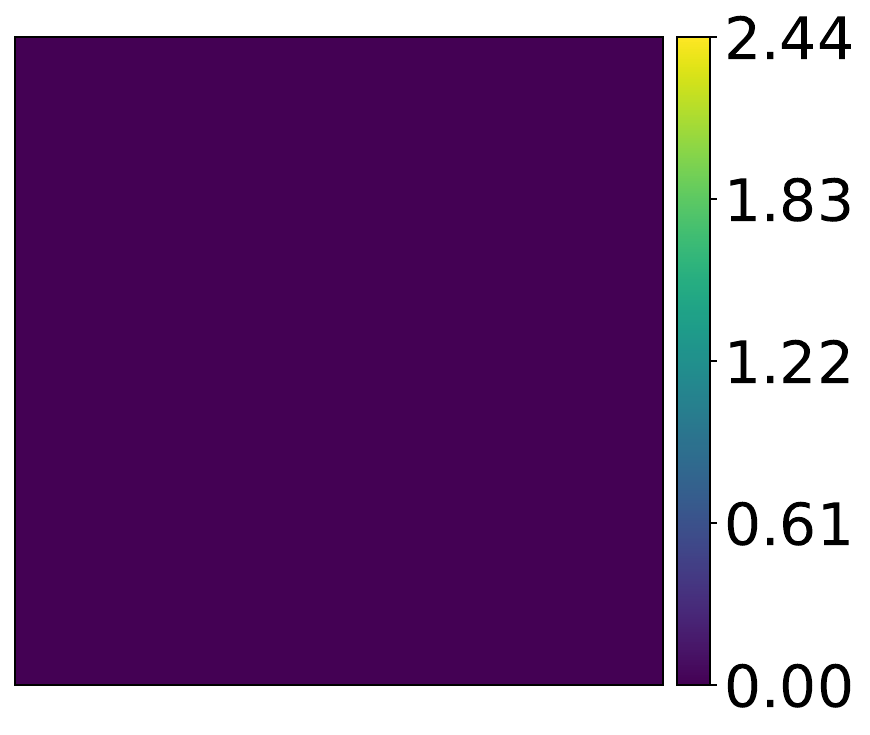}}\\
\subfloat[POD-GPR prediction]{\includegraphics[width = 1in]{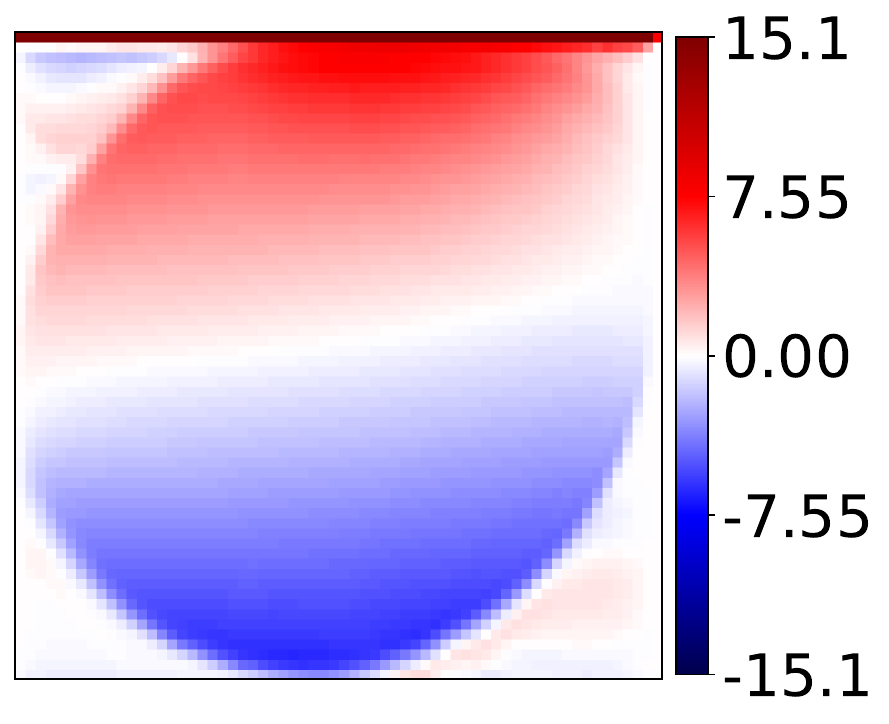}}
\qquad
\subfloat[error u POD-GPR]{\includegraphics[width = 1in]{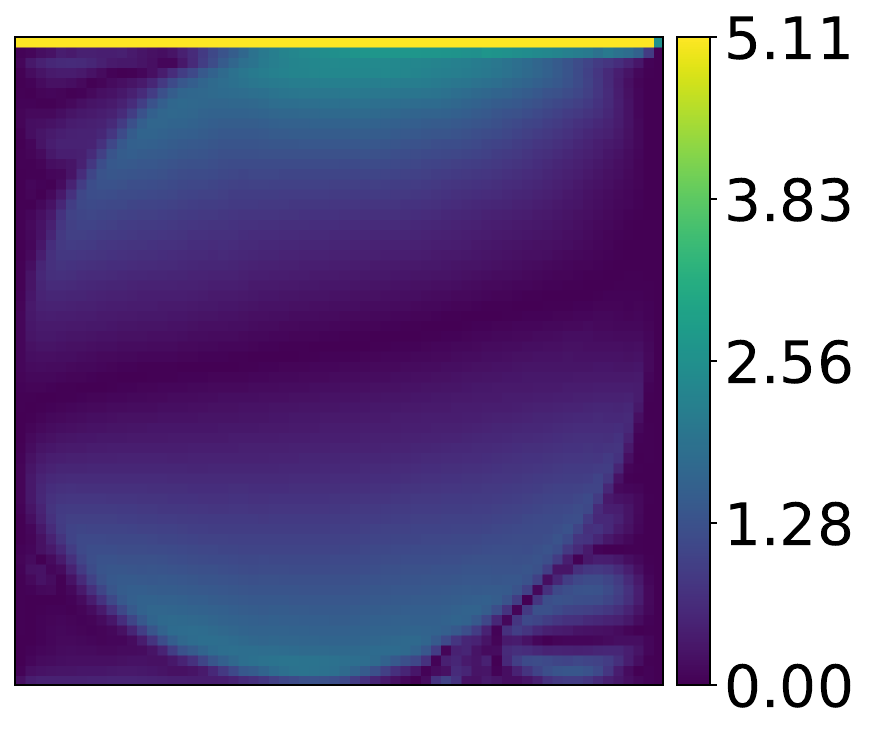}}
\qquad
\subfloat[POD-GPR prediction]{\includegraphics[width = 1in]{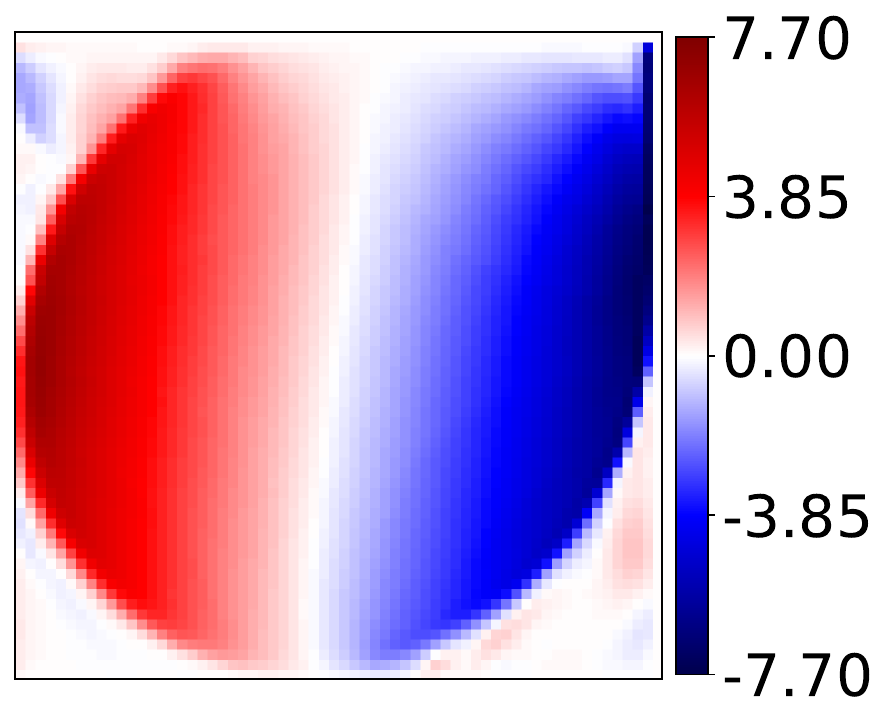}}
\qquad
\subfloat[error v POD-GPR]{\includegraphics[width = 1in]{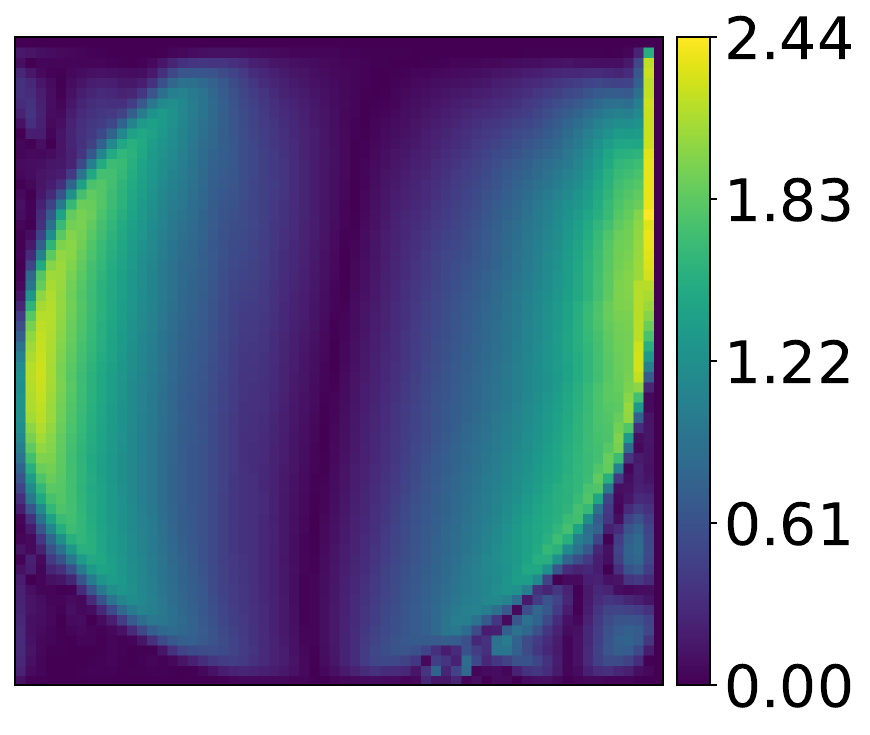}}\\
\subfloat[CAE-MLP prediction]{\includegraphics[width = 1in]{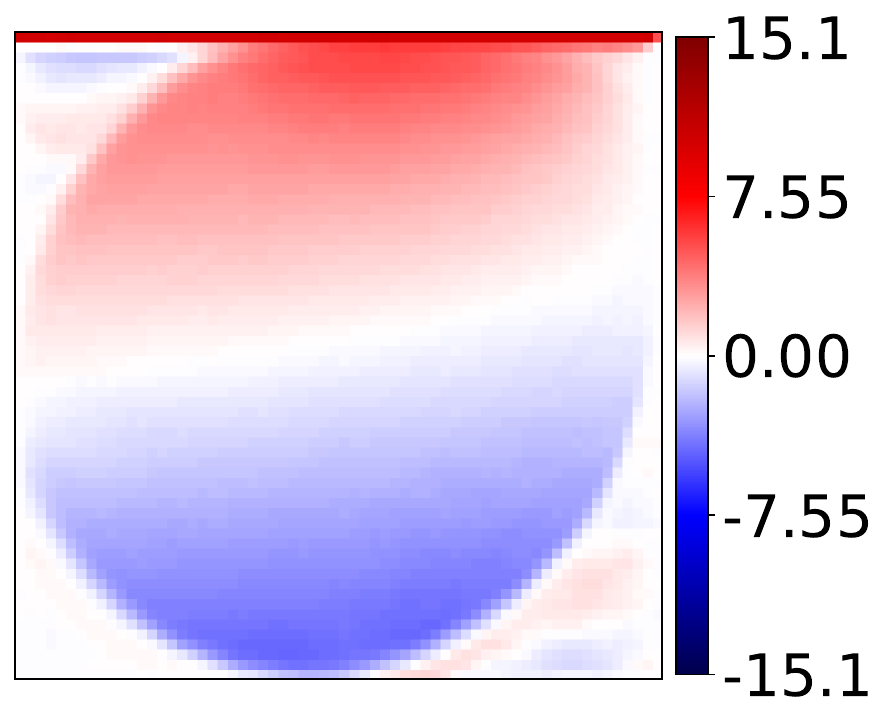}}
\qquad
\subfloat[error u CAE-MLP]{\includegraphics[width = 1in]{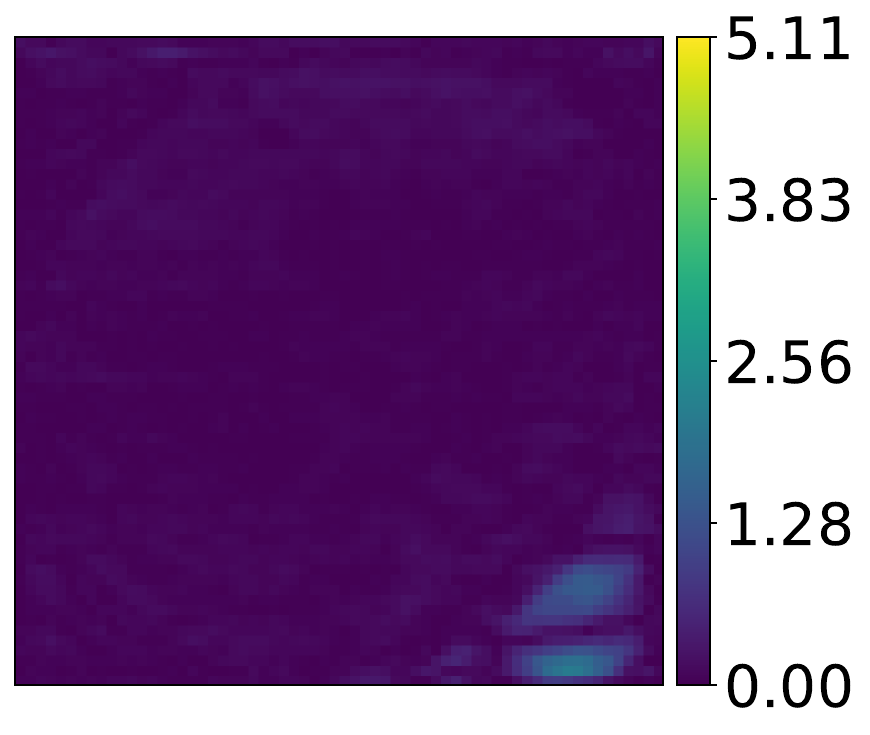}}
\qquad
\subfloat[CAE-MLP prediction]{\includegraphics[width = 1in]{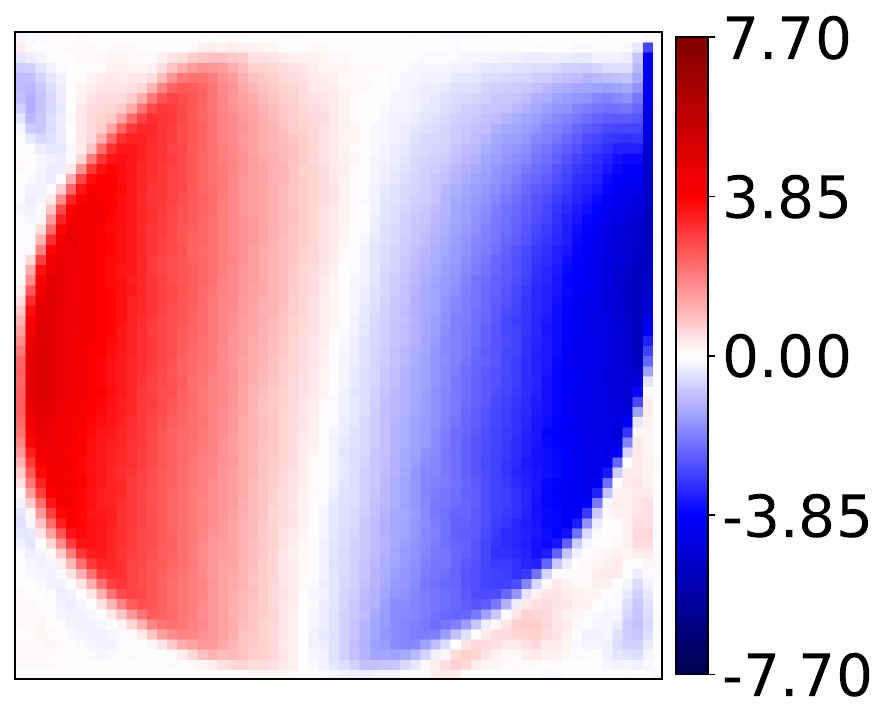}}
\qquad
\subfloat[error v CAE-MLP]{\includegraphics[width = 1in]{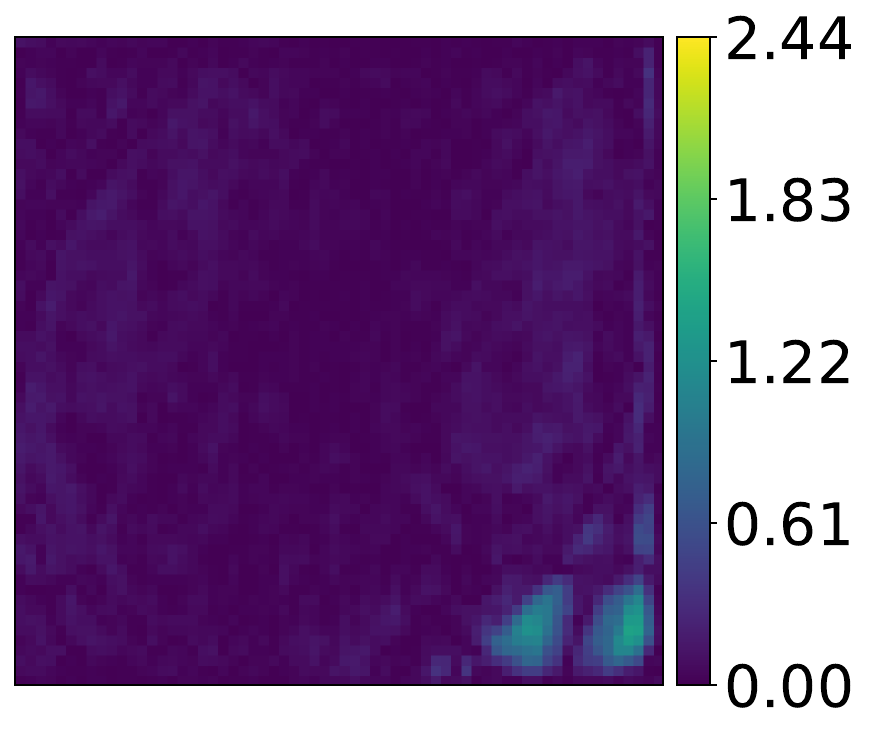}}\\
\subfloat[OACAE-MLP prediction]{\includegraphics[width = 1in]{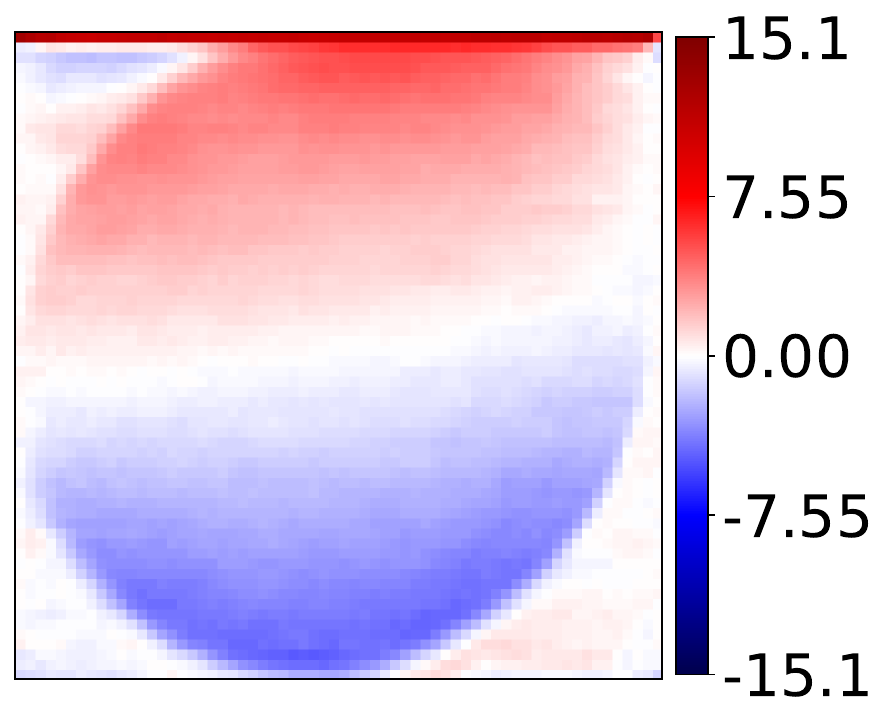}}
\qquad
\subfloat[error u OACAE-MLP]{\includegraphics[width = 1in]{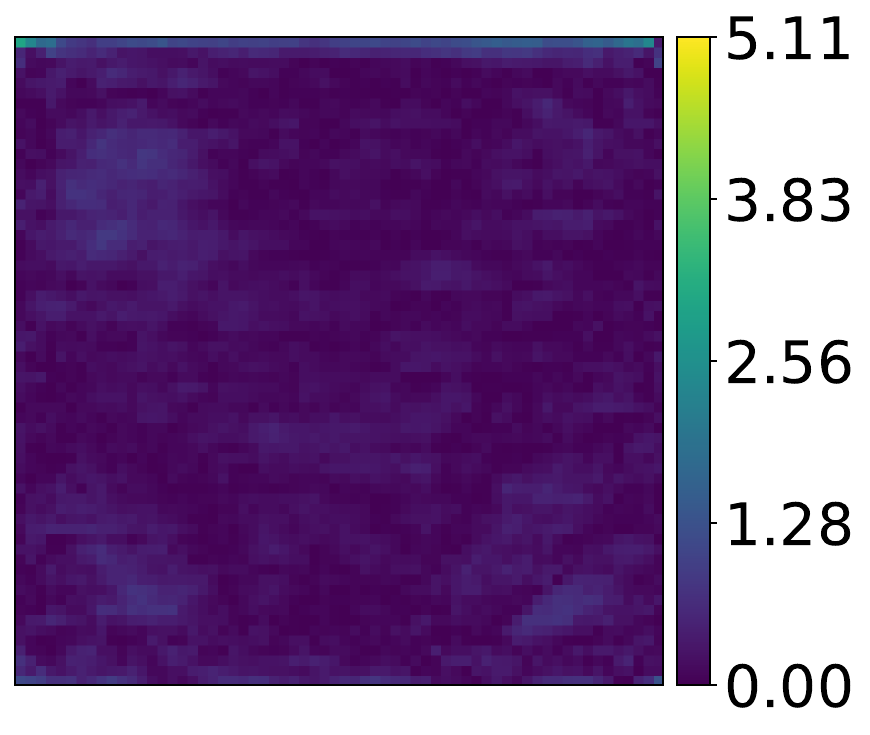}}
\qquad
\subfloat[OACAE-MLP prediction]{\includegraphics[width = 1in]{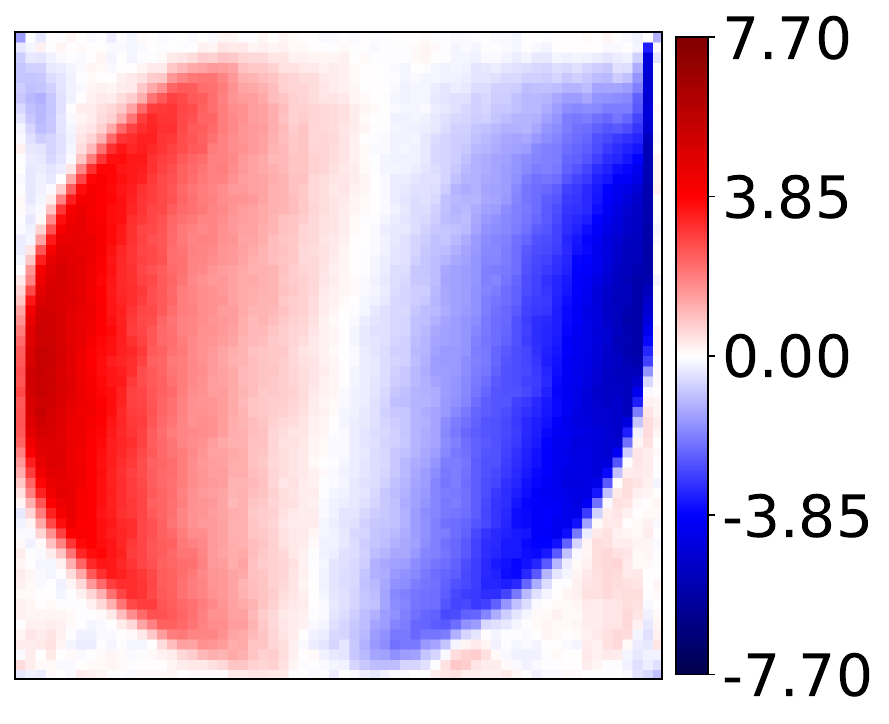}}
\qquad
\subfloat[error v OACAE-MLP]{\includegraphics[width = 1in]{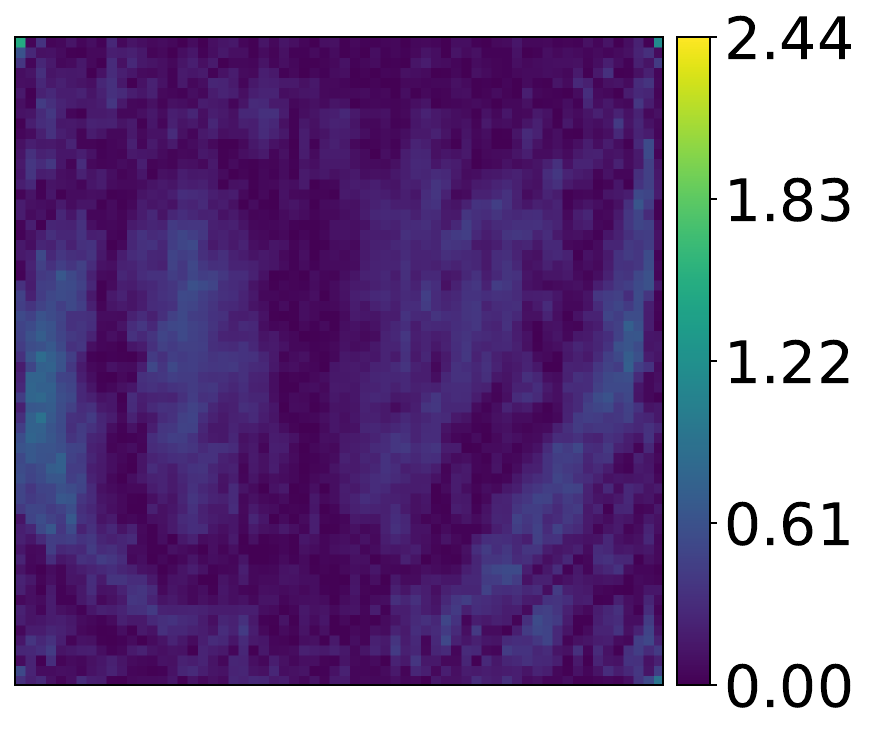}}\\
   \caption{Comparison of prediction results (dam flow and cavity flow)}
   \label{fig:prediction-dam-cavity}
\end{figure}

\subsection{Inverse modeling: Parameter calibration}
\label{sec:calibration}

We finally consider the parameter calibration problem, which aims to estimate key physical parameters from observed flow fields.
For the cavity flow, the lid velocity $u_{lid}$ and fluid density $\rho$ are treated as control variables, while for the dam-break flow, the barrier height $h$ and inlet velocity $u_{in}$ are considered.
All remaining parameters are fixed to their ground-truth values.

Parameter calibration is performed using variational DA frameworks (3D-Var and 4D-Var) combined with CAE-MLP and OACAE-MLP surrogates.
As a baseline, an Ensemble Kalman Filter (EnKF) coupled with a POD-GPR surrogate is also evaluated.
All experiments are conducted on held-out test cases.

For each flow configuration, a representative temporal evolution is selected.
We first examine single-time-step assimilation strategies, including EnKF with POD-GPR and 3D-Var with CAE-MLP and OACAE-MLP, applied over the entire evolution.
We then investigate multi-time-step assimilation strategies, namely 2T-EnKF with POD-GPR and 4D-Var with CAE-MLP and OACAE-MLP, using multiple time windows extracted from the same evolution. 
Since multi-time-step assimilation increases the number of surrogate evaluations, we further compare the online runtime of the EnKF-POD-GPR and the AE-MLP 4D-Var frameworks in~\ref{app: Computational efficiency}.

\subsubsection{Single-time-step assimilation.}
\label{sec: single-da}
We first examine the performance of single-time-step assimilation frameworks on one dam flow and cavity flow evolution case. The dam flow evolution case consists of 100 time steps, while the cavity flow evolution case includes 32 time steps. The availability of complete temporal evolutions with a sufficient number of time instants allows for a reliable evaluation of single-time-step parameter estimation performance. Table~\ref{tab:calibration-1T} reports the average parameter estimation errors, with the background errors included for comparison. Since all data assimilation procedures are initialized from the same prescribed background state, the comparison between analysis and background errors provides a direct measure of the correction capability of each method. The considered approaches include the one-step EnKF (1T-EnKF) coupled with POD-GPR, as well as 3D-Var combined with CAE-MLP and OACAE-MLP surrogates.

\begin{table}[htbp]
\centering
\caption{Average $\ell_2$ relative errors of parameters for single-time-step assimilation. Both the analysis results and the background are reported for reference.}
\label{tab:calibration-1T}
\begin{tabular}{lcc|cc}
\toprule
 & \multicolumn{2}{c}{Dam flow} & \multicolumn{2}{c}{Cavity flow} \\
\cmidrule(lr){2-3} \cmidrule(lr){4-5}
Method 
& $u_{in}$  & $h$ 
& $u_{lid}$ & $\rho$\\
\midrule
\textit{Background}
& 0.097873 & 0.632951 & 0.951456 & 9.325386\\
1T-EnKF (POD--GPR) 
& 0.144884 & 0.298331 & 0.563508 & 3.322671\\
3D-Var (CAE--MLP) 
& \textbf{0.054595} & 0.167084  & 0.562477 & \textbf{2.096463}\\
3D-Var (OACAE--MLP) 
& 0.068964 & \textbf{0.095677} & \textbf{0.159565} & 2.533160\\
\bottomrule
\end{tabular}
\end{table}

Several observations can be drawn from these results.
First, in both flow configurations, the lowest estimation errors are consistently achieved by 3D-Var combined with DL-based ROM frameworks.
For the dam flow problem, the inlet velocity $u_{in}$ is most accurately estimated using the 3D-Var-CAE-MLP framework, whereas the barrier height $h$ is better recovered with 3D-Var-OACAE-MLP.
In the cavity flow case, the most accurate estimates of the lid velocity $u_{lid}$ are obtained with 3D-Var-OACAE-MLP, while the fluid density $\rho$ is more accurately estimated using 3D-Var-CAE-MLP.
These results demonstrate the effectiveness of the proposed variational DA-DL-ROM framework for parameter calibration and its advantage over the EnKF-POD-GPR baseline in the same task. 
Second, although the 3D-Var-OACAE-MLP framework exhibits slightly higher estimation errors for one of the parameters in each configuration (namely $u_{in}$ for the dam flow and $\rho$ for the cavity flow), it provides substantially improved estimates for the other parameter ($h$ and $u_{lid}$, respectively).
Since the objective of the data assimilation procedure is to optimize the control variable vector, lower estimation errors in the analyzed parameter vector directly reflect improved calibration performance. These results suggest that the physics-informed latent representation learned by OACAE can enhance parameter calibration in the proposed variational-DA-DL-ROM framework.

To further investigate different inverse modeling performances between the CAE-MLP and OACAE-MLP, beyond the average parameter estimation errors reported in Table~\ref{tab:calibration-1T}, it is also essential to assess the impact of parameter calibration in the physical space. In particular, we compare the prediction errors of the flow fields obtained using the true parameters and the calibrated (analysis) parameters. To this end, we recall that the 3D-Var cost function~\eqref{eq:3dvar costfunc} consists of a prior term and an observation term. In the present setting, the observations correspond to the true physical fields, such that the observation term directly measures the discrepancy between the predicted and true flow fields. Moreover, because the observation term is given a large weight, its contribution dominates the cost function by several orders of magnitude over the prior term. 
\begin{figure}[htbp]
\centering
  \begin{subfigure}[t]{0.46\textwidth}
    \centering
    \includegraphics[width=\textwidth]{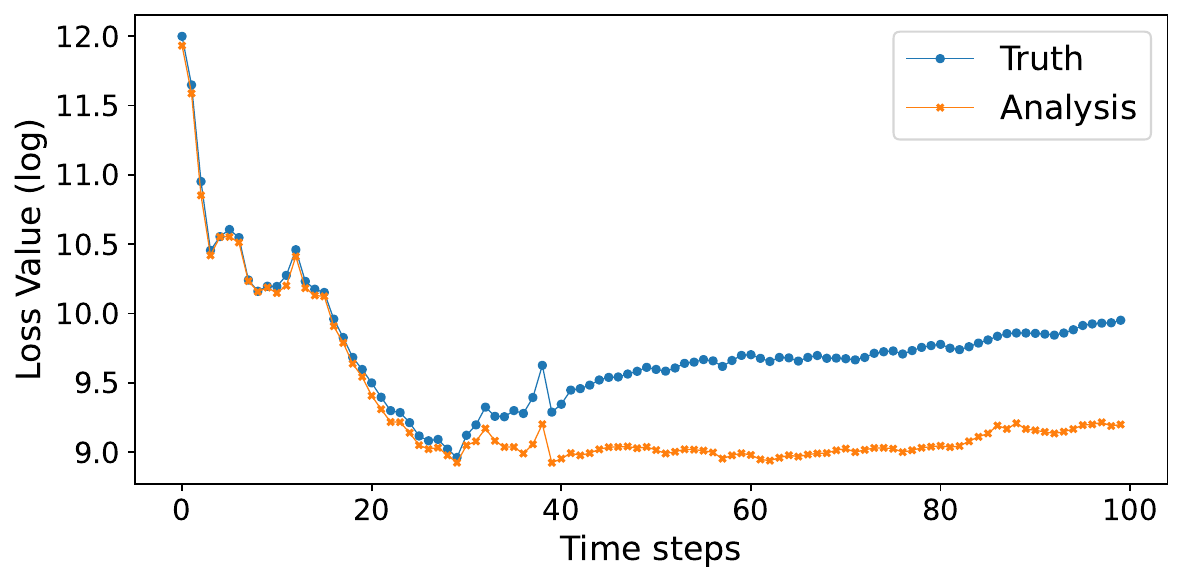}
    \caption{Dam flow --- OACAE-MLP}
  \end{subfigure}
  \quad\quad
  \begin{subfigure}[t]{0.46\textwidth}
    \centering
    \includegraphics[width=\textwidth]{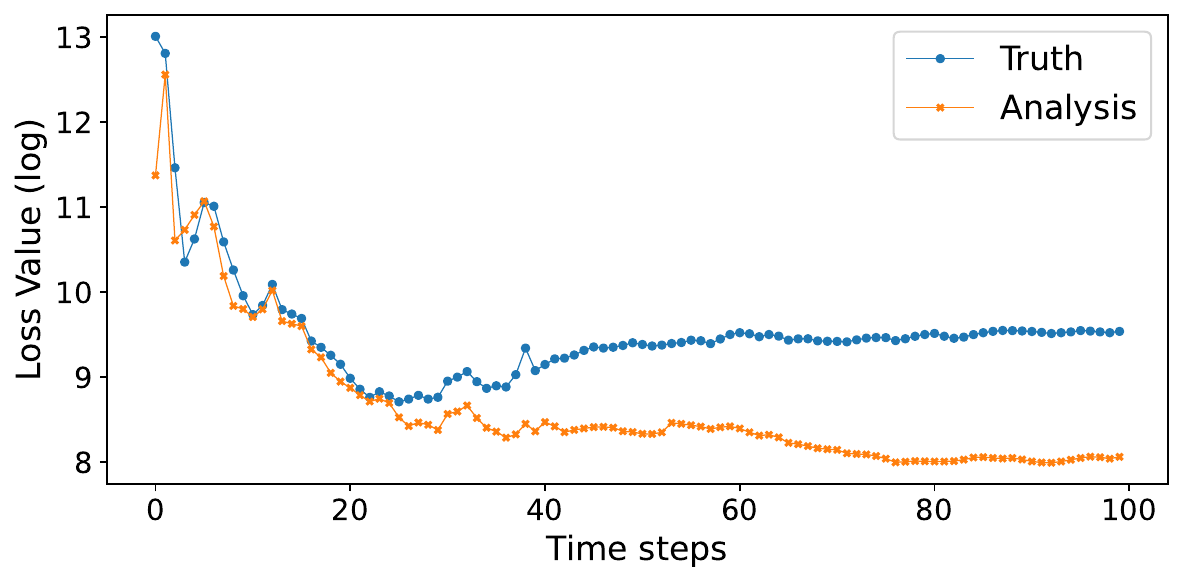}
    \caption{Dam flow --- CAE-MLP}
  \end{subfigure}

  \vspace{1em}

  \begin{subfigure}[t]{0.46\textwidth}
    \centering
    \includegraphics[width=\textwidth]{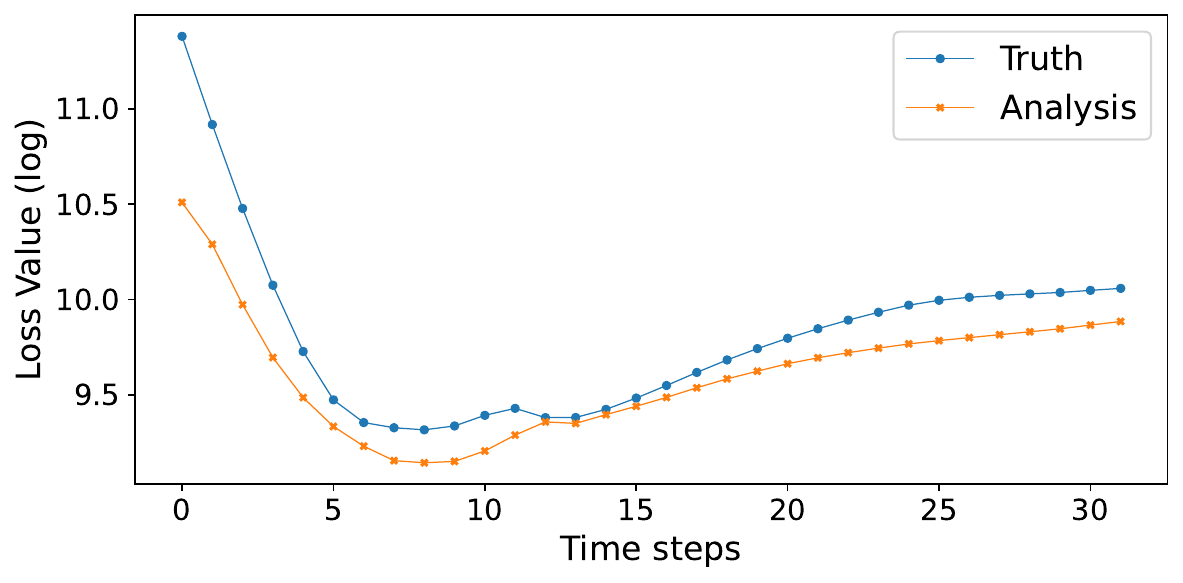}
    \caption{Cavity flow --- OACAE-MLP}
  \end{subfigure}
  \quad\quad
  \begin{subfigure}[t]{0.46\textwidth}
    \centering
    \includegraphics[width=\textwidth]{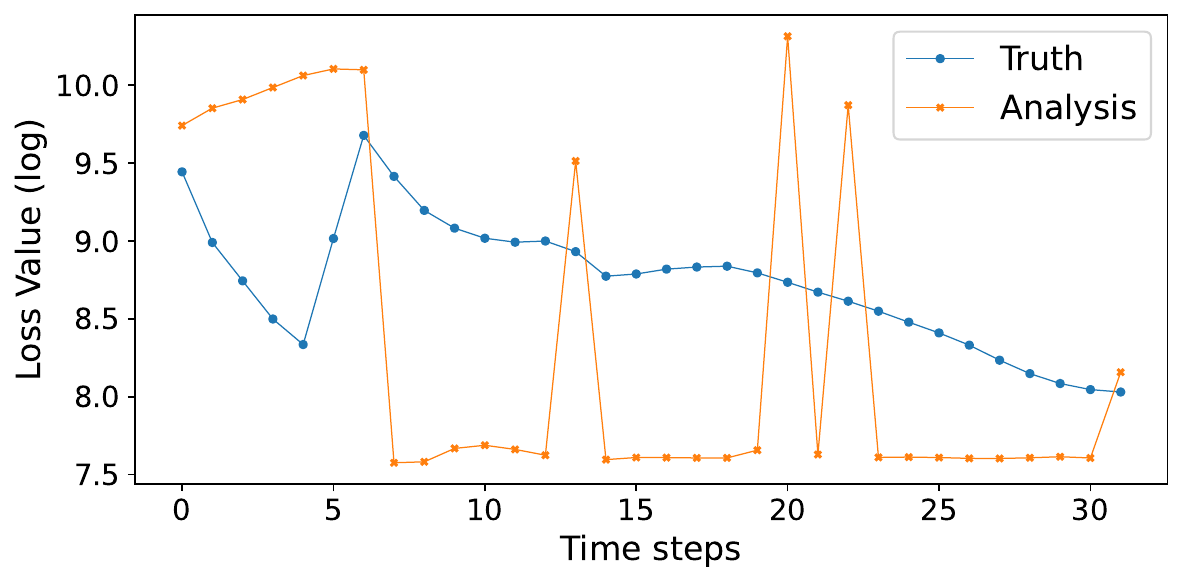}
    \caption{Cavity flow --- CAE-MLP}
  \end{subfigure}

\caption{Comparison of 3D-Var observation mismatch in the physical space, evaluated at each time step of the selected evolution case, using the true parameters and the calibrated (analysis) parameters for different surrogate models (OACAE-MLP and CAE-MLP) and flow configurations (dam flow and cavity flow).}
\label{fig:physics-loss-1T}
\end{figure}
As a consequence, the effectiveness of variational data assimilation can be assessed through the 3D-Var validation loss, defined as the value of the observation term in the 3D-Var cost function, which provides a physically meaningful criterion for comparing different parameter estimates.
Since the observation operator is a learned surrogate rather than the full-order model, the surrogate cannot exactly represent all spatial structures of the full-order flow. The inverse optimization may therefore adjust the control parameters to compensate for unresolved or distorted information in the reduced-order model. Consequently, the true physical parameters do not necessarily minimize the surrogate-induced observation mismatch.
In particular, obtaining a lower validation loss with the analysis parameters than with the true parameters at a given time step serves as a necessary consistency check for the optimization procedure, indicating improved flow predictions within the surrogate-based framework. Figure~\ref{fig:physics-loss-1T} illustrates the temporal evolution of the validation loss in the physical space for both flow configurations, comparing the predictions obtained using the true and calibrated parameters. For the OACAE-MLP framework, the validation loss associated with the analysis parameters is consistently lower than or equal to that obtained with the true parameters at all time steps, indicating robust convergence of the variational assimilation procedure. By contrast, in the CAE-MLP framework, although similar trends are observed at most time steps, several instances exhibit higher validation losses with the analysis parameters, suggesting a lack of convergence at specific time steps and less reliable flow predictions, which can be attributed to a less smooth latent space. Overall, these results further demonstrate the effectiveness and enhanced robustness of the proposed physics-informed variational DA-DL-ROM framework for parameter calibration.

\subsubsection{Multi-time-step assimilation.} 
\begin{figure}[htbp]
    \centering
    \begin{subfigure}[t]{0.46\textwidth}
        \centering
        \includegraphics[width=\textwidth]{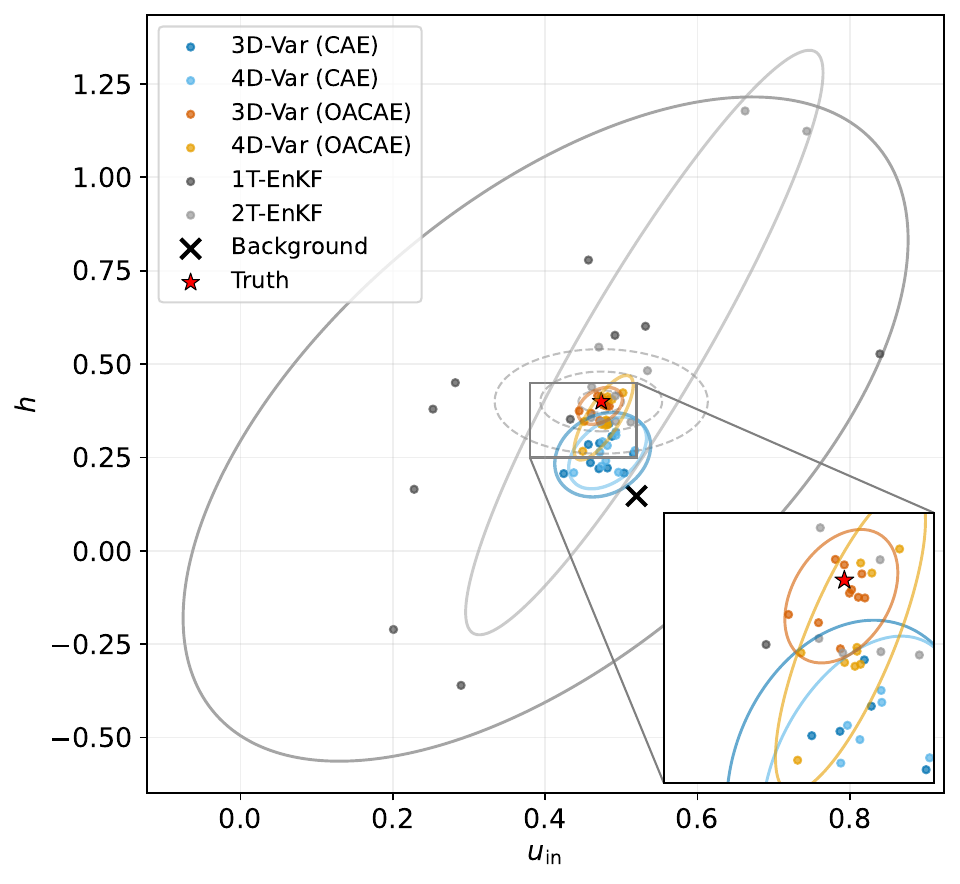}
        \caption{Inverse modeling results in parameter space (Dam flow)}
        \label{fig:ellipse-single}
    \end{subfigure}
    \hfill
    \begin{subfigure}[t]{0.46\textwidth}
        \centering
        \includegraphics[width=\textwidth]{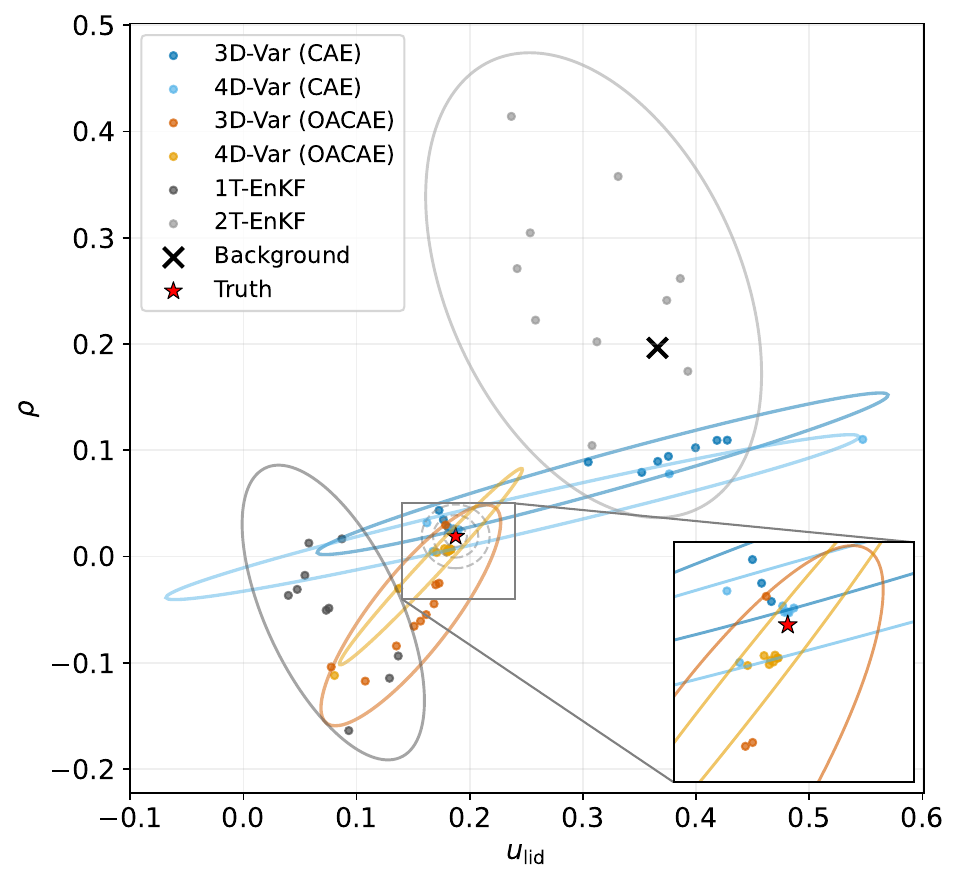}
        \caption{Inverse modeling results in parameter space (Cavity flow)}
        \label{fig:ellipse-multi}
    \end{subfigure}
    \caption{Distribution of calibrated parameters (ten realizations for each method) for the dam flow and cavity flow problems. Scatter markers indicate the estimated $(u_{\mathrm{in}}, h)$ and $(u_{\mathrm{lid}}, \rho)$ parameter pairs obtained using different data assimilation frameworks. The black cross denotes the background state, while the star indicates the ground-truth parameters. Ellipses represent the $2\sigma$ covariance bounds associated with each method.
    Overall, the variational DA-DL-ROM frameworks yield more concentrated parameter estimates around the target than the baseline approaches. In particular, the 3D-Var-OACAE-MLP and 4D-Var-OACAE-MLP strategies exhibit the tightest distributions around the reference values for both flow configurations.}
    \label{fig:calibration-ellipse}
\end{figure}

We now examine the performance of multi-time-step assimilation frameworks.
To ensure a consistent basis for comparison with the single-time-step experiments, we consider the same evolution cases from the dam flow and cavity flow datasets. Specifically, we design ten comparative experiments between the two-time-step EnKF (2T-EnKF) and the 4D-Var strategies.
In each experiment, one snapshot selected from $\mathbf{x}_{t=0}$ to $\mathbf{x}_{t=9}$ is used as the first observation, while the final snapshot $\mathbf{x}_{t=t_{\mathrm{end}}}$ is fixed as the second observation, with $t_{\mathrm{end}}=99$ for the dam flow evolution and $t_{\mathrm{end}}=31$ for the cavity flow evolution. The background state for multi-time-step assimilation is the estimated results from the corresponding single-time-step assimilation. This setup ensures that both ensemble-based and variational methods assimilate exactly the same observational information from two time instants. Moreover, since the first assimilation step coincides with that used in the single-time-step experiments, the corresponding results can be directly reused to facilitate a complementary comparison between stationary and non-stationary data assimilation strategies. The resulting distributions of the parameter estimates in the parameter space for both dam and cavity flow cases are illustrated in Figure~\ref{fig:calibration-ellipse}.

The 4D-Var procedure is initialized following the 3D-Var analysis and incorporates additional observational information over multiple time instants. As a result, the corresponding analysis is expected to achieve improved predictive capability in the physical space.
To assess this expected advantage of 4D-Var over 3D-Var, we compare the 4D-Var validation losses (observation term in the 4D-Var cost function~\eqref{eq:4dvar costfunc}) obtained using the analysis parameters produced by the two methods. 

\begin{figure}[htbp]
\centering
  \begin{subfigure}[t]{0.46\textwidth}
    \centering
    \includegraphics[width=\textwidth]{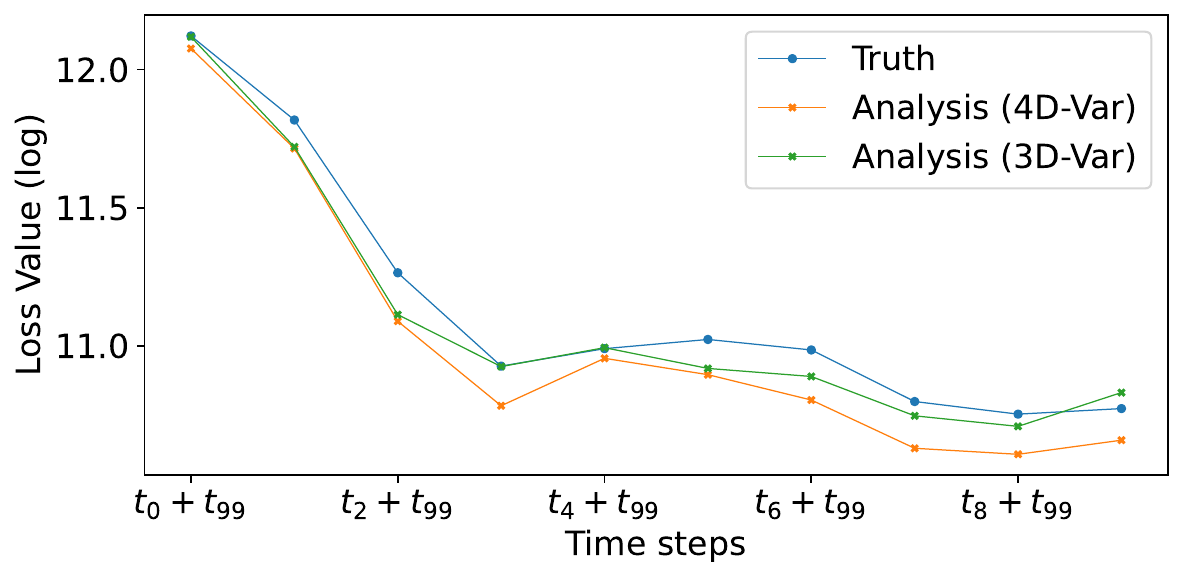}
    \caption{Dam flow --- OACAE-MLP}
  \end{subfigure}
  \quad\quad
  \begin{subfigure}[t]{0.46\textwidth}
    \centering
    \includegraphics[width=\textwidth]{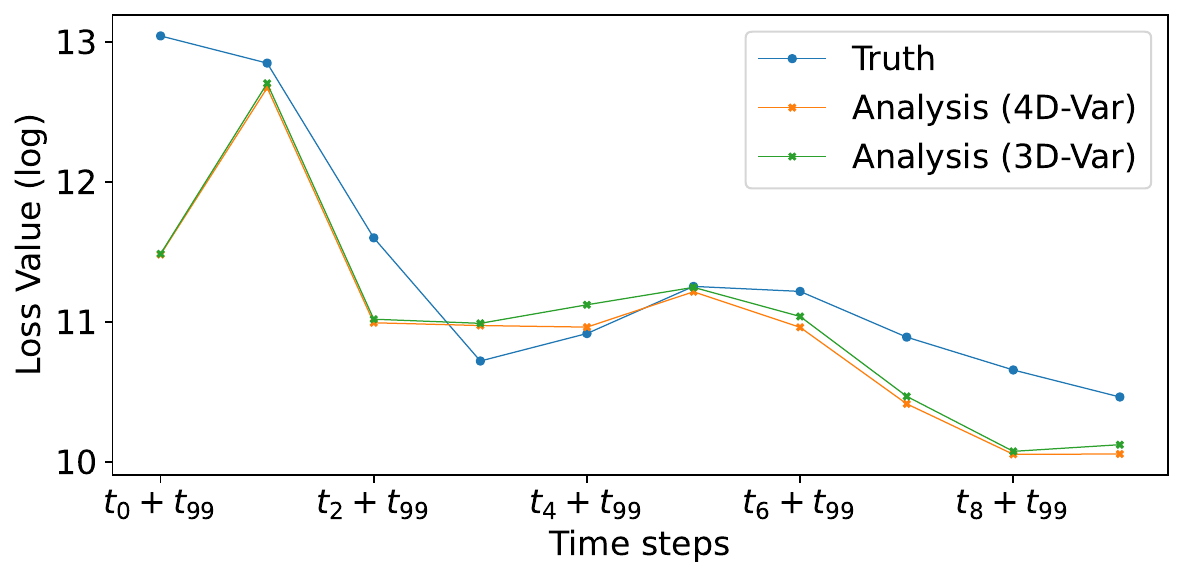}
    \caption{Dam flow --- CAE-MLP}
  \end{subfigure}

  \vspace{1em}

  \begin{subfigure}[t]{0.46\textwidth}
    \centering
    \includegraphics[width=\textwidth]{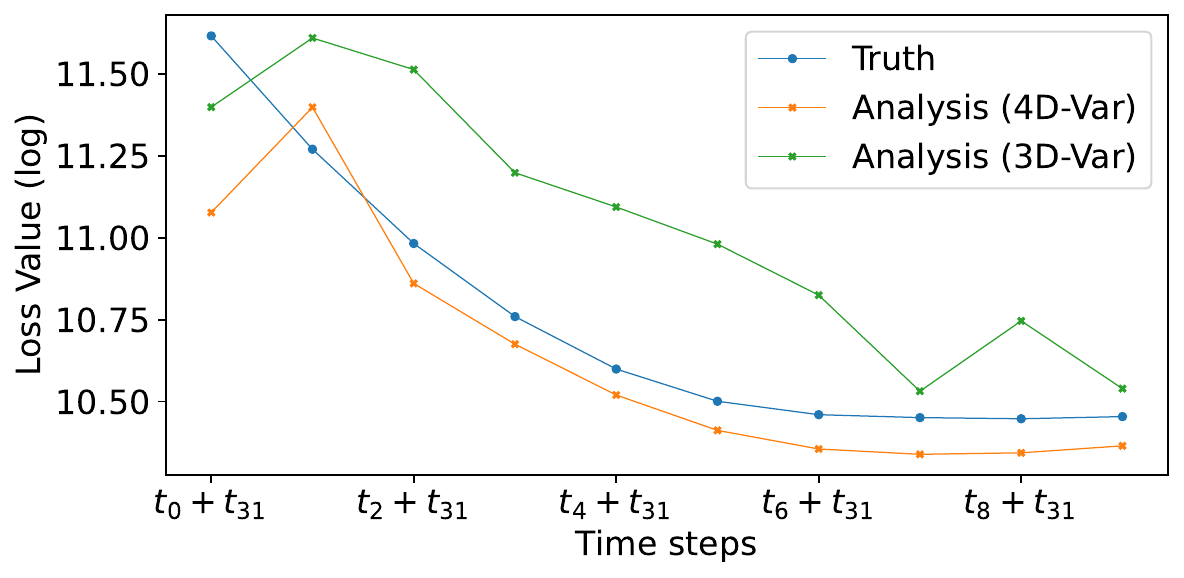}
    \caption{Cavity flow --- OACAE-MLP}
  \end{subfigure}
  \quad\quad
  \begin{subfigure}[t]{0.46\textwidth}
    \centering
    \includegraphics[width=\textwidth]{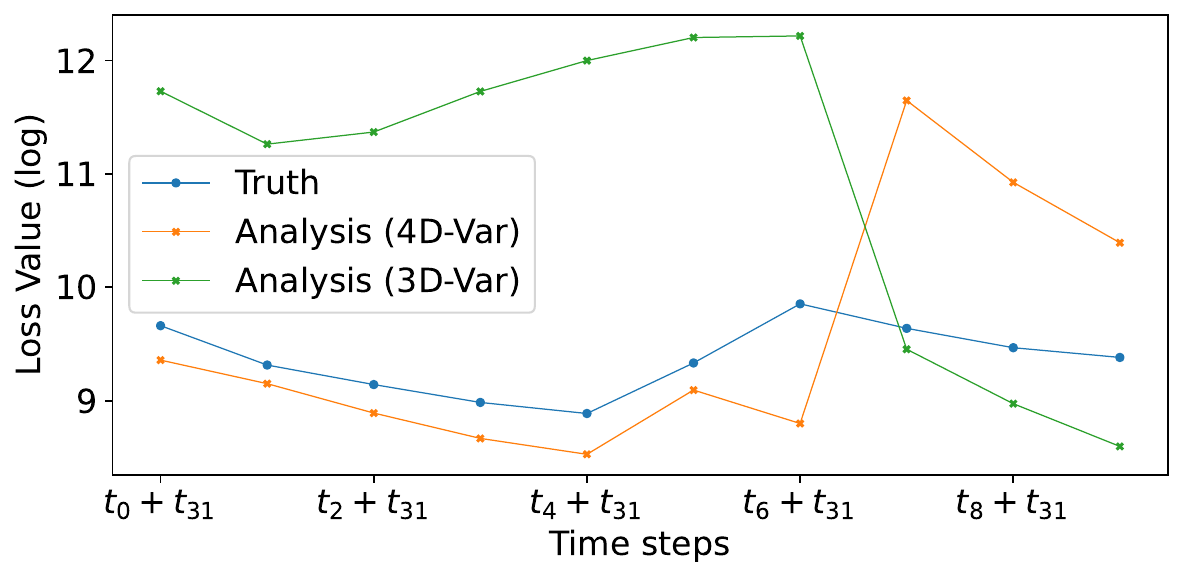}
    \caption{Cavity flow --- CAE-MLP}
  \end{subfigure}

\caption{Comparison of 4D-Var observation mismatch in the physical space, evaluated at selected 4D-Var assimilation time steps, using the true parameters and the analysis parameters from 4D-Var and 3D-Var methods for different surrogate models (OACAE-MLP and CAE-MLP) and flow configurations (dam flow and cavity flow).}
\label{fig:physics-loss-2T}
\end{figure}

Figure~\ref{fig:physics-loss-2T} compares the 4D-Var validation losses obtained using the true parameters and the analysis parameters from 3D-Var and 4D-Var for different surrogate models and flow configurations. For the OACAE-MLP surrogate, the 4D-Var framework consistently achieves lower or comparable validation losses than 3D-Var for both dam and cavity flows, demonstrating that incorporating temporal consistency effectively improves physical-space predictions when a physics-informed latent representation is employed.
By contrast, for the CAE-MLP surrogate, although 4D-Var outperforms 3D-Var in most cases, noticeable degradation is observed for certain test windows, particularly in the cavity flow configuration, indicating reduced robustness.
Moreover, in the majority of tests, the validation losses obtained with the 4D-Var analysis parameters are lower than or equal to those obtained with the true parameters, highlighting the strong capability of the proposed inverse modeling framework to achieve physically consistent parameter calibration.

\begin{figure}[H]
  \centering
  \begin{subfigure}[t]{0.85\textwidth}
    \centering
    \includegraphics[height=0.28\textheight, keepaspectratio]{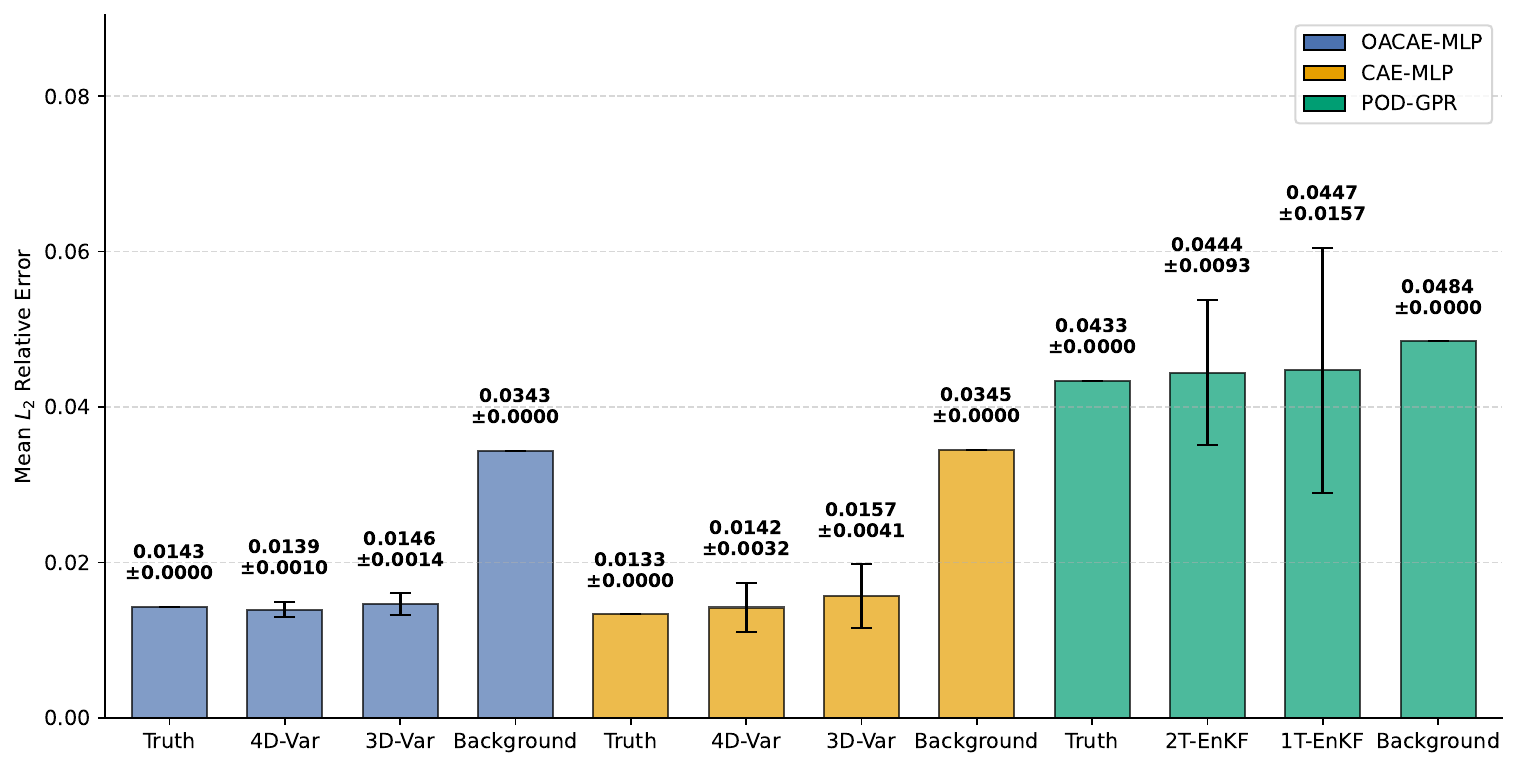}
    \caption{Entire evolution prediction error (Dam flow)}
  \end{subfigure}

  \vspace{0.25cm}

  \begin{subfigure}[t]{0.85\textwidth}
    \centering
    \includegraphics[height=0.28\textheight, keepaspectratio]{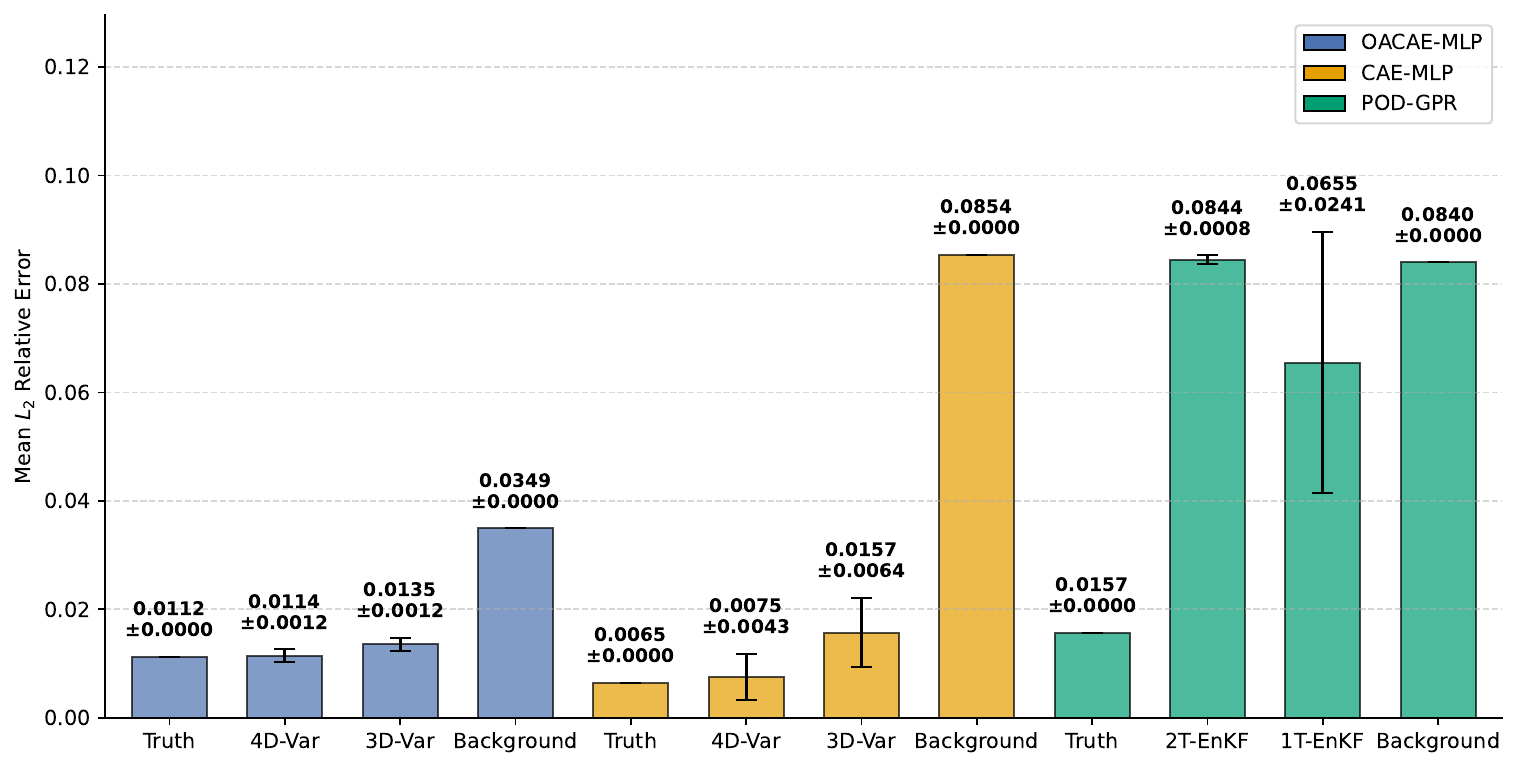}
    \caption{Entire evolution prediction error (Cavity flow)}
  \end{subfigure}

  \caption{Mean relative $\ell_2$ errors in the physical space over the entire temporal evolution using parameters estimated by different data assimilation frameworks.}
  \label{fig:full-evolution-errors}
\end{figure}

While the validation losses reported above evaluate the physical-space consistency at assimilation time steps, the ultimate objective of parameter calibration is to improve long-term prediction of flow dynamics. To this end, we further assess the predictive performance of the calibrated parameters over the entire temporal evolution. Figure~\ref{fig:full-evolution-errors} reports the mean $\ell_2$ relative errors with standard deviations computed over all time steps of the evolution for different calibration strategies. For each estimate, the discrepancy of its predicted trajectory with the reference is averaged over all time steps, yielding a single error value per estimate.  The long-term predictive performance of the calibrated parameters is evaluated based on two complementary criteria: prediction robustness and prediction accuracy.
Robustness is assessed through the uncertainty of the prediction errors, quantified by the standard deviation across different calibration instances, while accuracy is evaluated by comparing the mean prediction error with that obtained using the true parameters. As shown in Figure~\ref{fig:full-evolution-errors}, the variational DA-DL-ROM frameworks, particularly when combined with the OACAE-MLP surrogate, consistently exhibit low prediction uncertainty for both dam and cavity flow configurations. Notably, for each flow configuration, the uncertainty associated with the 3D-Var-OACAE-MLP framework is even lower than that of the 4D-Var-CAE-MLP framework, highlighting the strong impact of physics-aware latent representations on stabilizing the assimilation process. In terms of accuracy, the prediction errors obtained using the calibrated parameters are either lower than or comparable to those obtained using true parameters. For the dam flow configuration, the 4D-Var-OACAE-MLP framework achieves lower prediction errors than the reference level, whereas for the cavity flow configuration, the corresponding errors remain close to those obtained with the true parameters. In both cases, the OACAE-based predictions are consistently closer to the reference than their CAE-based counterpart. A quantitative comparison further highlights this difference. For the dam flow, the relative deviations with respect to the true prediction are 2.80\% and 2.10\% for the OACAE-based framework using 4D-Var and 3D-Var, whereas the corresponding deviations for the CAE-based framework reach 6.77\% for 4D-Var and 18.05\% for 3D-Var. For the cavity flow, the OACAE-based framework yields relative deviations of 1.79\% for 4D-Var and 20.54\% for 3D-Var, while the CAE-based framework exhibits substantially larger deviations of 15.38\% for 4D-Var and 141.54\% for 3D-Var. 
By contrast, the EnKF-POD-GPR baseline exhibits substantially larger prediction errors and uncertainties, highlighting its limited robustness for long-term prediction, particularly in the cavity flow configuration where the 2T-EnKF strategy leads to increased errors over the entire evolution. Overall, these results demonstrate that the proposed variational DA-DL-ROM framework not only improves parameter estimation but also leads to accurate and robust long-term predictions in physical space.

\begin{figure}[htbp]
\centering
\subfloat[Background u]{\includegraphics[width = 1in]{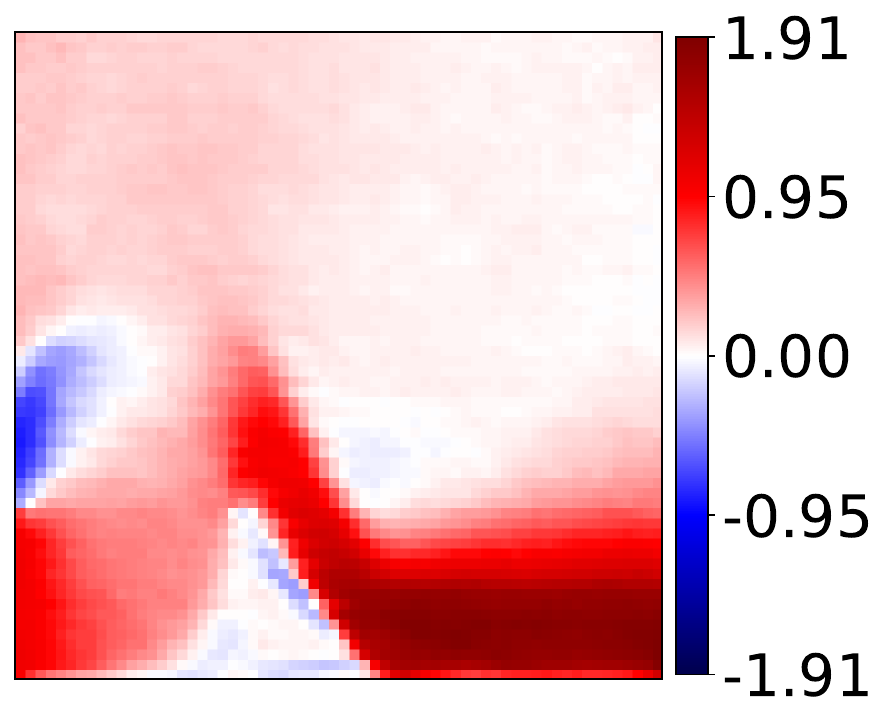}}
\qquad
\subfloat[Background error-u]{\includegraphics[width = 1in]{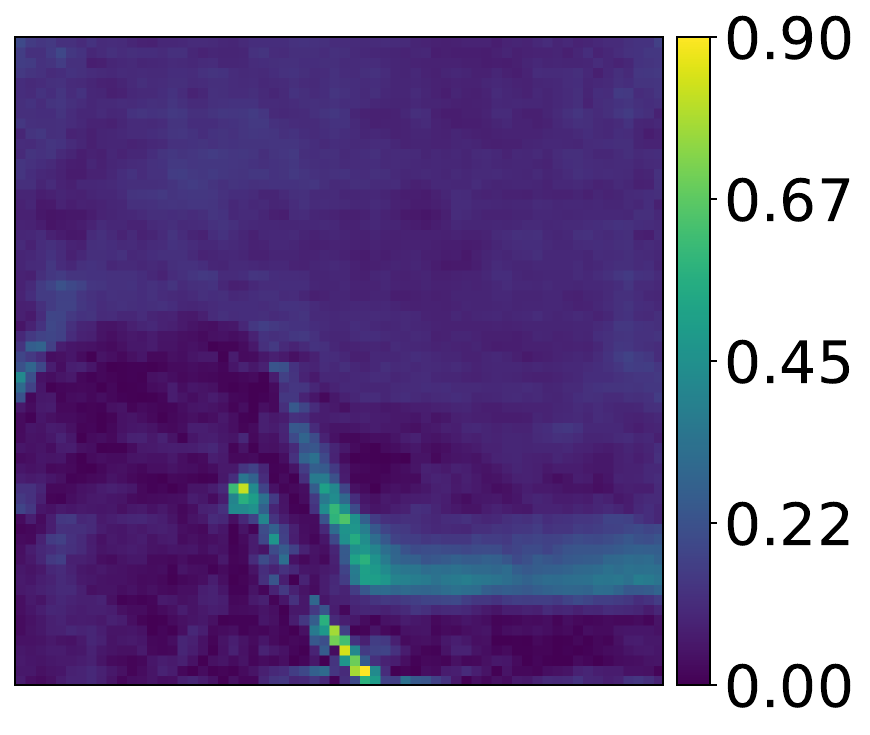}}
\qquad
\subfloat[Background v]{\includegraphics[width = 1in]{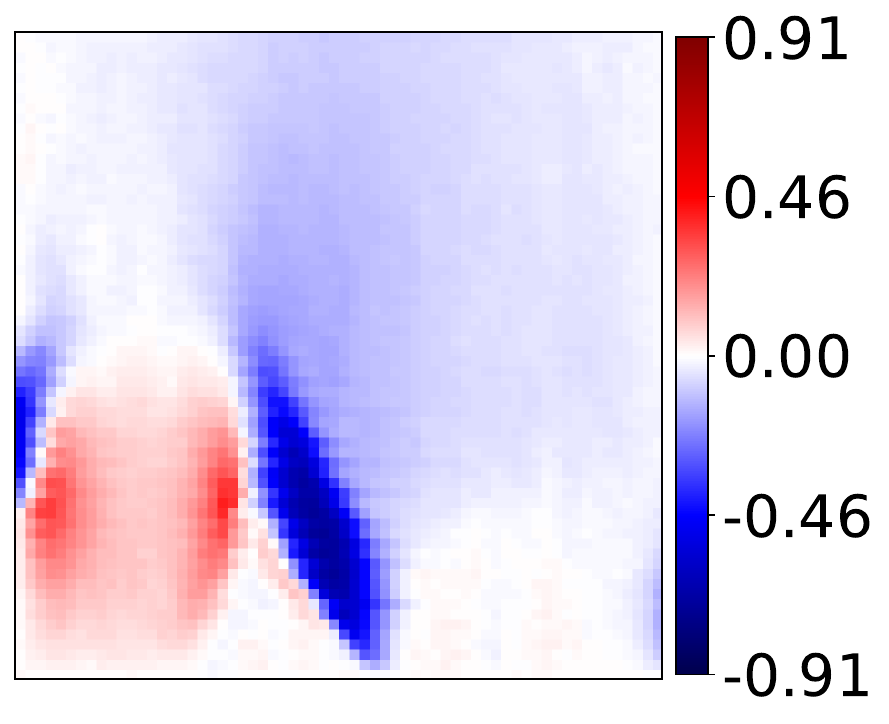}}
\qquad
\subfloat[Background error-v]{\includegraphics[width = 1in]{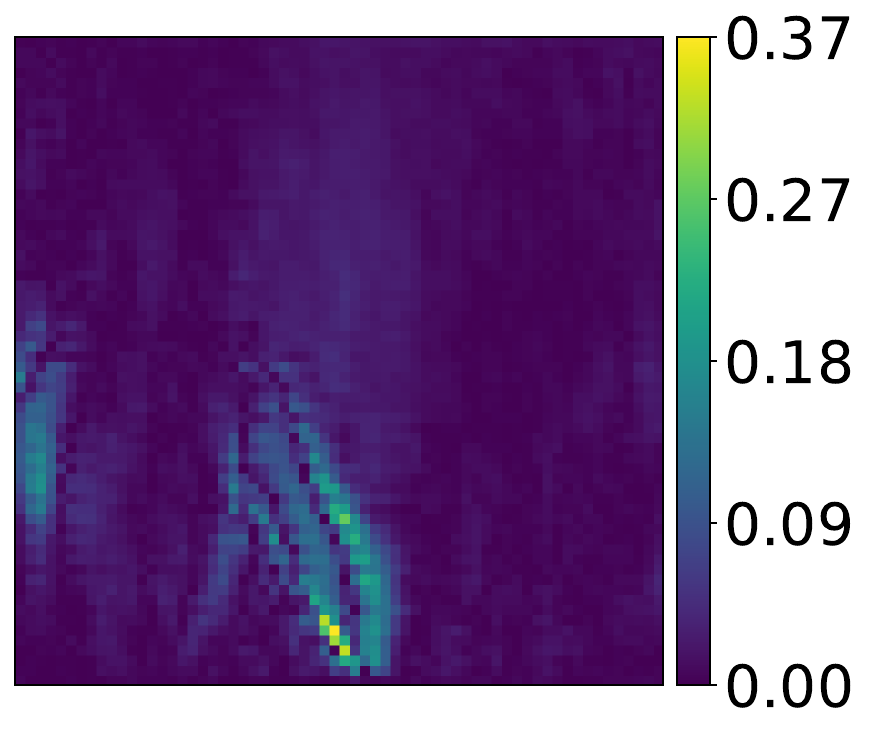}}\\
\subfloat[3D-Var u]{\includegraphics[width = 1in]{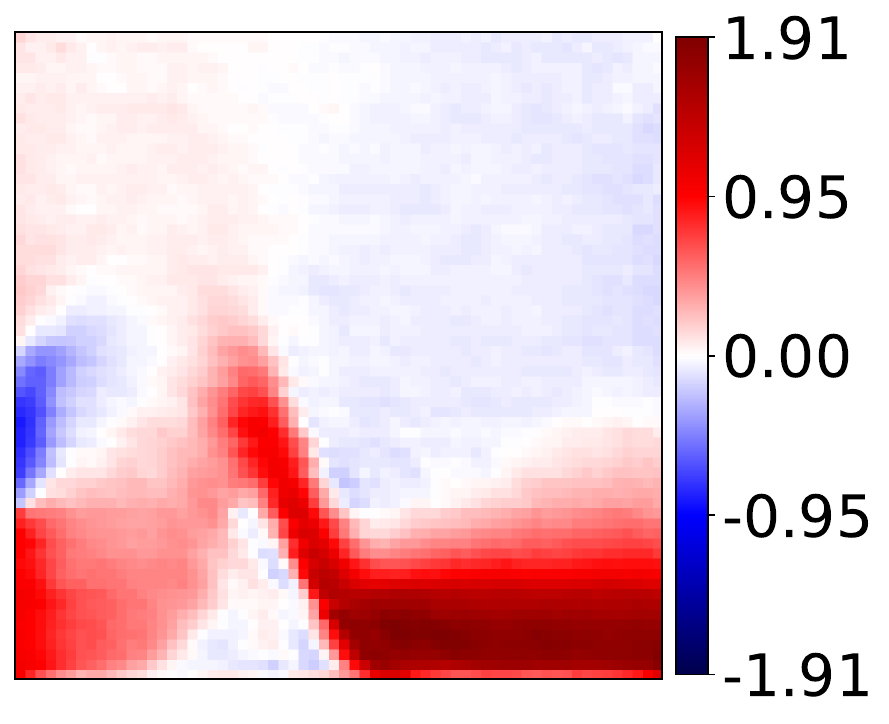}}
\qquad
\subfloat[3D-Var-error-u]{\includegraphics[width = 1in]{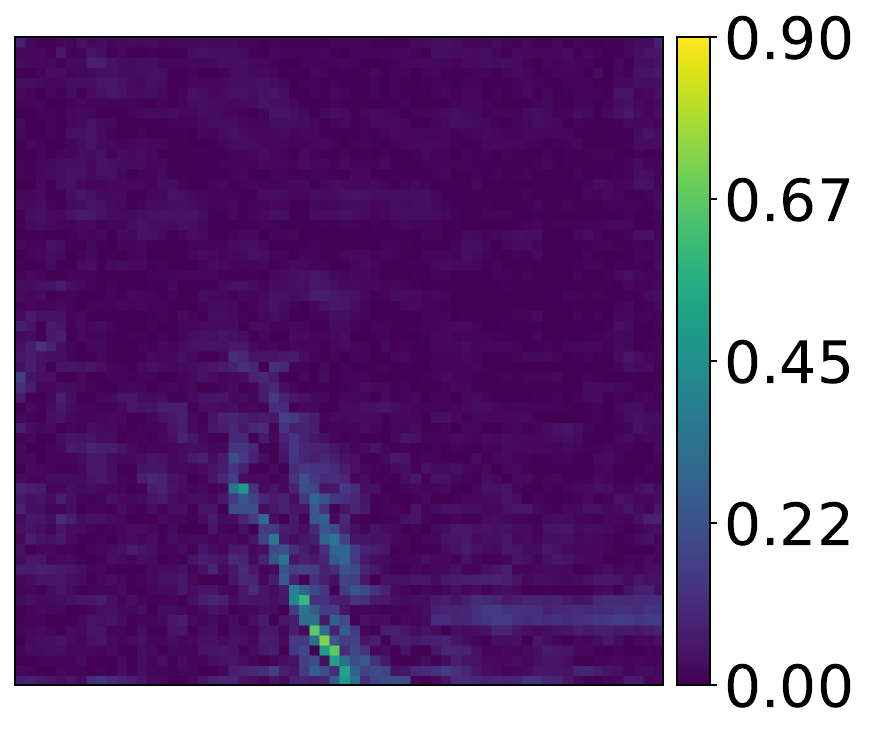}}
\qquad
\subfloat[3D-Var v]{\includegraphics[width = 1in]{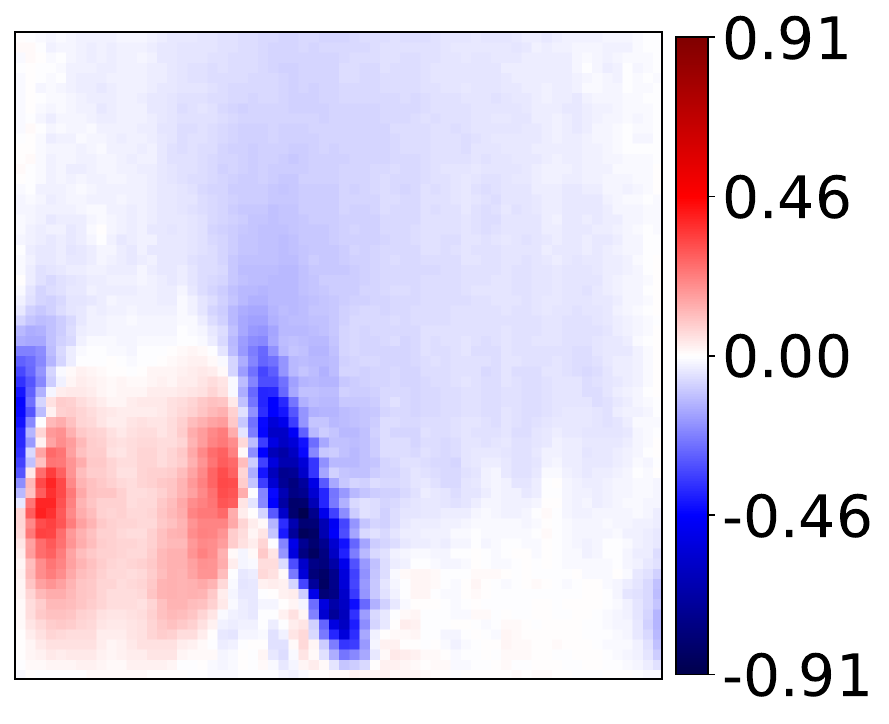}}
\qquad
\subfloat[3D-Var-error-v]{\includegraphics[width = 1in]{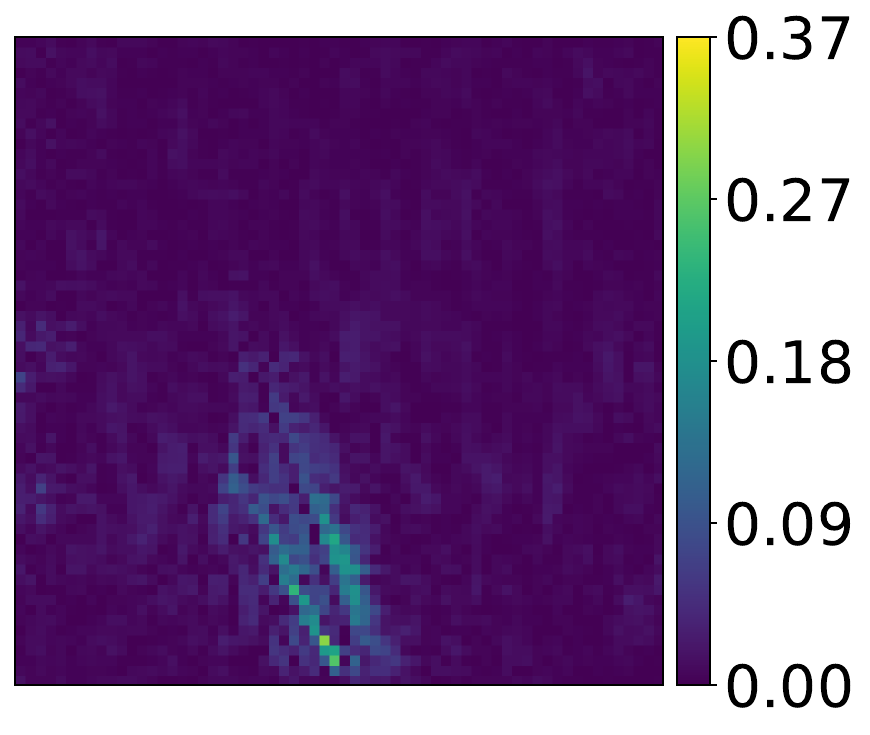}}\\
\subfloat[4D-Var u]{\includegraphics[width = 1in]{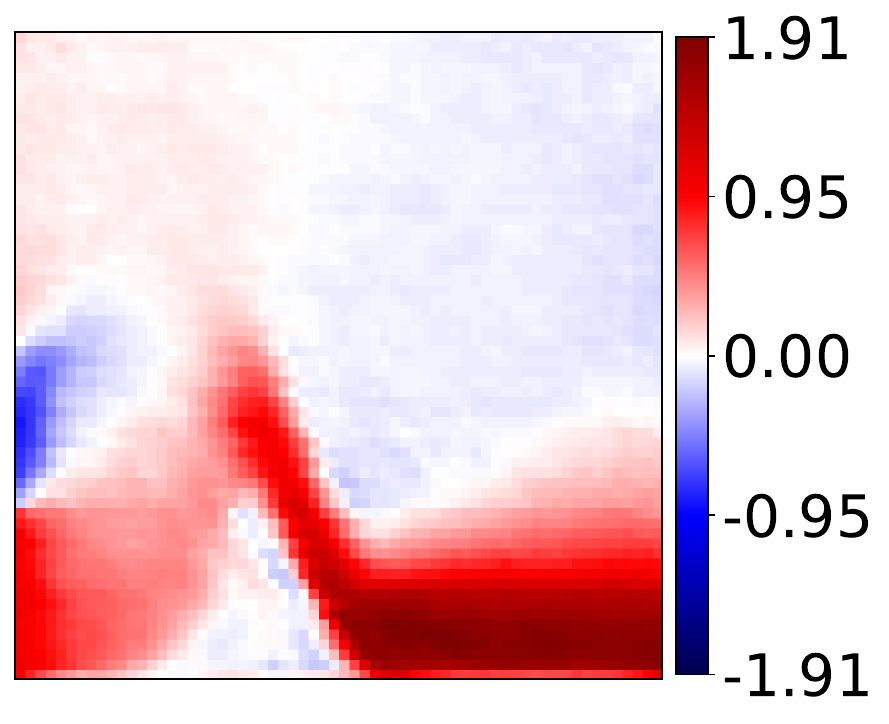}}
\qquad
\subfloat[4D-Var-error u]{\includegraphics[width = 1in]{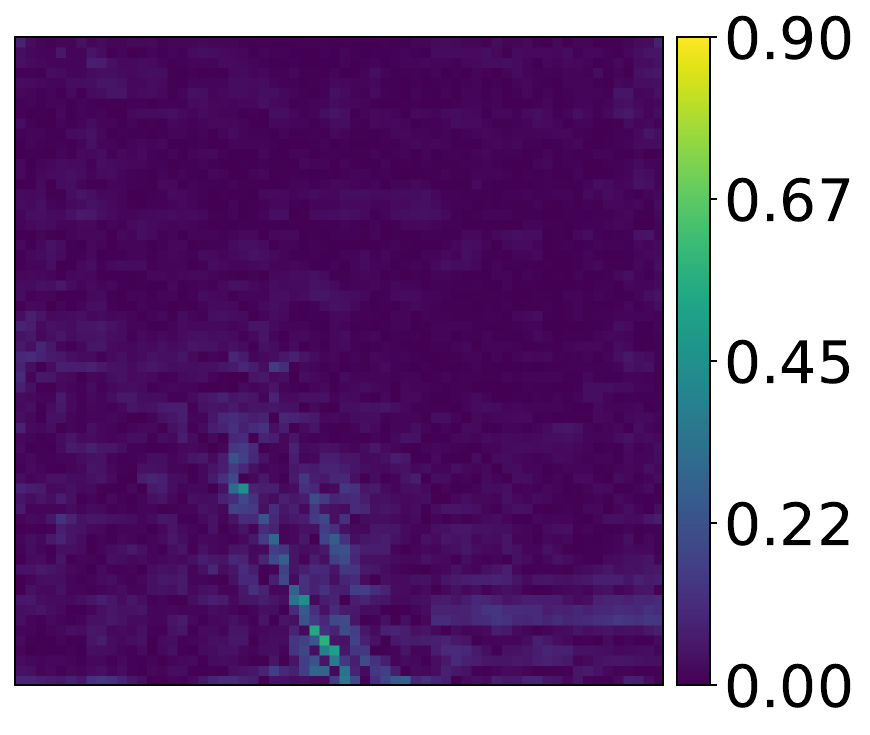}}
\qquad
\subfloat[4D-Var v]{\includegraphics[width = 1in]{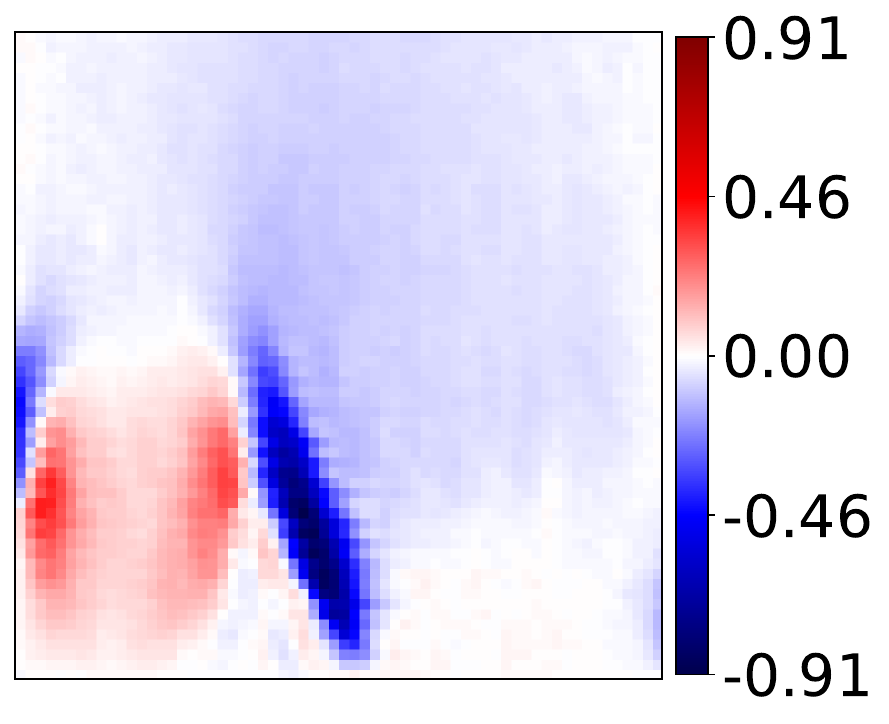}}
\qquad
\subfloat[4D-Var error v]{\includegraphics[width = 1in]{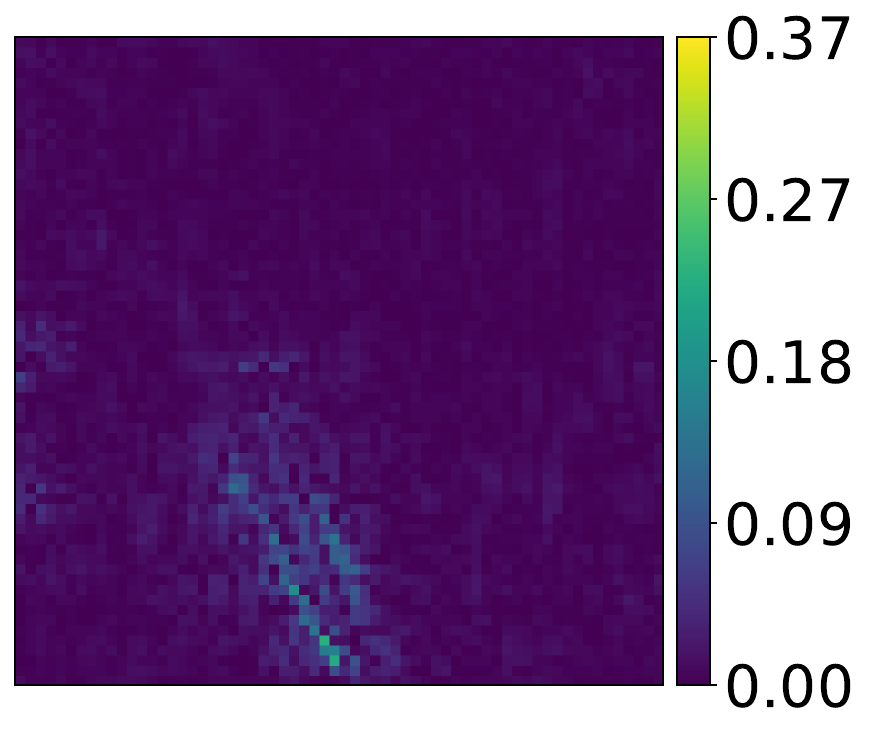}}\\
\subfloat[True prediction u]{\includegraphics[width = 1in]{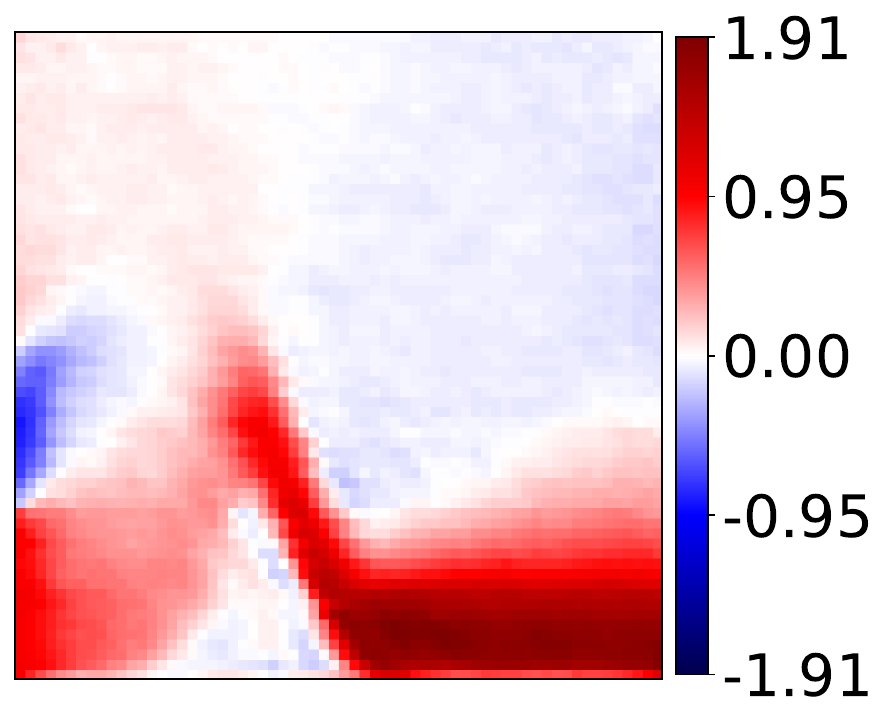}}
\qquad
\subfloat[True prediction error u]{\includegraphics[width = 1in]{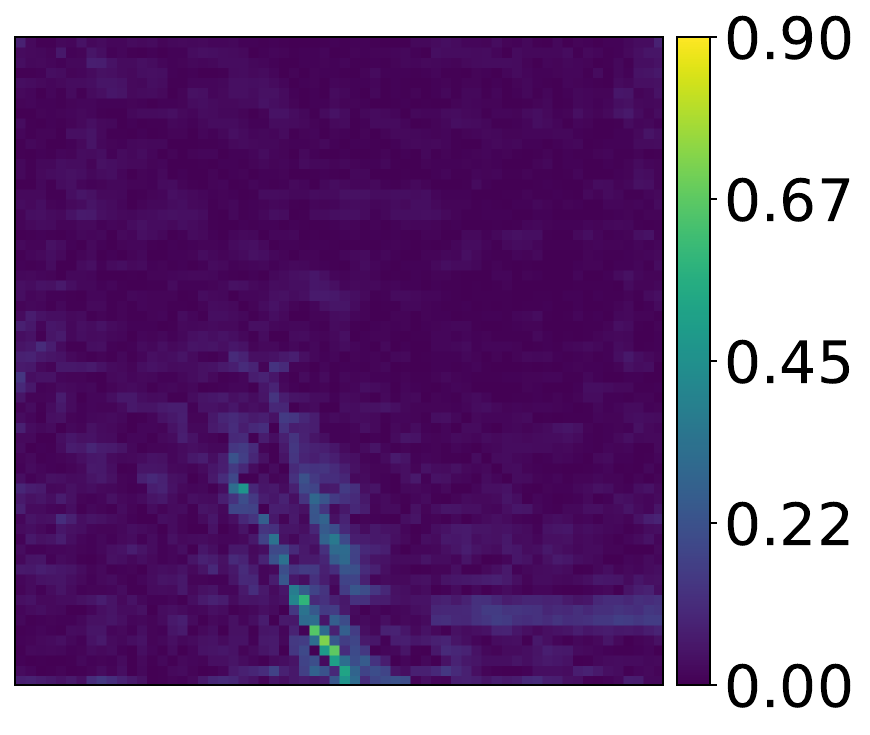}}
\qquad
\subfloat[True prediction v]{\includegraphics[width = 1in]{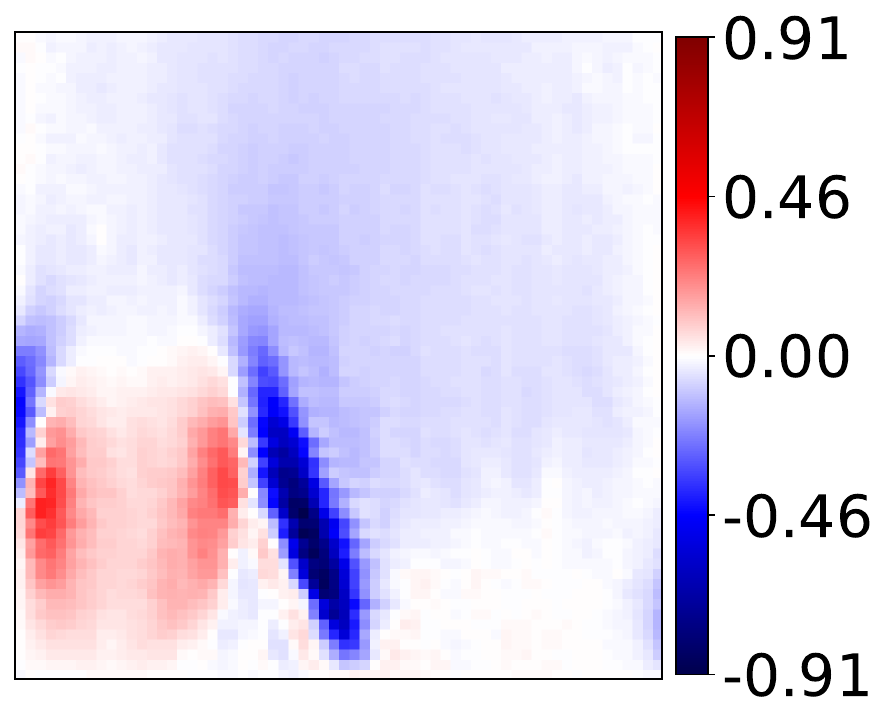}}
\qquad
\subfloat[True prediction error v]{\includegraphics[width = 1in]{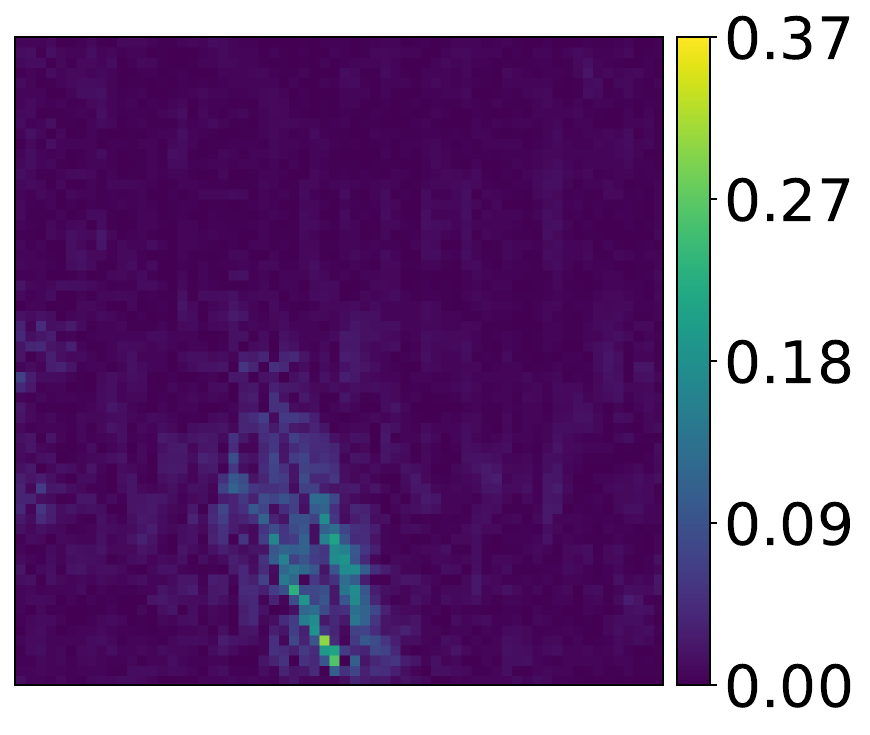}}\\
\subfloat[Background u]{\includegraphics[width = 1in]{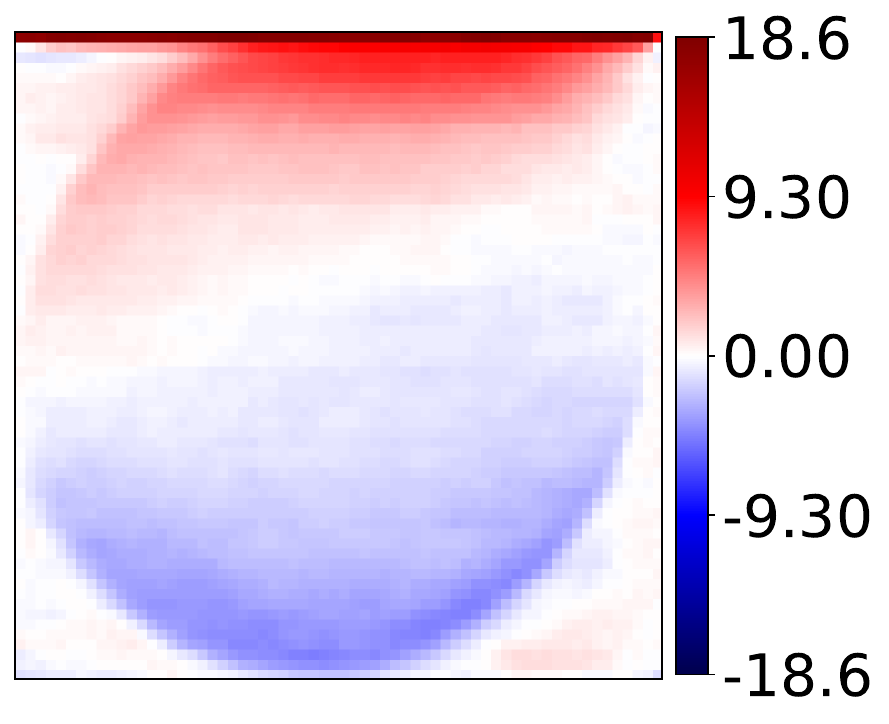}}
\qquad
\subfloat[Background error-u]{\includegraphics[width = 1in]{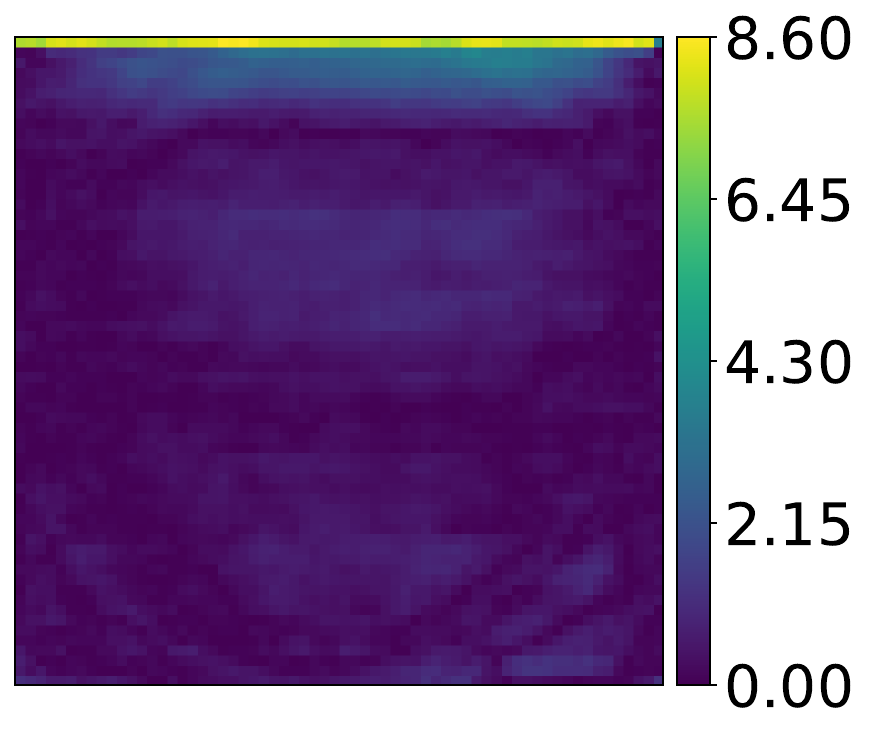}}
\qquad
\subfloat[Background v]{\includegraphics[width = 1in]{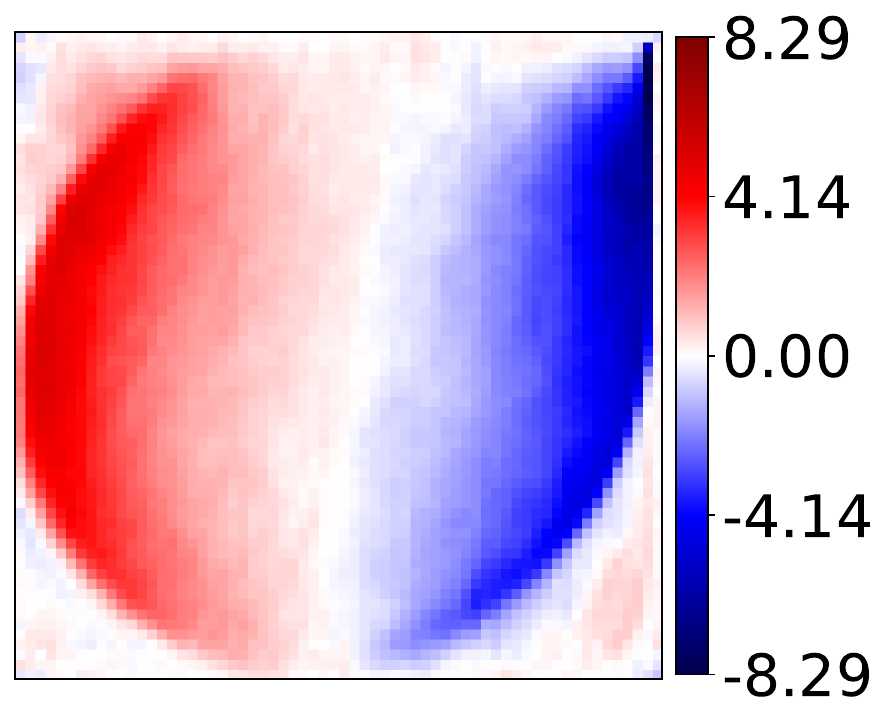}}
\qquad
\subfloat[Background error-v]{\includegraphics[width = 1in]{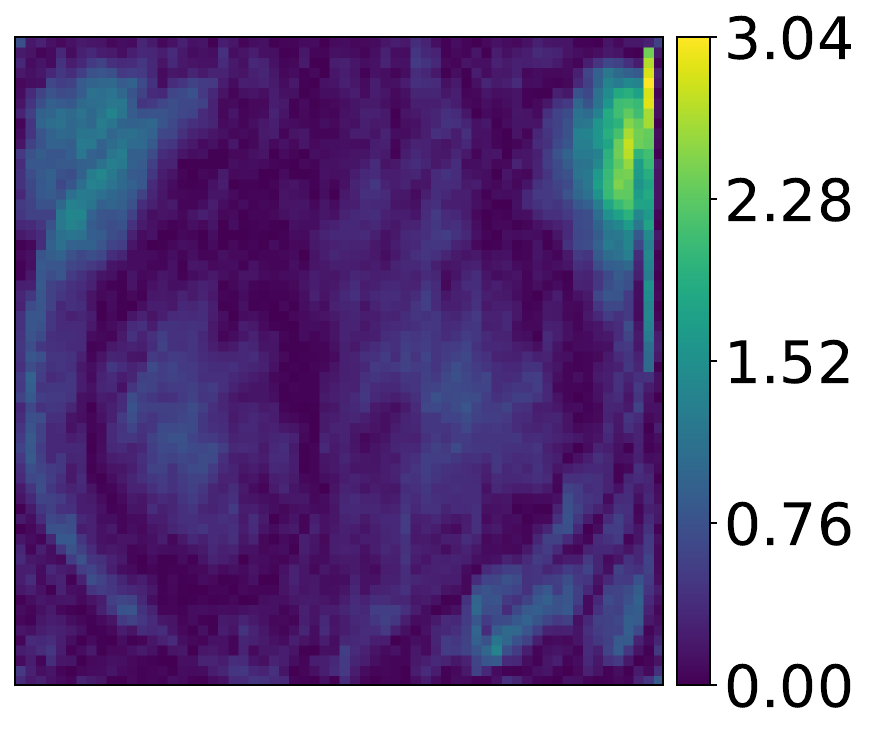}}\\
\subfloat[3D-Var u]{\includegraphics[width = 1in]{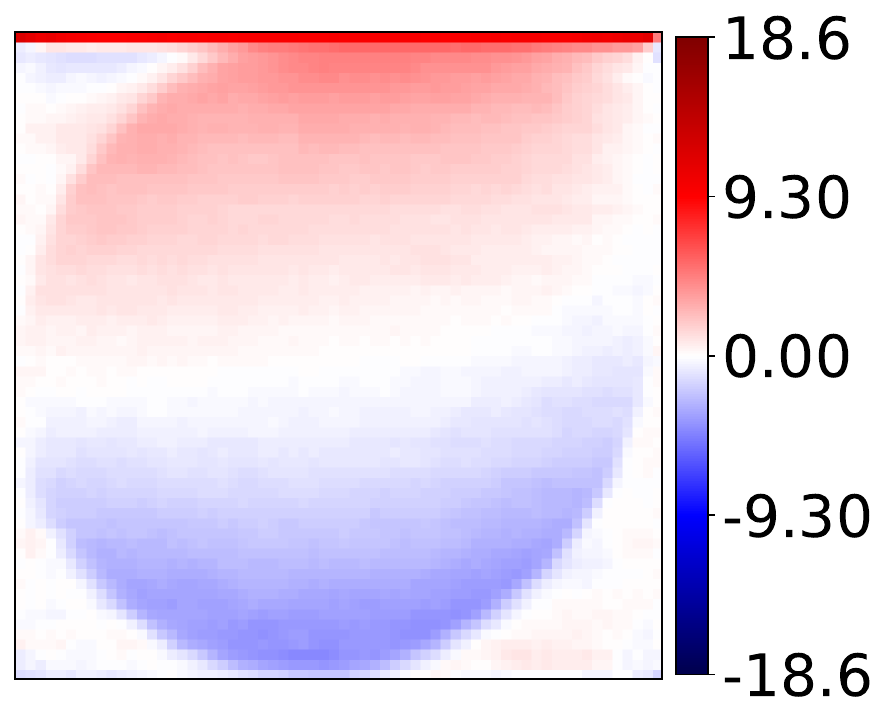}}
\qquad
\subfloat[3D-Var-error-u]{\includegraphics[width = 1in]{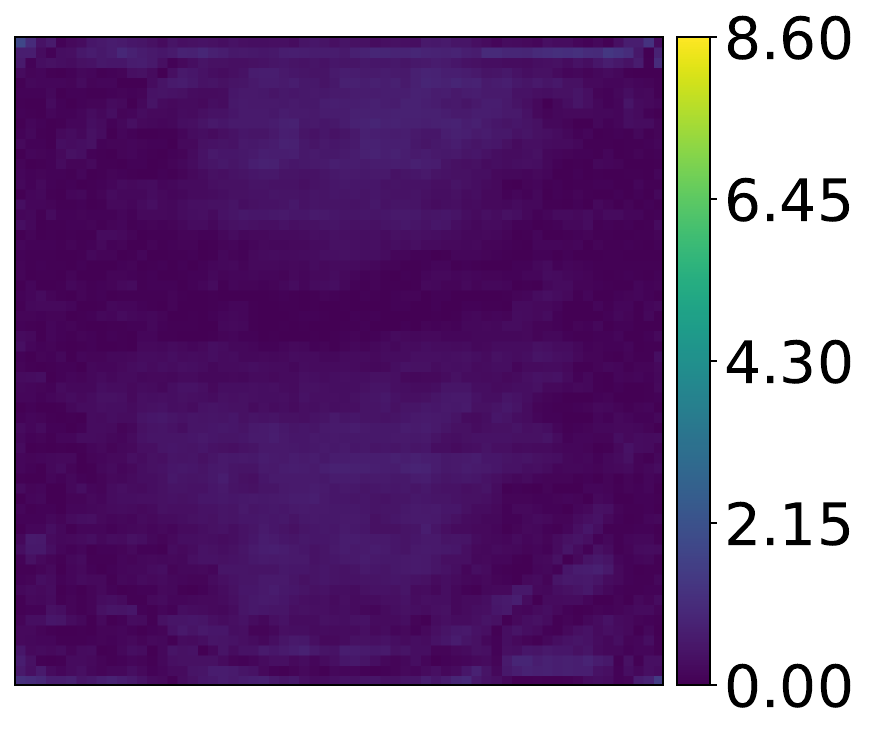}}
\qquad
\subfloat[3D-Var v]{\includegraphics[width = 1in]{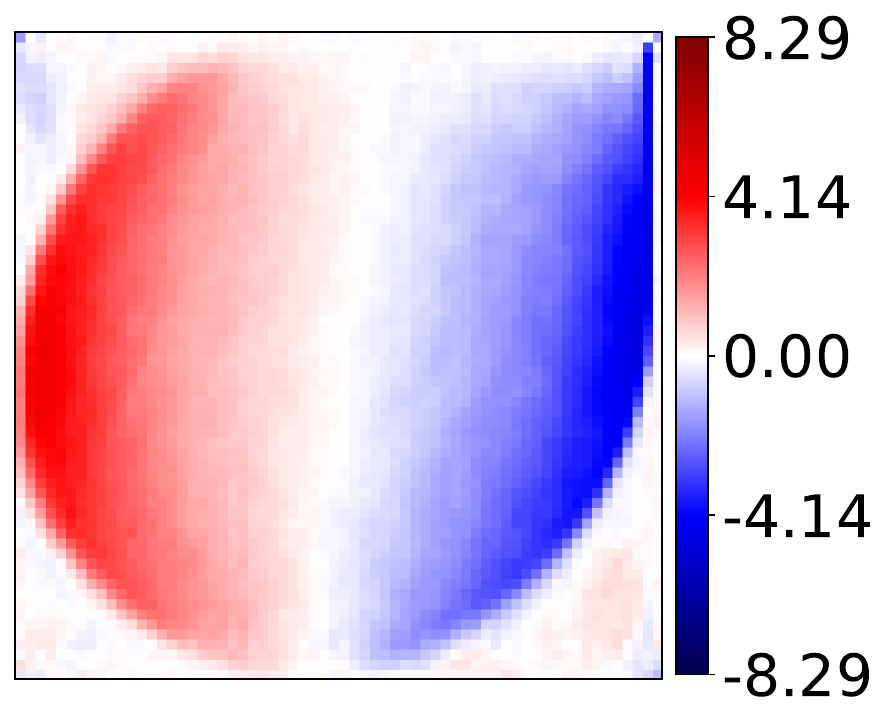}}
\qquad
\subfloat[3D-Var-error-v]{\includegraphics[width = 1in]{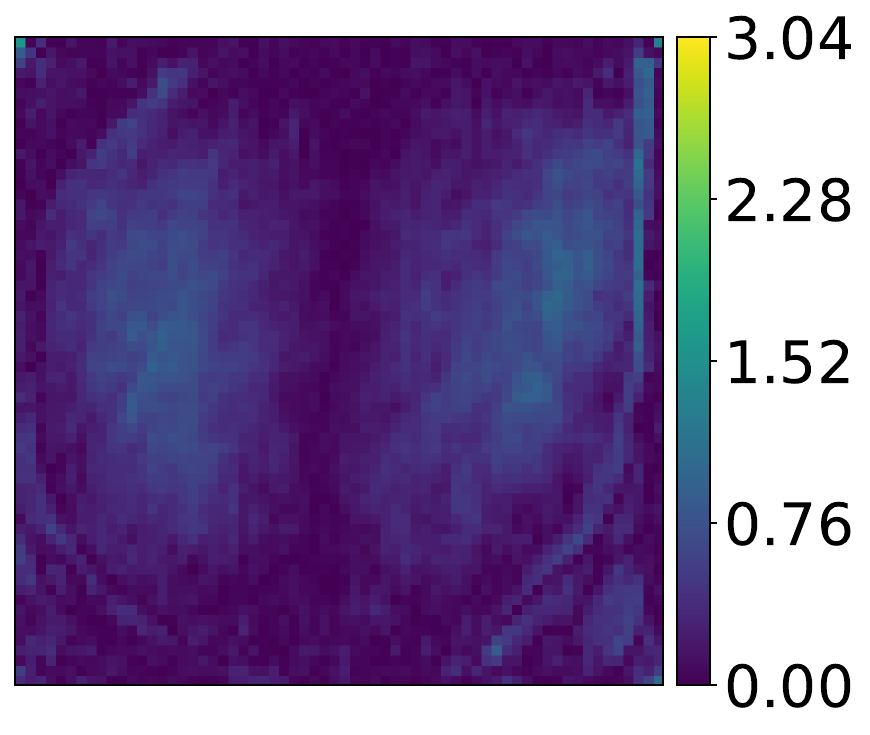}}\\
\subfloat[4D-Var u]{\includegraphics[width = 1in]{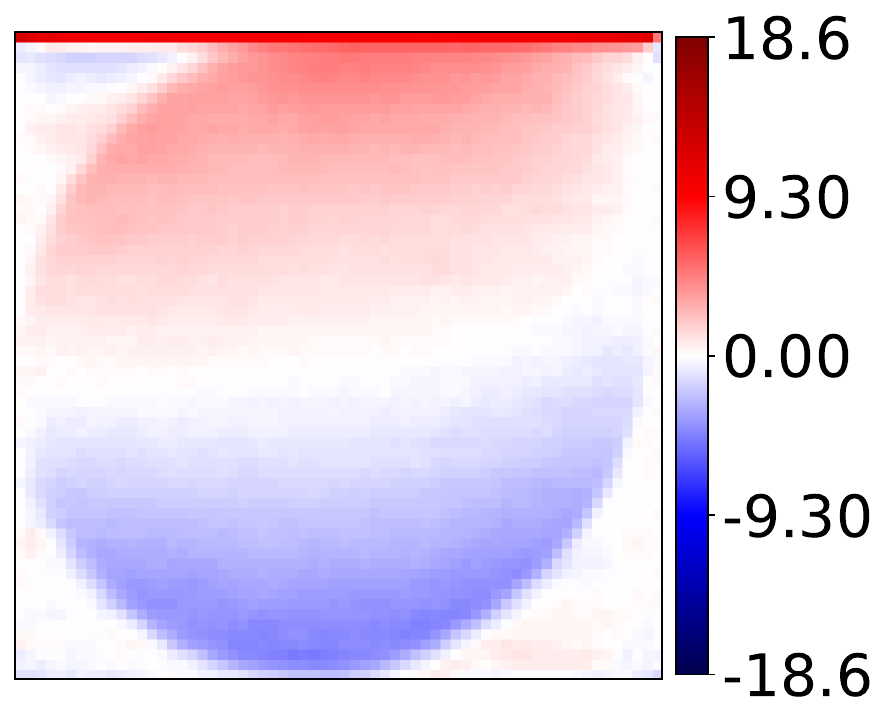}}
\qquad
\subfloat[4D-Var-error u]{\includegraphics[width = 1in]{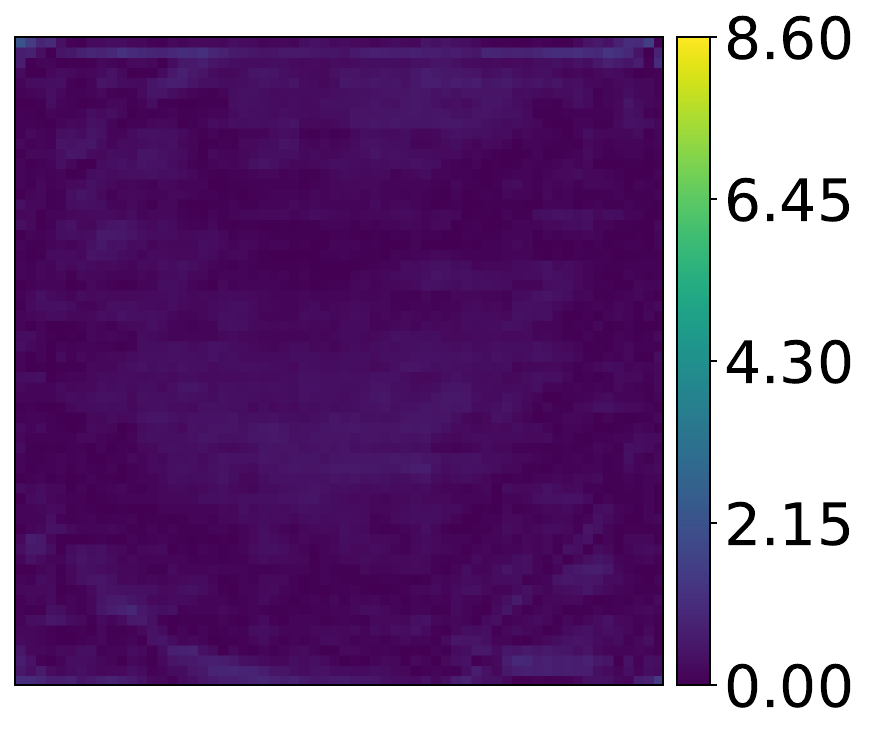}}
\qquad
\subfloat[4D-Var v]{\includegraphics[width = 1in]{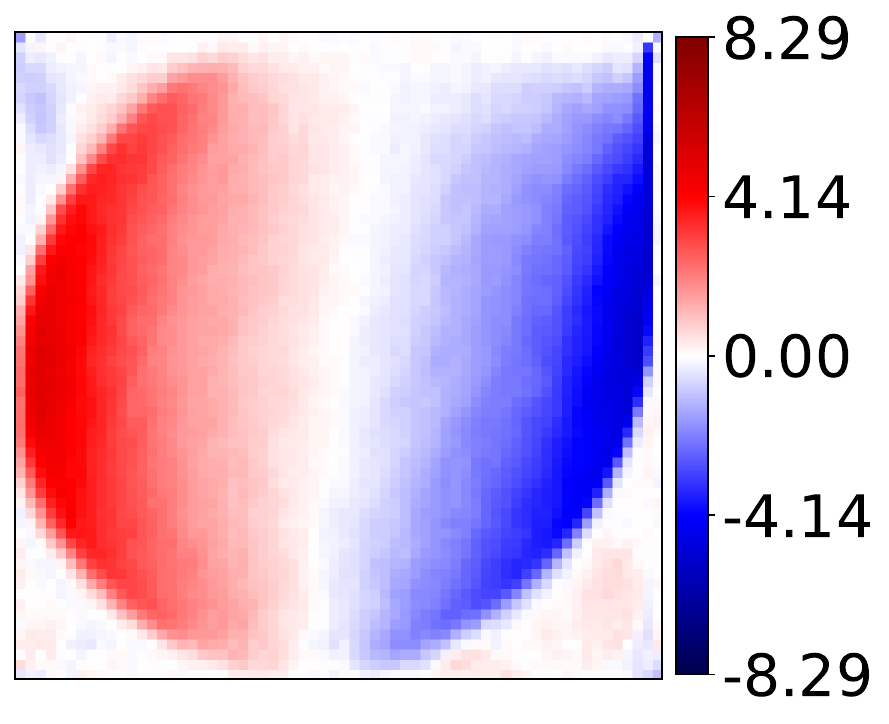}}
\qquad
\subfloat[4D-Var error v]{\includegraphics[width = 1in]{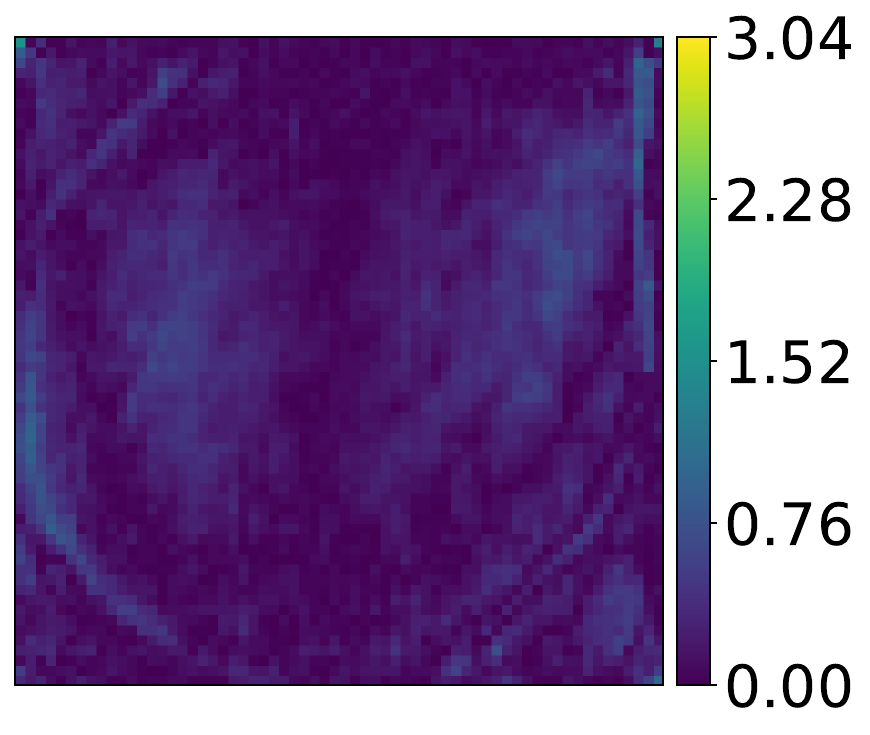}}\\
\subfloat[True prediction u]{\includegraphics[width = 1in]{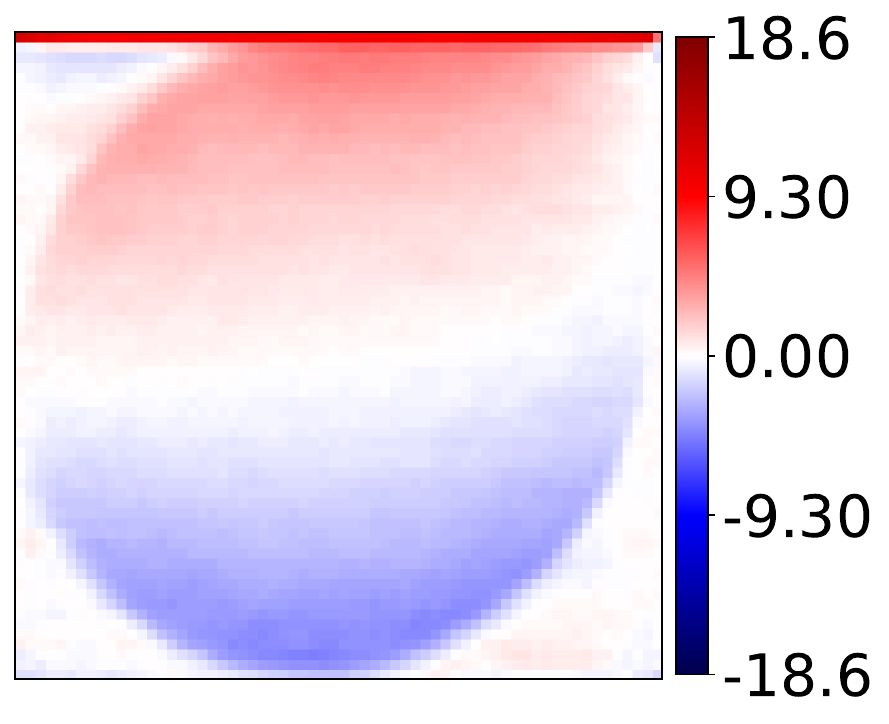}}
\qquad
\subfloat[True prediction error u]{\includegraphics[width = 1in]{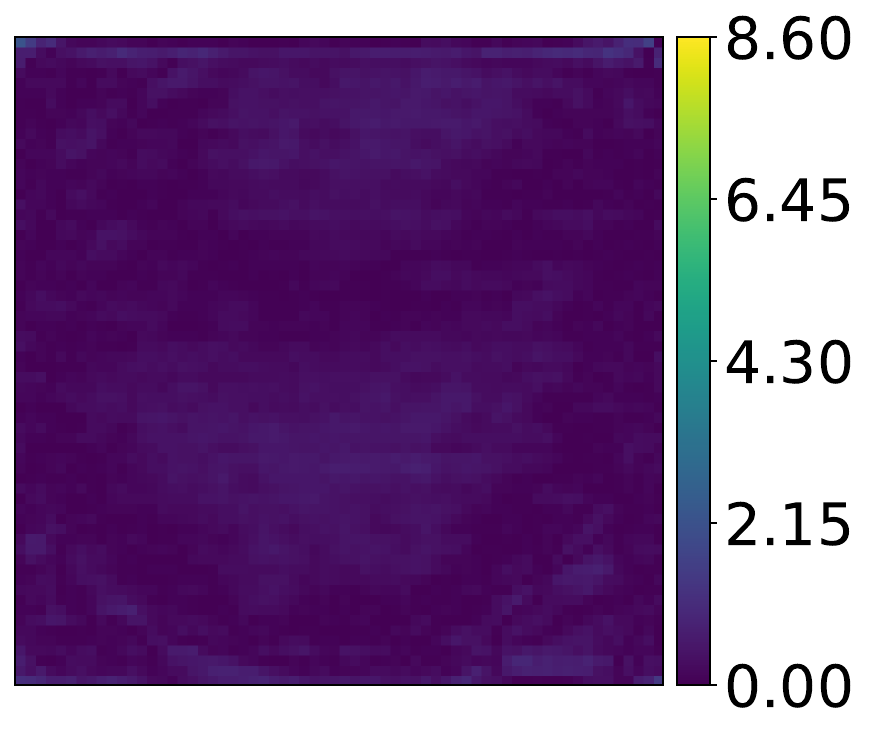}}
\qquad
\subfloat[True prediction v]{\includegraphics[width = 1in]{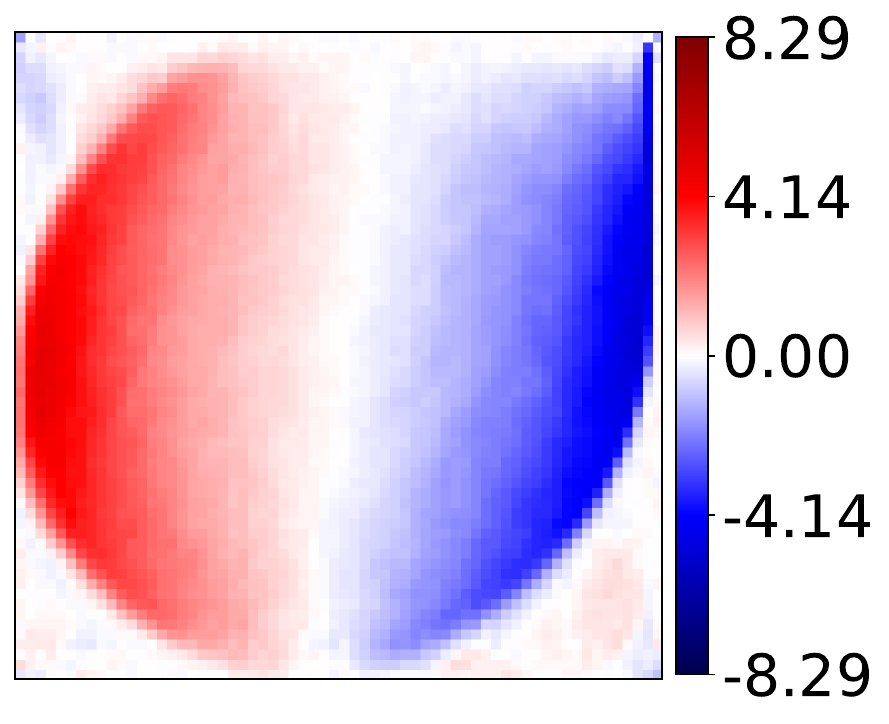}}
\qquad
\subfloat[True prediction error v]{\includegraphics[width = 1in]{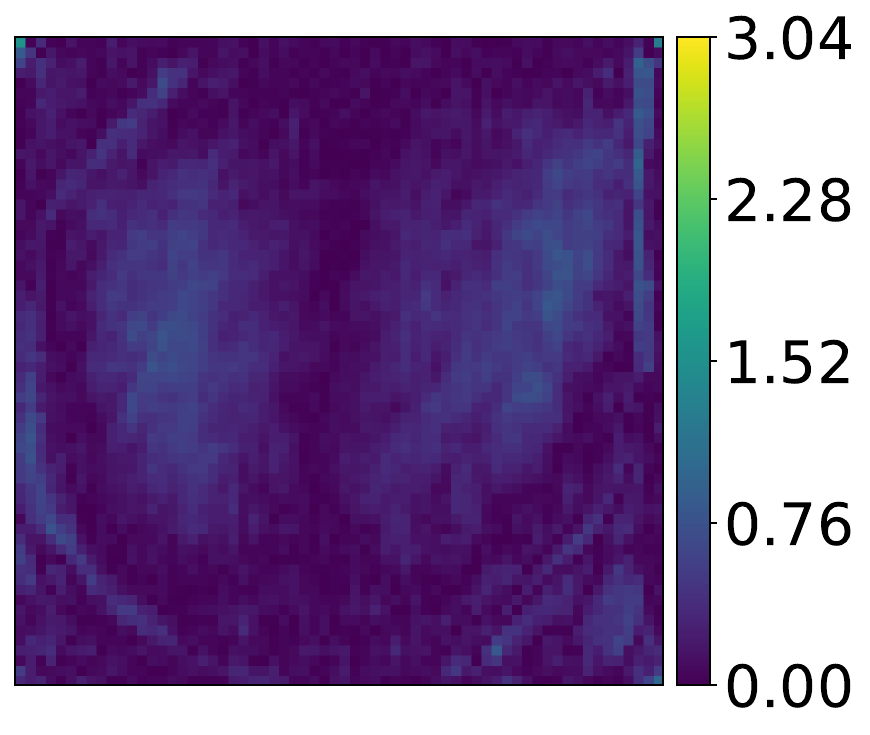}}\\
   \caption{}
   \label{fig:DA-visualize-dam-cavity}
\end{figure}

To further illustrate the physical impact of parameter calibration, in Figure~\ref{fig:DA-visualize-dam-cavity}, we visualize representative flow fields obtained using the background parameters, the parameters calibrated by 3D-Var and 4D-Var, and the corresponding true parameter with the OACAE-MLP model.

\subsubsection{Robustness to different evolution cases.}
As discussed in Section~\ref{sec: single-da}, the effectiveness of the proposed physics-informed variational DA-DL-ROM framework for parameter calibration is first demonstrated using all time steps from a fixed evolution case in the test dataset. To further assess the robustness of the proposed framework across different evolution cases, we perform an additional sensitivity analysis on the dam flow test dataset. The dam flow test dataset contains 44 distinct evolution cases, each consisting of 100 time steps, which provides a suitable basis for evaluating case-to-case robustness. For each evolution case, a single snapshot at the same time index (the 50th time step) is selected for assimilation, and the same single-time-step data assimilation procedures are applied without any modification to the data assimilation settings or surrogate models. As a result, each framework produces 44 independent parameter estimates, enabling consistent comparison of robustness across different evolution cases.

\begin{figure}[h!]
    \centering
    \includegraphics[width=0.7\textwidth]{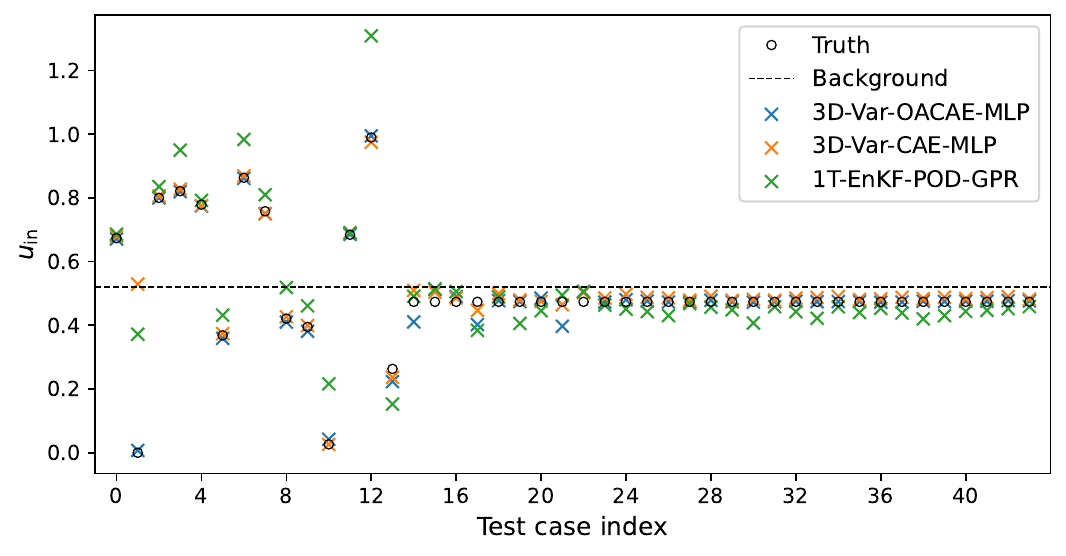}

    \vspace{0.5em}

    \includegraphics[width=0.7\textwidth]{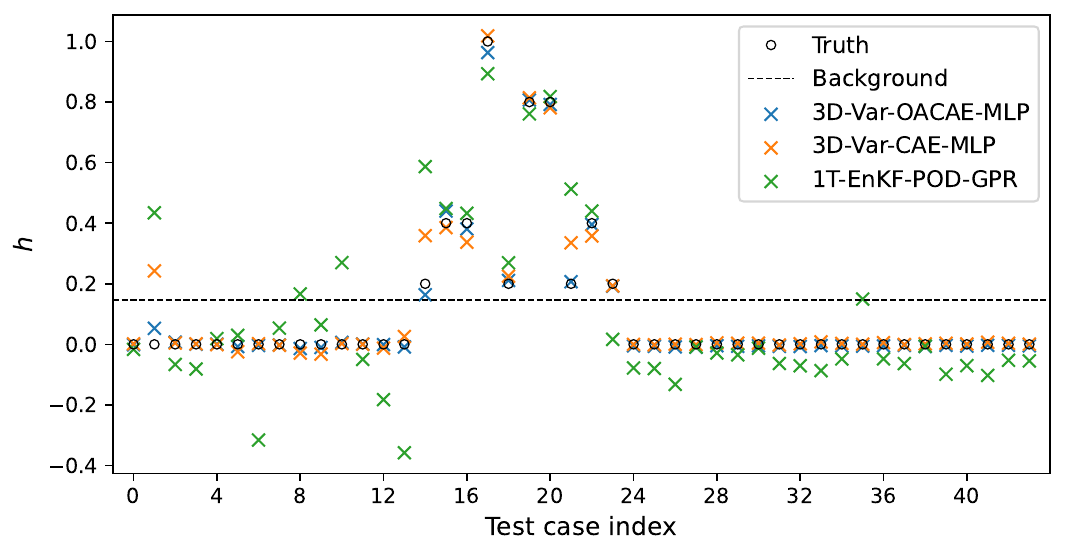}

    \caption{Distribution of calibrated parameters across different evolution cases for the dam flow problem.}
    \label{fig:case-to-case-params-distribution}
\end{figure}

Figure~\ref{fig:case-to-case-params-distribution} shows the calibrated parameter values for each case and each framework, together with the corresponding ground-truth and background values for reference. Each marker corresponds to the parameter estimate obtained from an evolution case using single-time-step data assimilation. The horizontal dashed line indicates the background value, while the circle markers denote the ground-truth parameters. Results obtained with different data assimilation frameworks are shown for comparison. The variational DA-DL-ROM approaches yield more concentrated and consistent parameter estimates across cases, whereas larger dispersion is observed for the EnKF-POD-GPR baseline, indicating improved case-to-case robustness of the proposed framework.

Table~\ref{tab:case-to-case-param-error} reports the average parameter estimation errors together with their standard deviations over all 44 test cases.

\begin{table}[htbp]
\centering
\caption{Parameter estimation errors for the dam flow problem using single-time-step data assimilation across 44 evolution cases. The results show that the variational DA-DL-ROM frameworks yield significantly lower estimation errors and reduced parameter-space variability compared with the EnKF-POD-GPR baseline.
In particular, the 3D-Var-OACAE-MLP approach achieves the lowest errors and the smallest uncertainty for both parameters, indicating improved robustness of parameter calibration across different evolution cases.
}
\label{tab:case-to-case-param-error}
\begin{tabular}{lccccc}
\toprule
& \multicolumn{2}{c}{$u_{in}$} & \multicolumn{2}{c}{$h$} \\
\cmidrule(lr){2-3} \cmidrule(lr){4-5}
Method & MSE~$\downarrow$ & STD~$\downarrow$ & MSE~$\downarrow$ & STD~$\downarrow$ \\
\midrule
1T-EnKF-POD-GPR   & $8.70\times10^{-3}$ & $9.18\times10^{-2}$ & $2.24\times10^{-2}$ & $1.49\times10^{-1}$ \\
3D-Var-CAE-MLP   & $6.58\times10^{-3}$ & $7.86\times10^{-2}$ & $2.58\times10^{-3}$ & $4.98\times10^{-2}$ \\
3D-Var-OACAE-MLP & $\mathbf{4.66\times10^{-4}}$ & $\mathbf{2.13\times10^{-2}}$ & $\mathbf{2.04\times10^{-4}}$ & $\mathbf{1.40\times10^{-2}}$ \\
\bottomrule
\end{tabular}
\end{table}

\begin{figure}[h!]
    \centering
    \includegraphics[width=0.85\textwidth]{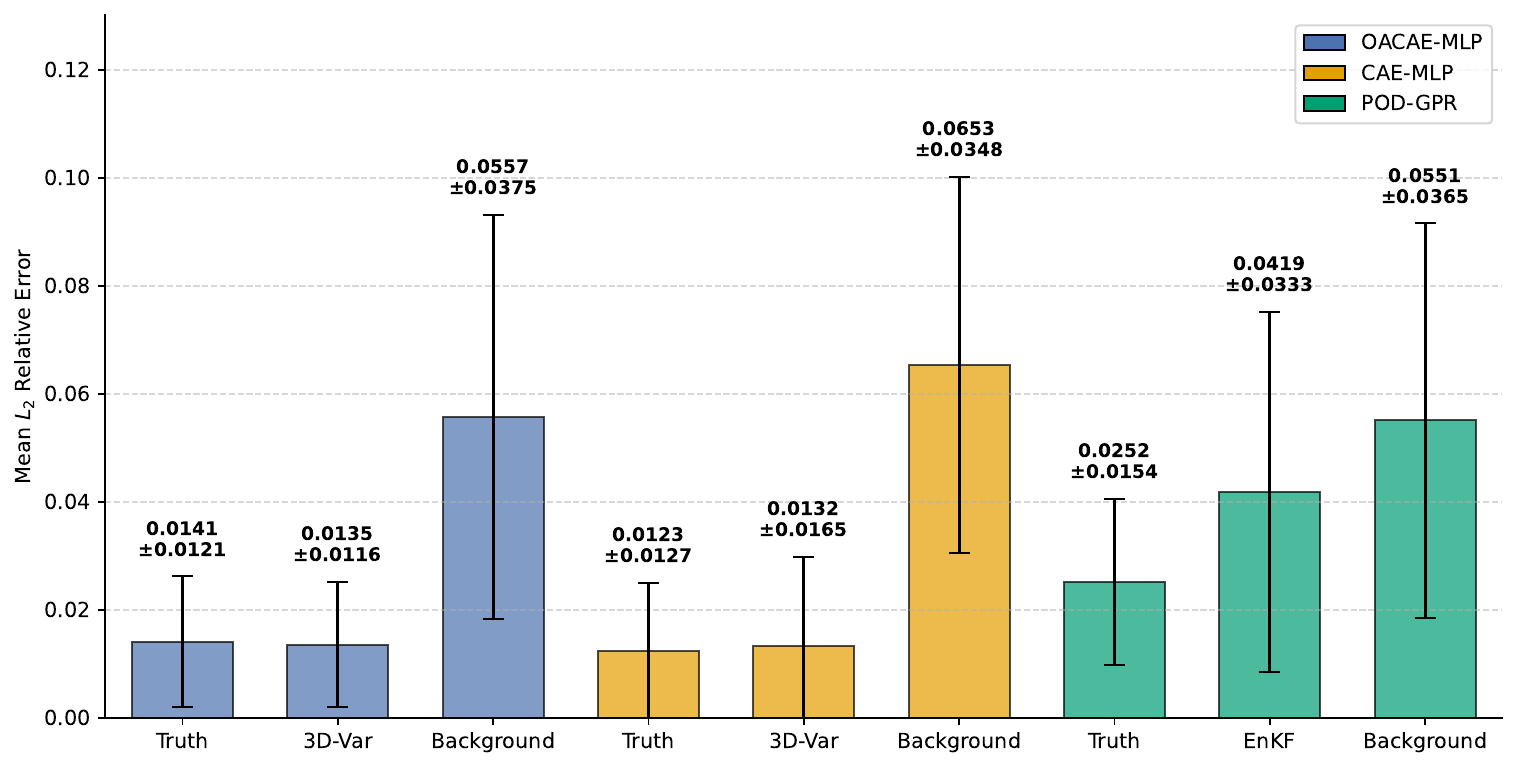}
    \caption{Case-to-case physical-space prediction errors (relative $\ell_2$ error) for the dam flow problem. All data assimilation frameworks reduce both the prediction error and its uncertainty compared with the background. In particular, the variational DA-DL-ROM approaches yield lower average errors and reduced uncertainty than the EnKF-POD-GPR baseline. Moreover, the 3D-Var-OACAE-MLP framework achieves even lower prediction error and reduced uncertainty than that obtained using the true parameters.}
    \label{fig:case-to-case-physics-error}
\end{figure}

While the parameter-space statistics provide a direct measure of calibration accuracy, it is still necessary to examine whether the observed case-to-case robustness is preserved in the physical space. To this end, Figure~\ref{fig:case-to-case-physics-error} reports the physical-space prediction errors obtained using the calibrated parameters for each evolution case.
For each method, the calibrated parameters are used to generate predictions over the entire temporal evolution of that case. The resulting prediction errors are then computed and averaged over time in the physical space and summarized using their mean and standard deviation across all 44 evolution cases.

Across the tested evolution cases, consistent trends are observed. In particular, the variational DA--DL--ROM frameworks continue to provide more stable parameter estimates and improved physical-space predictions compared with the EnKF-POD-GPR baseline. Moreover, the OACAE-MLP surrogate exhibits greater robustness than the CAE-MLP surrogate. These results indicate that the main conclusions drawn above are not sensitive to the choice of test case.

\subsection{Robustness under degraded-observation settings}
\label{sec:degraded_observations}

To evaluate the robustness of the proposed framework for inverse problem under more realistic measurement
conditions, we consider four degraded-observation scenarios for the dam flow example: noisy observations,
low-resolution observations, randomly masked observations, and block-wise partial
observations. These scenarios represent cases where the available observations are
corrupted, spatially coarsened, or incomplete. 

For the noisy-observation experiment, additive Gaussian noise is applied to the full-field
observation. For each sample, a random noise field \(\varepsilon\) is generated and rescaled
to satisfy
\[
\frac{\|\varepsilon\|_2}{\|y\|_2}=\eta,
\]
where
\[
\eta \in \{0,0.01,0.05,0.10,0.20\}
\]
denotes the prescribed relative noise level. The noisy observation is then defined as
\[
y^{\eta}=y+\varepsilon.
\]

For the low-resolution observation experiment, the full-resolution field is downsampled
using average pooling over non-overlapping spatial blocks. Given a downsampling factor
\(s\), the low-resolution observation is defined as
\[
y^{(s)}=P_s(y),
\]
where \(P_s\) denotes the average-pooling operator with kernel size and stride equal to
\(s\). We consider
\[
s\in\{1,2,4,8,16\},
\]
corresponding to observation resolutions
\[
64\times64,\quad 32\times32,\quad 16\times16,\quad 8\times8,\quad 4\times4,
\]
respectively. Note that \(s=1\) corresponds to the original full-field observation.

For the randomly masked-observation experiment, binary spatial masks are applied to the
full-resolution field. The mask is defined over spatial locations and is shared by both
physical channels, so that either both velocity components are observed or both are masked
at a given grid point. Let \(M_{\rho}\in\{0,1\}^{H\times W}\) denote a binary spatial mask,
where \(\rho\) is the masked ratio. The partial observation is written as
\[
y^{\rho}=M_{\rho}(y),
\]
where \(M_{\rho}\) extracts the observed entries from \(y\) according to the mask. We use
\[
\rho\in
\left\{
0,\frac{1}{8},\frac{3}{8},\frac{5}{8},\frac{7}{8}
\right\},
\]
corresponding to observed ratios
\[
100\%,\quad 87.5\%,\quad 62.5\%,\quad 37.5\%,\quad 12.5\%.
\]
The random masks are generated with a fixed seed and are nested, so that higher masked
ratios remove a superset of the grid points removed at lower masked ratios.

For the block-wise partial observation experiment, the full resolution spatial domain is
divided into four non-overlapping blocks of equal size using a \(2\times2\) partition. At
each time snapshot, only one block is retained as the observation, while the remaining
three blocks are unobserved. The same block mask is applied to both physical channels.
Specifically, we consider
\[
y^b=B_b(y),\qquad b\in\{A,B,C,D\},
\]
where \(B_b\) extracts the entries of \(y\) located in block \(b\). The four blocks correspond
to the lower-left, lower-right, upper-left, and upper-right quadrants of the spatial domain,
respectively. In addition, the full-field observation is included as a reference case.
Therefore, the tested block-observation cases are
\[
\mathrm{Full},\quad A,\quad B,\quad C,\quad D.
\]

\begin{figure}[htbp]
\centering
\includegraphics[width=0.95\textwidth]{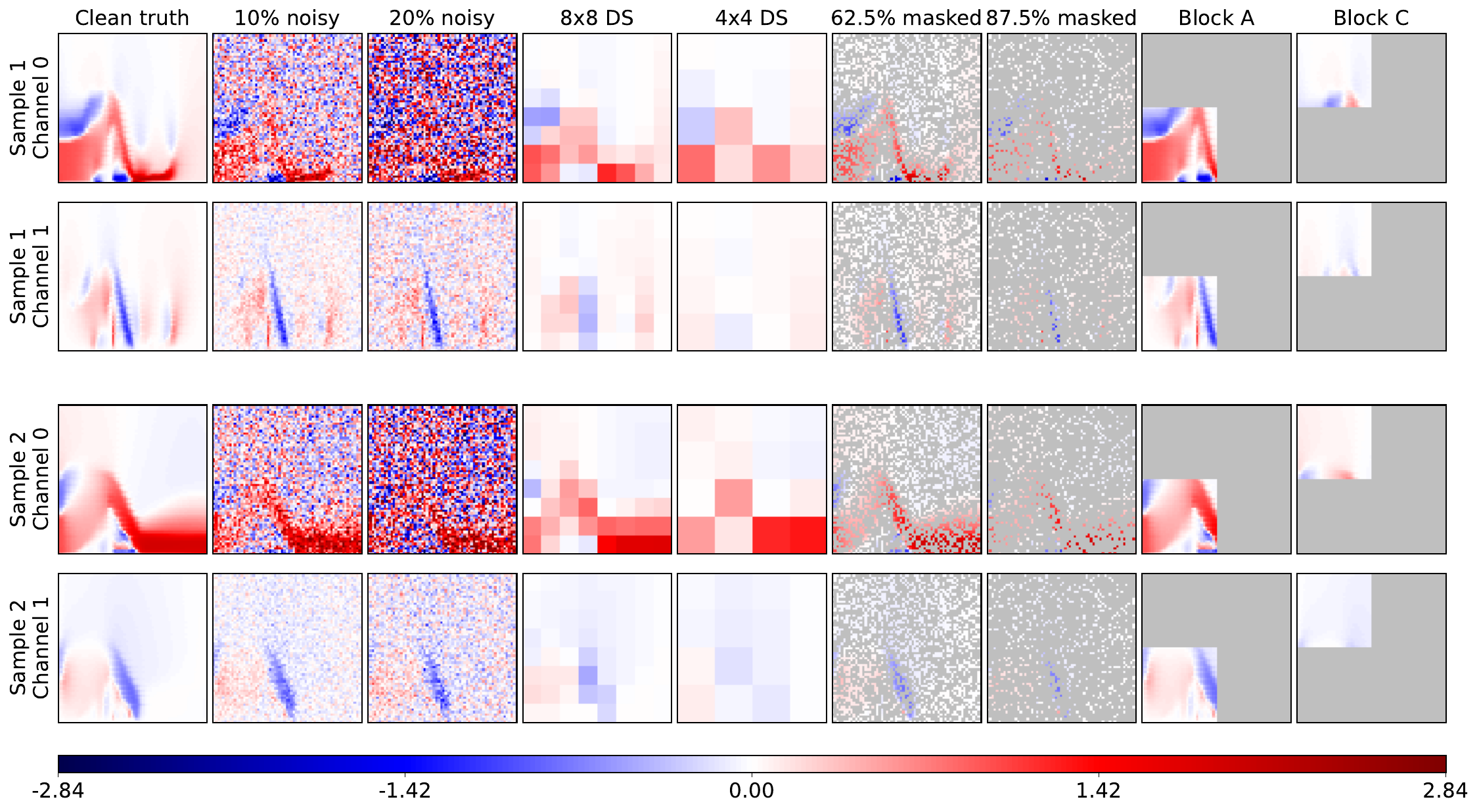}
\caption{Visualization of different observation settings, including clean full-field observations,
noisy observations, low-resolution observations, random masked observations, and partial
observations.}
\label{fig:degraded_observation_overview}
\end{figure}

Figure~\ref{fig:degraded_observation_overview} shows representative examples of the
different observation settings, including clean full-field observations, noisy observations,
low-resolution observations, randomly masked observations, and block-wise partial
observations.

For each degraded-observation setting, the observation operator in the variational cost
function is modified consistently with the available observation. More precisely, let
\(H(\theta)\) denote the full-field surrogate prediction obtained from CAE-MLP or
OACAE-MLP. For a given observation transformation \(G\), the observation term is
evaluated as
\[
J_o(\theta)
=
\frac{1}{2}
\left(G(y)-G(H(\theta))\right)^T
R_G^{-1}
\left(G(y)-G(H(\theta))\right),
\]
where \(G\) is the identity operator for full-field and noisy observations, the average-pooling
operator \(P_s\) for low-resolution observations, the masking operator \(M_{\rho}\) for
randomly masked observations, and the block operator \(B_b\) for block-wise partial
observations. The covariance matrix \(R_G\) is defined in the corresponding observation
space, with a dimension consistent with the number of observed entries.

\begin{figure}[htbp]
    \centering

    \begin{subfigure}[t]{0.48\textwidth}
        \centering
        \includegraphics[width=\textwidth]{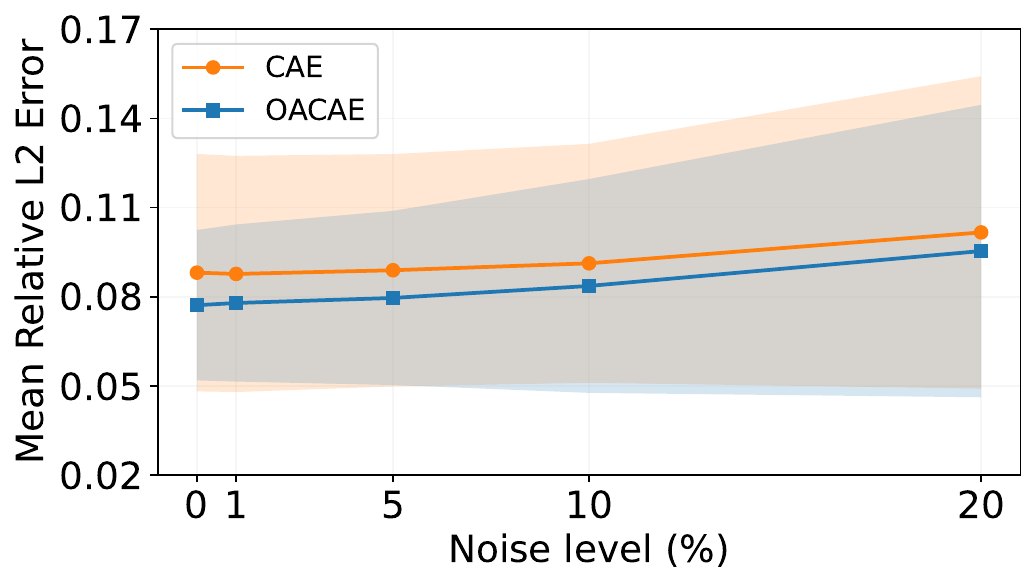}
        \caption{Noisy observations}
        \label{fig:noisy_observation_error}
    \end{subfigure}
    \hfill
    \begin{subfigure}[t]{0.48\textwidth}
        \centering
        \includegraphics[width=\textwidth]{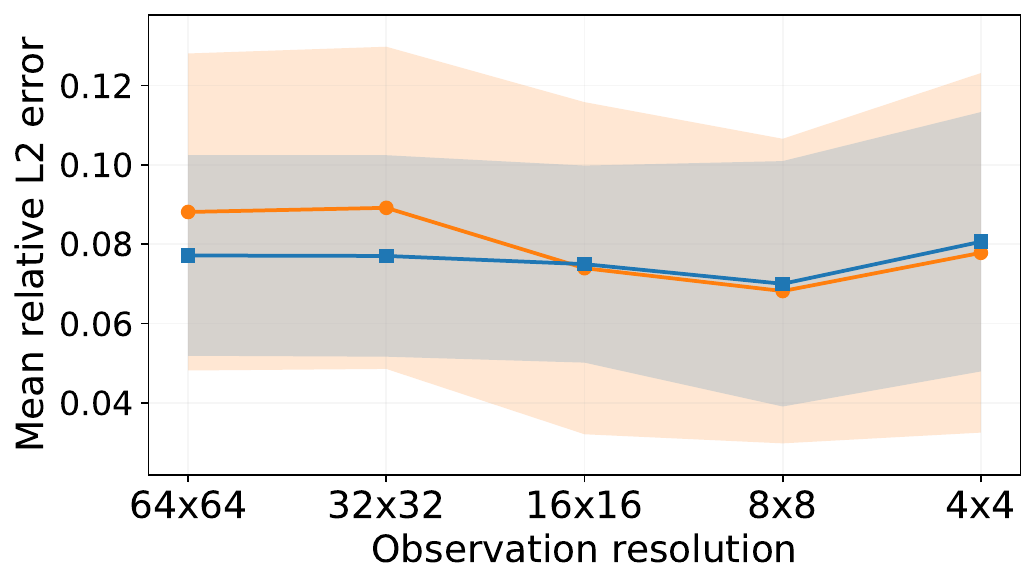}
        \caption{Low-resolution observations}
        \label{fig:low_resolution_observation_error}
    \end{subfigure}

    \vspace{0.3cm}

    \begin{subfigure}[t]{0.48\textwidth}
        \centering
        \includegraphics[width=\textwidth]{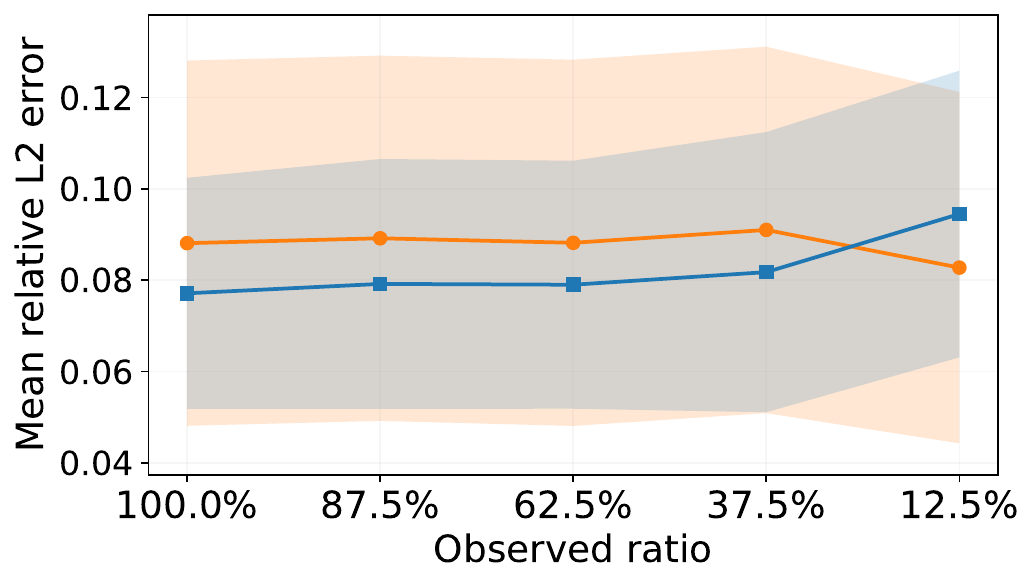}
        \caption{Masked observations}
        \label{fig:partial_observation_error}
    \end{subfigure}
    \hfill
    \begin{subfigure}[t]{0.48\textwidth}
        \centering
        \includegraphics[width=\textwidth]{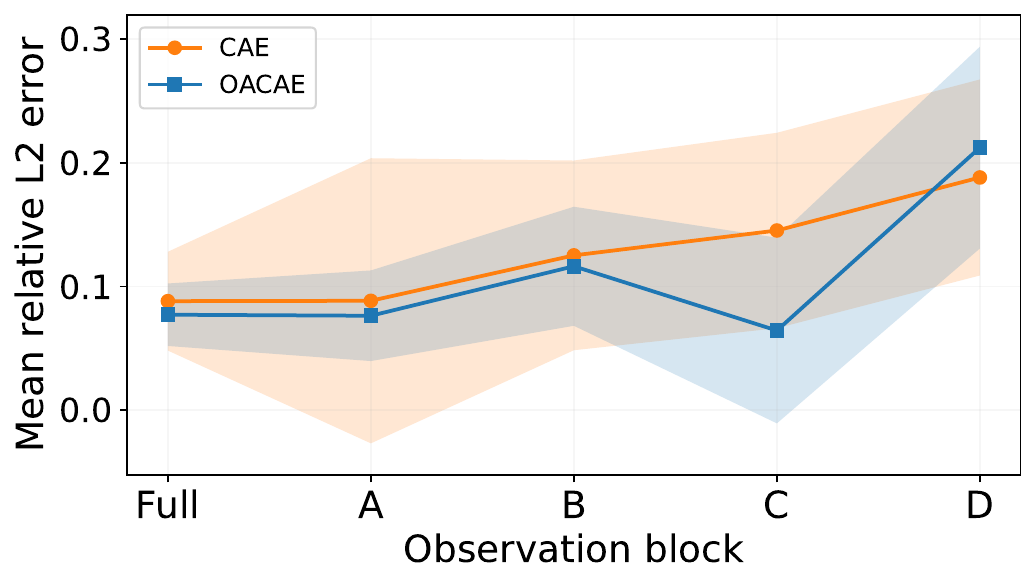}
        \caption{Block observations}
        \label{fig:block_observation_error}
    \end{subfigure}

    \caption{
    Parameter calibration performances under degraded-observation settings. For each setting, inverse modeling performances of CAE-MLP and OACAE-MLP are compared using 3D-Var. The curves report the mean error over multiple observations, and the shaded regions indicate standard deviation.
    }
    \label{fig:degraded_observation_results}
\end{figure}

Figure~\ref{fig:degraded_observation_results} reports the resulting parameter-calibration
errors under the four degraded-observation scenarios. In most
degraded-observation settings, the OACAE-MLP framework achieves lower calibration
errors than the CAE-MLP framework. Exceptions are observed in several highly degraded
cases, including low-resolution observations at \(16\times16\), \(8\times8\), and
\(4\times4\), randomly masked observations with only \(12.5\%\) observed entries, and the
block-observation setting \(D\). In these cases, the difference between CAE-MLP and
OACAE-MLP becomes less pronounced, which is expected when the available observations
contain limited parameter-sensitive information. More importantly, the OACAE-MLP framework exhibits lower variability across all observation settings, as indicated by the smaller standard deviations. Overall, both CAE-MLP and OACAE-MLP show robustness under
degraded-observation settings, demonstrating the stability of the proposed DL-ROM-DA
framework itself. Within this robust framework, OACAE-MLP further improves the
calibration accuracy and reduces the variability of the inverse estimates.

\section{Conclusion and Future work}
\label{sec:Conclusion}
This work aims to address the challenge of forward prediction and inverse parameter calibration in high-dimensional parametric dynamical systems. This requires a reduced-order surrogate that not only accurately reconstructs or predicts flow fields but also preserves parameter-relevant information essential for inverse modeling. To this end, we developed an end-to-end differentiable physics-aware latent-space framework that combines observable-augmented autoencoder-based reduced-order modeling with variational data assimilation. 

The numerical results on the CFDbench benchmark show that reconstruction accuracy alone is not sufficient to determine the quality of a surrogate model for inverse problems. Although POD achieves the best reconstruction accuracy, its reconstruction-optimal projection basis may not preserve the parameter-informative directions needed for inverse calibration. In contrast, the observable-augmented autoencoder slightly sacrifices pure reconstruction accuracy, but provides a latent representation that is better aligned with the physical quantities involved in the inverse problem. More precisely, when integrated into a variational data assimilation framework, the observable-augmented model yields low physical-space prediction errors and reduced variability across calibration instances. These advantages persist under noisy, low-resolution, randomly masked, and block-wise partial observations. 

Since the present method relies on a deterministic surrogate, the reported parameter errors reflect the combined effects of the potential ill-posedness of the inverse problem, observation uncertainty, and surrogate-model discrepancy. A complete decomposition of surrogate-induced bias and a systematic uncertainty quantification of the calibrated parameters are therefore left for future work \cite{Zhang_2020, Reiser_2025}. Moreover, since the numerical experiments mainly assess model performance within the available data distribution, reliable extrapolation is generally not guaranteed. Future work will address these limitations by extending the framework to more realistic experimental data and by developing uncertainty-aware and probabilistic variants of the proposed framework \cite{guo2026parametric, pmlr-v9-titsias10a}. Potential directions include probabilistic manifold learning, Bayesian reduced-order modeling, and ensemble-based data-assimilation strategies. Further applications to higher-Reynolds-number turbulent flows, higher-resolution simulations, structural mechanics \cite{SimoneBrivio2025, li2024mechanicsinformedautoencoderenablesautomated}, and fluid-structure interaction problems will also be considered. Finally, quantum neural networks \cite{xiao2024physics, li2025quantummachinelearningefficient, farea2025qcpinnquantumclassicalphysicsinformedneural} and hybrid quantum--classical learning architectures may be investigated as a longer-term direction for physics-informed reduced-order modeling and inverse problems, although their integration into the present data-assimilation framework would require a dedicated methodological study.


\section*{Data and code availability}
The code scripts used in this paper is available at \url{https://github.com/qiyaozhou963/rom-nn-da}. The numerical experiments are implemented using the CFDbench dataset, available at \url{https://github.com/luo-yining/CFDBench}.

\section*{Author contributions}
Qiyao Zhou: Methodology, Software, Validation, Formal analysis, Investigation, Writing - Original Draft, Visualization
\vskip\baselineskip
Xujia Zhu: Conceptualization, Methodology, Formal analysis, Investigation,  Writing - Review and Editing, Visualization
\vskip\baselineskip
Pierre Joli: Resources, Supervision, Formal analysis, Writing - Review and Editing
\vskip\baselineskip
Yu Cong: Resources, Supervision, Writing - Review and Editing
\vskip\baselineskip
Sibo Cheng: Project administration, Methodology, Software, Resources, Supervision, Formal analysis, Writing - Review and Editing

\section*{Acknowledgement}
Qiyao gratefully acknowledges the UnivEvry, Université Paris-Saclay, for support through a Ph.D. scholarship.
Sibo Cheng acknowledges the support of the French Agence Nationale de la Recherche (ANR) under reference ANR-22-CPJ2-0143-01. 

\bibliographystyle{ieeetr}
\bibliography{Bibliography}
\nocite{*}

\newpage

\appendix
\section{Gaussian process regression of POD coefficients}  
\label{appendix: GPR}

Once the POD basis $\mathbf{L}_{\mathcal{X},q}$ is computed and the training snapshots are projected onto the reduced space, the corresponding POD coefficients are obtained for each training parameter instance. Collecting these reduced representations over the training set yields the POD coefficient matrix
\[
\hat{\mathbf{x}} = 
\Big[\{k_{j}(\boldsymbol{\theta}^{(0)})\}_{j=1}^{q}, \; \ldots, \; \{k_{j}(\boldsymbol{\theta}^{(N_{\text{train}}-1)})\}_{j=1}^{q}\Big] 
\in \mathbb{R}^{q \times N_{\text{train}}},
\]
which serves as the regression target for the subsequent surrogate modeling.
Following the common assumption that POD coefficients are approximately decorrelated, the regression task is split into $q$ independent scalar problems, one for each mode.  

We adopt a Gaussian Process Regression (GPR) model for each coefficient $k_j$, with an additive Gaussian noise\cite{lumet2025POD-GPR}:  
\begin{equation}
k_{j}(\boldsymbol{\theta}) = f_{j}(\boldsymbol{\theta}) + \varepsilon_{j}, \quad f_j \sim \mathcal{GP}\big(0,\, r_j\big), 
\quad \varepsilon_{j} \sim \mathcal{N}(0, s_{j}^{2}),
\end{equation}
where $r_j$ is the covariance function defined on $\Omega_{\theta}^2$ and $s_j^2$ is the noise variance.  
Given training samples $\{\boldsymbol{\theta}^{(i)}, k_j^{(i)}\}_{i=1}^{N_{train}}$, 
the posterior distribution of the predicted POD coefficient $k_j^*(\boldsymbol{\theta}^*)$ at a new parameter $\boldsymbol{\theta}^*$, 
conditional on the training set $\{\Theta^{train}, \mathbf{K}_j^{train}\}=\{(\boldsymbol{\theta}^{(i)}, k_j^{(i)})\}_{i=1}^{N_{train}}$, is
\begin{equation}
k_j^*(\boldsymbol{\theta}^*) \;\big|\;\{\Theta^{train},\mathbf{K}_j^{train}\}
\;\sim\;
\mathcal{N}\!\big(\mu_j,\, \sigma^2_{\mathrm{GP}}(k_j^*)\big),
\end{equation}
with
\begin{subequations}
\begin{align}
\mu_j &= r_j(\boldsymbol{\theta}^*, \Theta^{train})
\Big[\, r_j(\Theta^{train}, \Theta^{train}) + s_j^2 \mathbf{I}\,\Big]^{-1}
\mathbf{K}_j^{train}, \\[6pt]
\sigma^2_{\mathrm{GP}}(k_j^*) &= r_j(\boldsymbol{\theta}^*, \boldsymbol{\theta}^*) + s_j^2 \notag\\
&\quad - r_j(\boldsymbol{\theta}^*, \Theta^{train})
\Big[\, r_j(\Theta^{train}, \Theta^{train}) + s_j^2 \mathbf{I}\,\Big]^{-1}
r_j(\Theta^{train}, \boldsymbol{\theta}^*). 
\end{align}
\end{subequations}

For regression, we train $q$ independent Gaussian processes using a multi-output wrapper that enables parallel computation. This implementation directly adopts the code architecture proposed in \cite{lumet2025POD-GPR}. Each GP employs a radial basis function kernel of the form
\[
r_j(\boldsymbol{\theta},\boldsymbol{\theta}') 
= \rho_j \exp\!\Big(-\tfrac12\|\boldsymbol{\theta}-\boldsymbol{\theta}'\|^2_{\Lambda_j}\Big),
\]
where $\Lambda_j=\mathrm{diag}(\lambda_{j,1}^{-2},\ldots,\lambda_{j,d_{\theta}}^{-2})$ contains 
the characteristic length-scales associated with each parameter dimension, 
and $\rho_j$ denotes the maximum allowable covariance. 
For convenience, we adopt the notation $\|\mathbf{a}\|_{\mathbf{A}}^2 = \mathbf{a}^\top \mathbf{A}^{-1}\mathbf{a}$.  

\begin{figure}[H]
    \centering
    \includegraphics[width=0.8\textwidth]{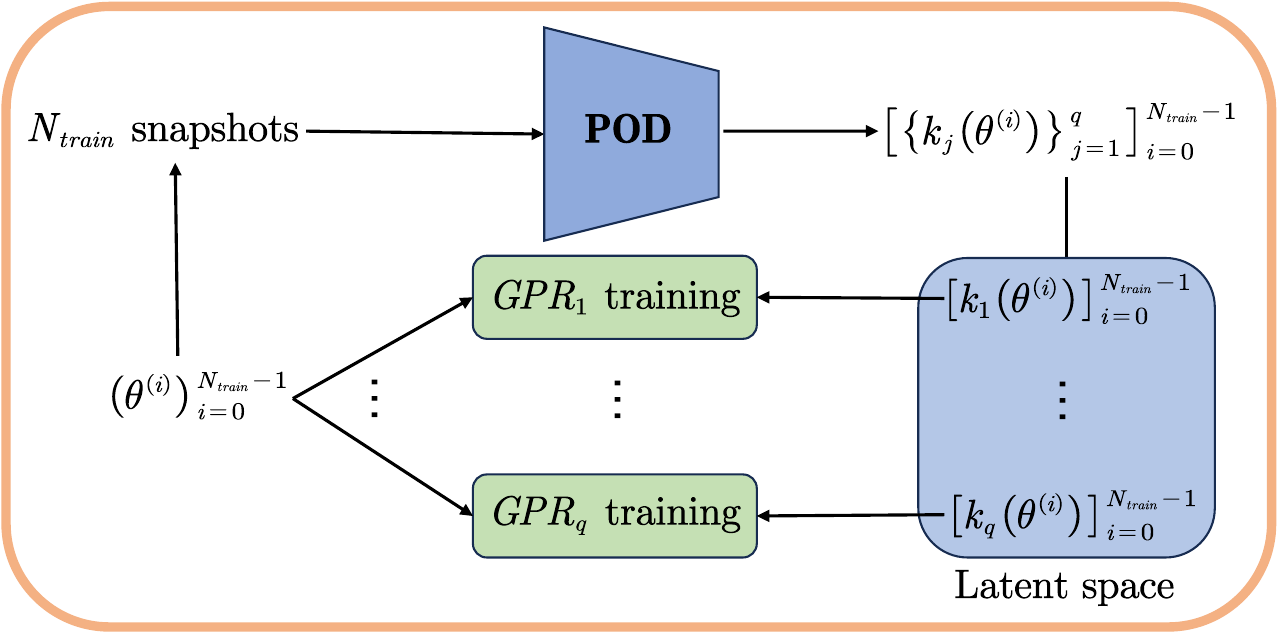}

    \vspace{0.5em}

    \includegraphics[width=0.8\textwidth]{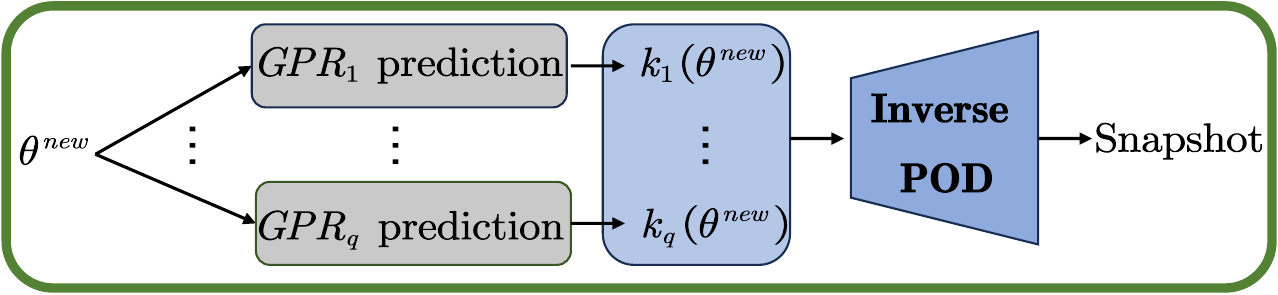}

    \caption{}
    \label{fig:POD-GPR}
\end{figure}

The GP predictive variance reflects two distinct sources of uncertainty:  
(i) observation noise in the training data, represented by the term $s_j^2 \mathbf{I}$, and  
(ii) regression uncertainty, which increases with the distance between new input parameters 
$\boldsymbol{\theta}^*$ and the training set $\Theta^{\mathrm{train}}$.  
The set of hyperparameters for each GP is given by
\[
\gamma_j = \big(\rho_j,\, s_j,\, \lambda_{j,1},\ldots,\lambda_{j,d_{\theta}}\big) \in \mathbb{R}_+^{d_{\theta}+2},
\]
and is determined by maximizing the log-marginal likelihood (MLL) with multiple random restarts. Finally, the posterior predictions of the reduced coordinates
\[
\hat{\mathbf{x}}^* = 
\Big[\{k_{j}^*(\boldsymbol{\theta}^{*(0)})\}_{j=1}^{q}, \; \ldots, \; \{k_{j}^*(\boldsymbol{\theta}^{*(N_{\text{test}}-1)})\}_{j=1}^{q}\Big] 
\in \mathbb{R}^{q \times N_{\text{test}}},
\]
are mapped back to the full physical space through the POD basis computed from the training set, 
thus performing the inverse POD reconstruction to predict the velocity fields corresponding to 
new parameter configurations in the test dataset. For clarity, the complete offline training and online prediction workflows of POD-GPR surrogate are summarized in Fig~\ref{fig:POD-GPR}.

\section{Neural network architectures}  
\label{appendix:nn-structures}

\begin{table}[h!]
\centering
\caption{Neural Network structure of the OACAE encoder/decoder (input shape: 2\,$\times$\,64\,$\times$\,64, output shape: 2\,$\times$\,64\,$\times$\,64).}
\begin{tabular}{ccc} \toprule
\textbf{Layer (type)} & \textbf{Output shape} & \textbf{Activation} \\ \midrule
\multicolumn{3}{l}{\emph{Encoder}} \\
Input & $(2,\,64,\,64)$ & \\
Conv2d $(3\times3)$ & $(16,\,64,\,64)$ & Tanh \\
MaxPooling $(2\times2)$ & $(16,\,32,\,32)$ & \\
Conv2d $(3\times3)$ & $(8,\,32,\,32)$ & Tanh \\
MaxPooling $(2\times2)$ & $(8,\,16,\,16)$ & \\
Conv2d $(3\times3)$ & $(4,\,16,\,16)$ & Tanh \\
MaxPooling $(2\times2)$ & $(4,\,8,\,8)$ & \\
Flatten & $(256)$ & \\
Dense $(256\!\to\!256)$ & $(256)$ & Tanh \\
Dense $(256\!\to\!128)$ & $(128)$ & Tanh \\
Dense $(128\!\to\!64)$ & $(64)$ & Tanh \\
Dense $(64\!\to\!32)$ & $(32)$ & Tanh \\
Latent (Dense $32\!\to\!32$) & $(32)$  \\ \midrule
\multicolumn{3}{l}{\emph{Decoder}} \\
Dense $(32\!\to\!32)$ & $(32)$ & Tanh \\
Dense $(32\!\to\!64)$ & $(64)$ & Tanh \\
Dense $(64\!\to\!128)$ & $(128)$ & Tanh \\
Dense $(128\!\to\!256)$ & $(256)$ & Tanh \\
Dense $(256\!\to\!256)$ & $(256)$ & Tanh \\
Reshape & $(4,\,8,\,8)$ & \\
Conv2d $(3\times3)$ & $(8,\,8,\,8)$ & Tanh \\
ConvTranspose2d $(4\times4)$ & $(16,\,16,\,16)$ & Tanh \\
ConvTranspose2d $(4\times4)$ & $(32,\,32,\,32)$ & Tanh \\
ConvTranspose2d $(4\times4)$ & $(2,\,64,\,64)$  \\
\bottomrule
\end{tabular}
\label{tab:oacae_enc_dec}
\end{table}

\begin{table}[H]
\centering
\caption{Neural Network structure of the observable branch (MLP $\mathcal{E}_a$) used in OACAE training (input shape: 32, output shape: 6).}
\begin{tabular}{ccc} \toprule
\textbf{Layer (type)} & \textbf{Output shape} & \textbf{Activation} \\ \midrule
Input (latent) & $(32)$ & \\
Dense $(32\!\to\!32)$ & $(32)$ & Tanh \\
Dense $(32\!\to\!64)$ & $(64)$ & Tanh \\
Dense $(64\!\to\!32)$ & $(32)$ & Tanh \\
Dense $(32\!\to\!6)$ & $(6)$ \\ \bottomrule
\end{tabular}
\label{tab:oacae_obs_branch}
\end{table}

\begin{table}[H]
\centering
\caption{Neural Network structure of the parameter-to-latent regressor (MLP $\mathcal{E}_b$, input shape: 6, output shape: 32).}
\begin{tabular}{ccc} \toprule
\textbf{Layer (type)} & \textbf{Output shape} & \textbf{Activation} \\ \midrule
Input (parameters) & $(6)$ & \\
Dense $(6 \!\to\! 50)$   & $(50)$   & ReLU \\
Dense $(50 \!\to\! 200)$ & $(200)$  & ReLU \\
Dense $(200 \!\to\! 200)$& $(200)$  & ReLU \\
Dense $(200 \!\to\! 50)$ & $(50)$   & ReLU \\
Dense $(50 \!\to\! 32)$  & $(32)$ \\ \bottomrule
\end{tabular}
\label{tab:oacae_param2latent}
\end{table}

\section{Reconstruction snapshots}  
\label{appendix: Reconstruction snapshots}

\begin{figure}[H]
\centering
\subfloat[truth u]{\includegraphics[width = 1in]{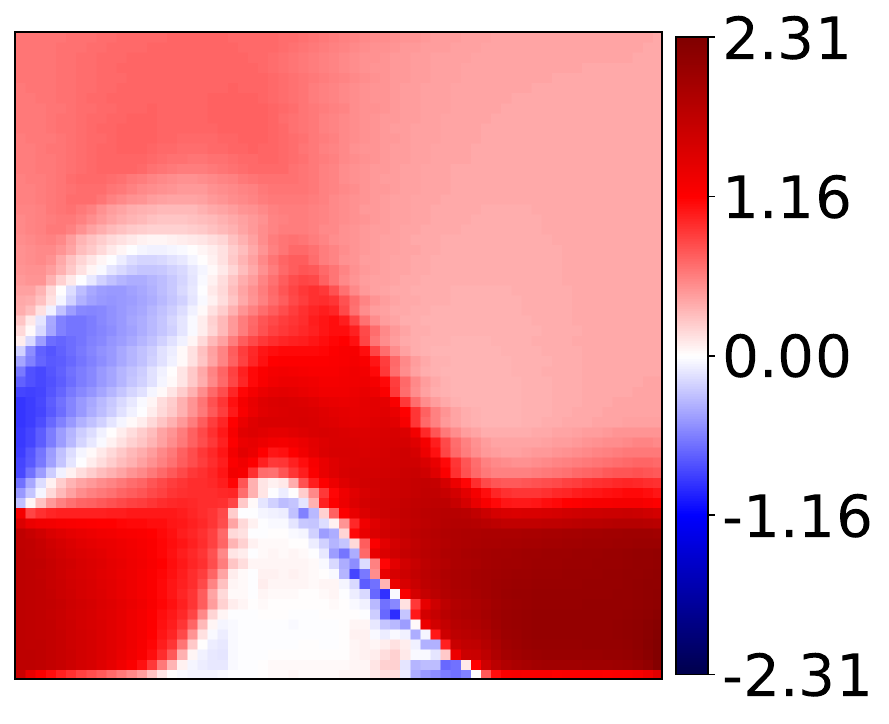}}
\qquad
\subfloat[no error]{\includegraphics[width = 1in]{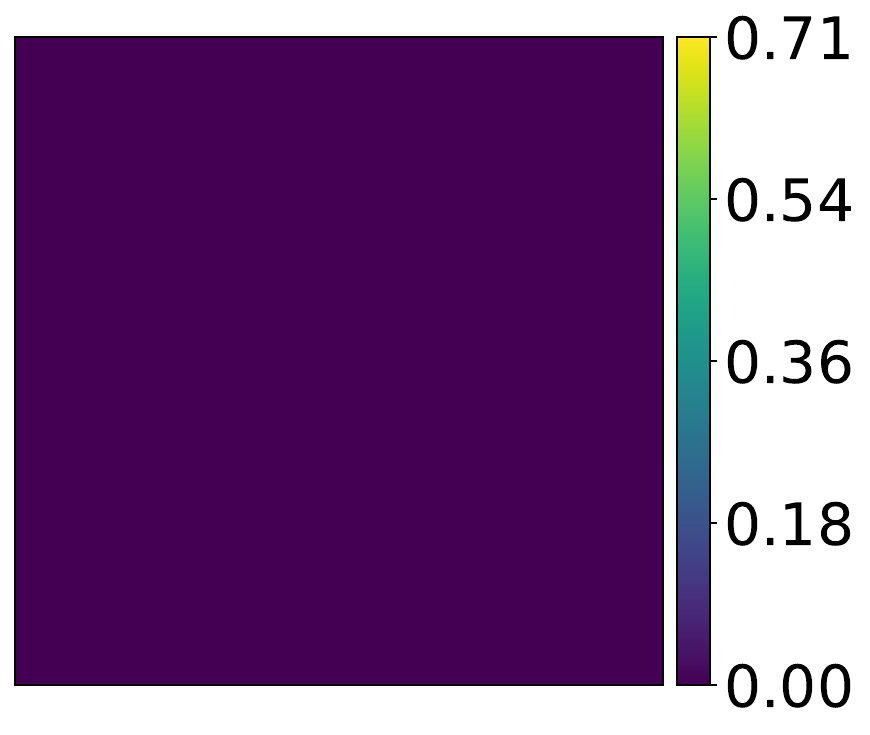}}
\qquad
\subfloat[truth v]{\includegraphics[width = 1in]{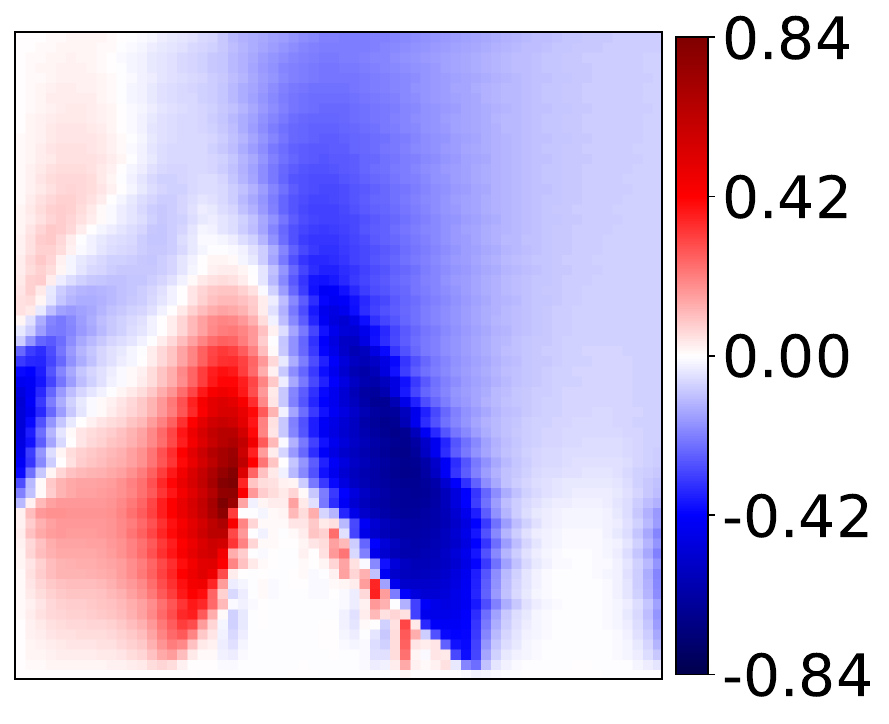}}
\qquad
\subfloat[no error]{\includegraphics[width = 1in]{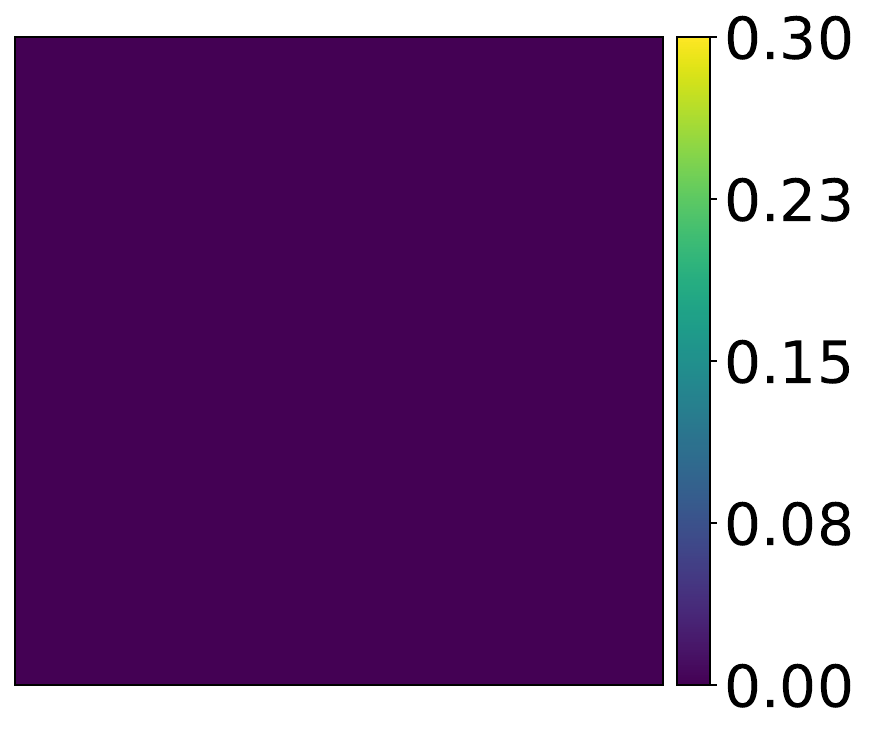}}\\
\subfloat[POD reconstruction]{\includegraphics[width = 1in]{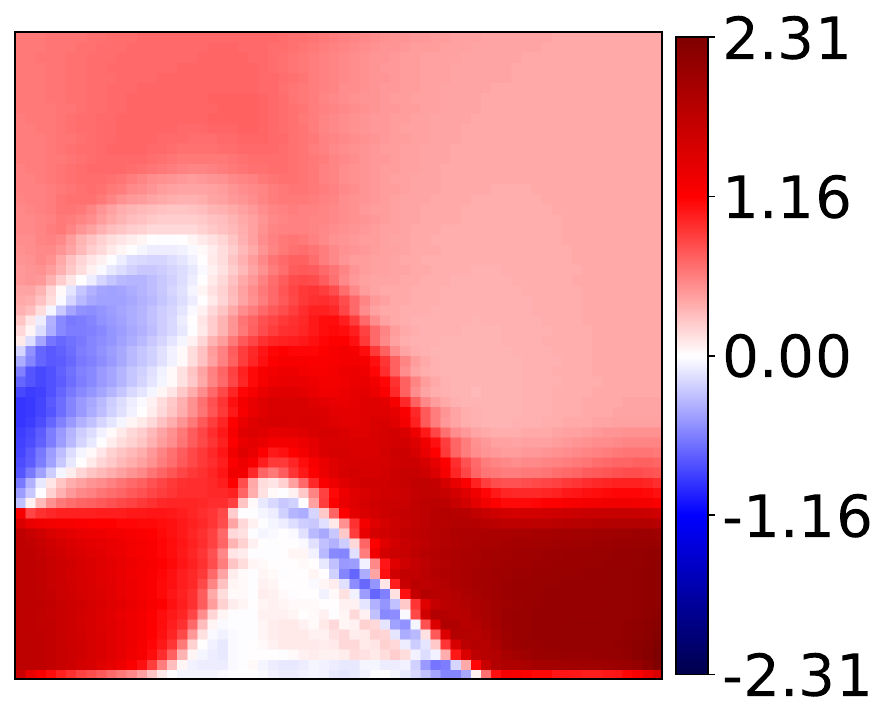}}
\qquad
\subfloat[error u POD]{\includegraphics[width = 1in]{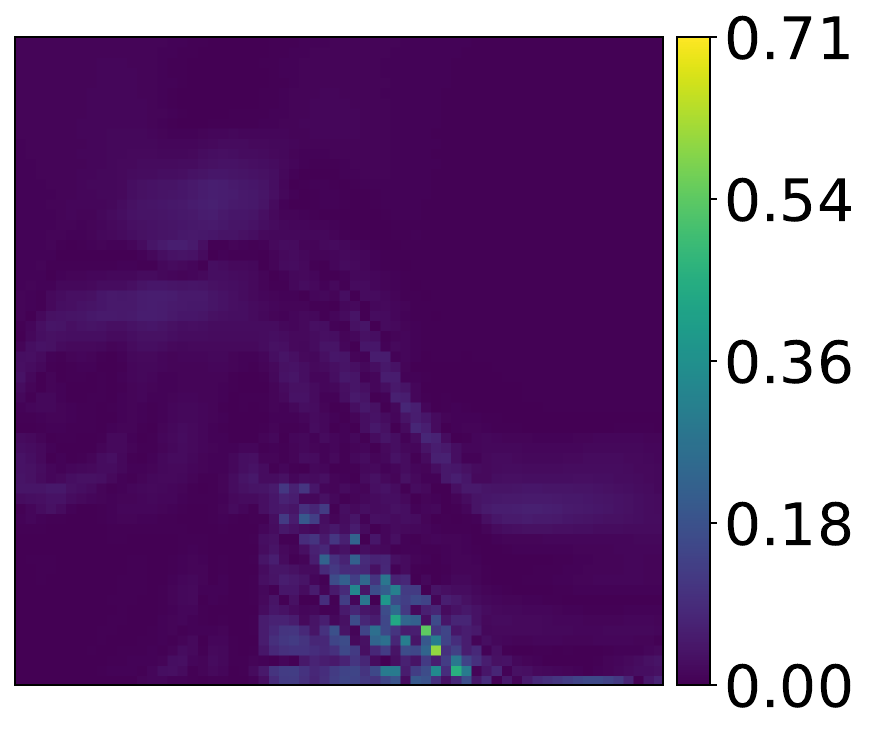}}
\qquad
\subfloat[POD reconstruction]{\includegraphics[width = 1in]{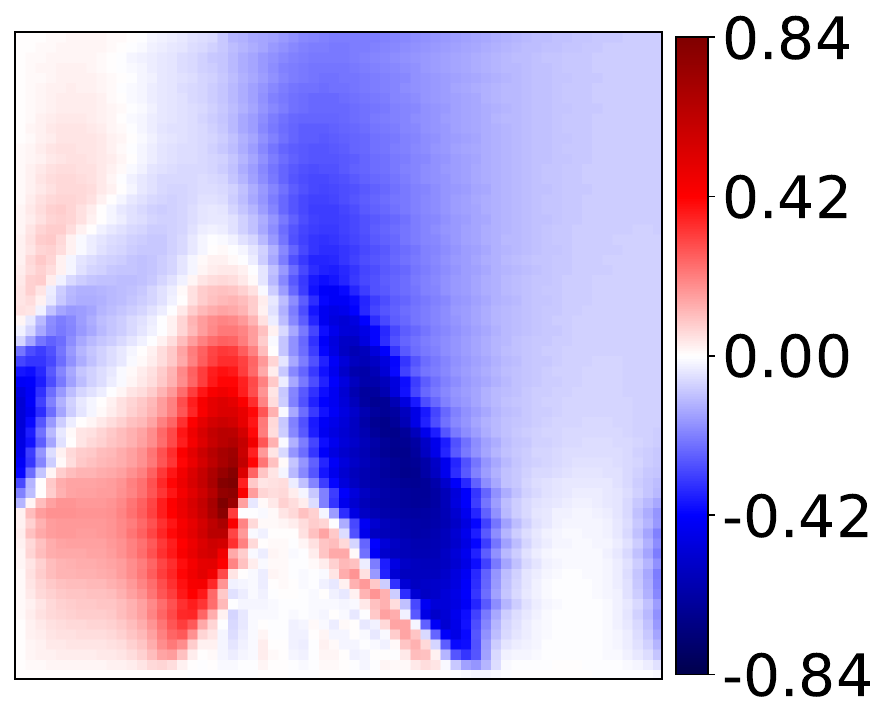}}
\qquad
\subfloat[error v POD]{\includegraphics[width = 1in]{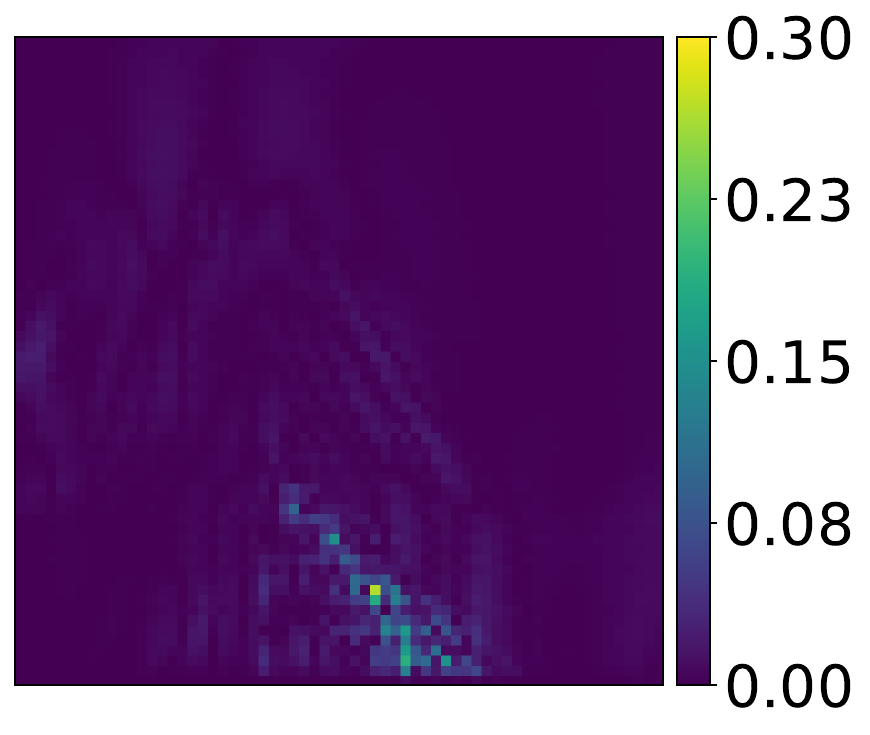}}\\

\subfloat[CAE-MLP reconstruction]{\includegraphics[width = 1in]{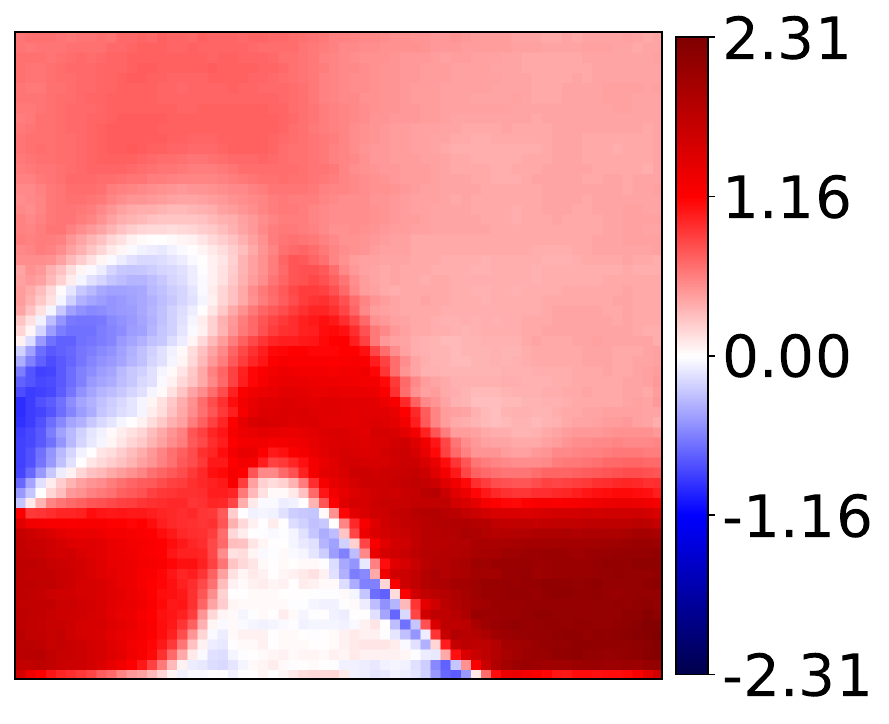}}
\qquad
\subfloat[error u CAE-MLP]{\includegraphics[width = 1in]{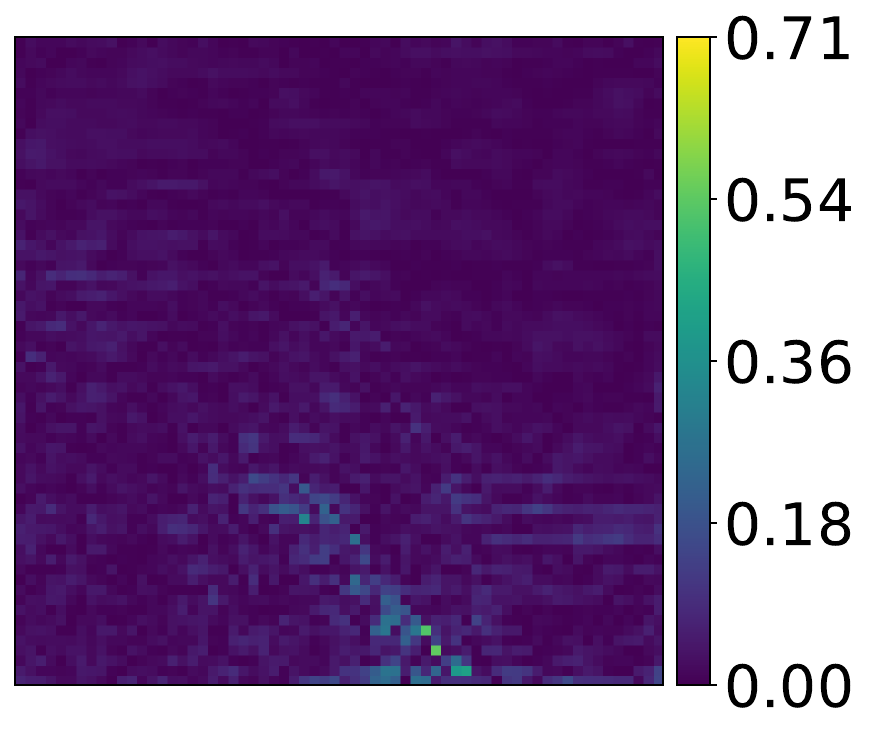}}
\qquad
\subfloat[CAE-MLP reconstruction]{\includegraphics[width = 1in]{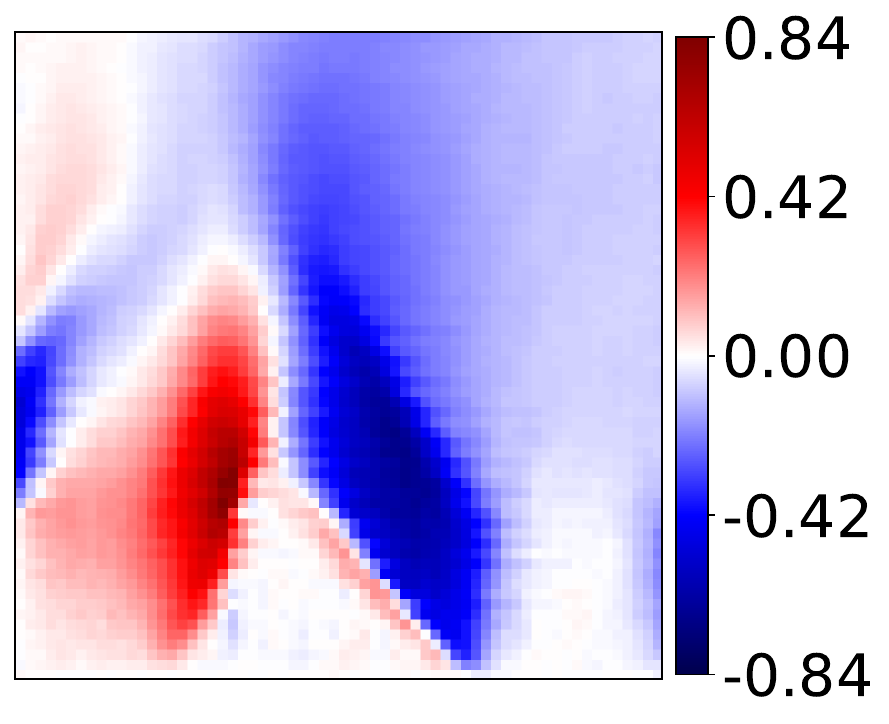}}
\qquad
\subfloat[error v CAE-MLP]{\includegraphics[width = 1in]{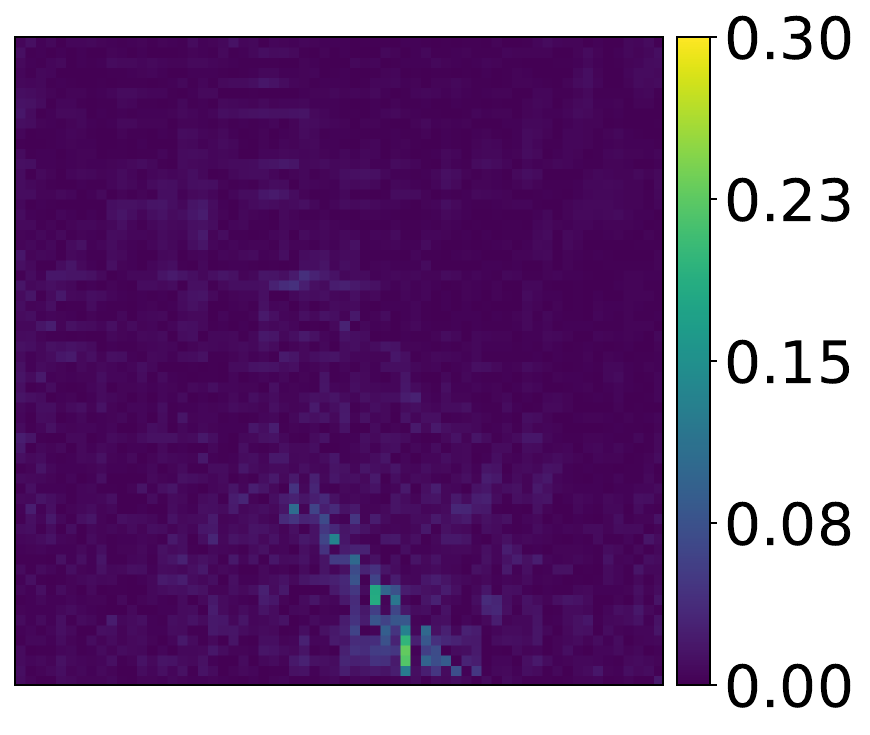}}\\
\subfloat[OACAE-MLP reconstruction]{\includegraphics[width = 1in]{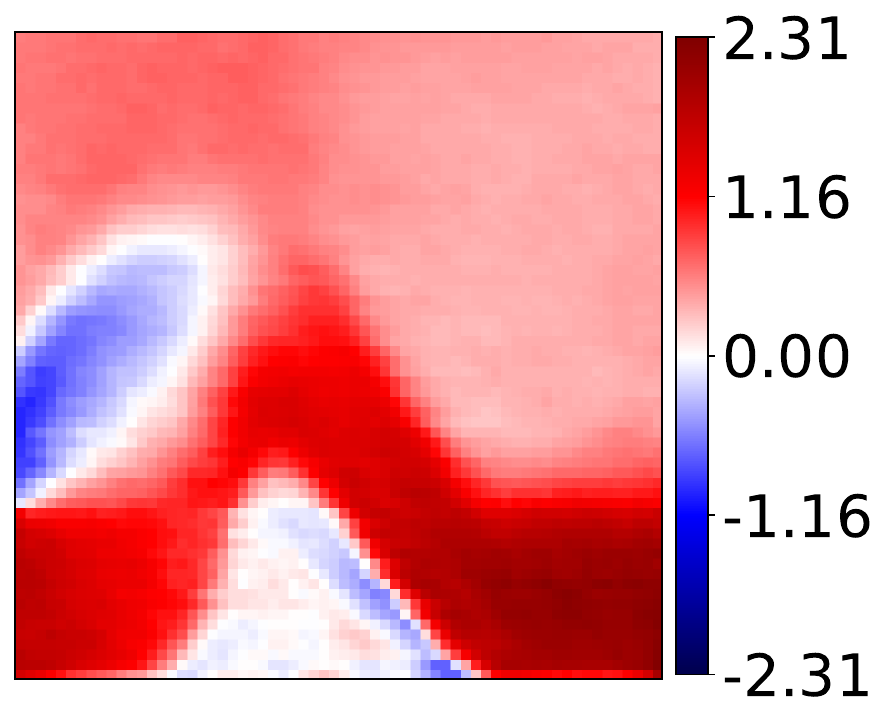}}
\qquad
\subfloat[error u OACAE-MLP]{\includegraphics[width = 1in]{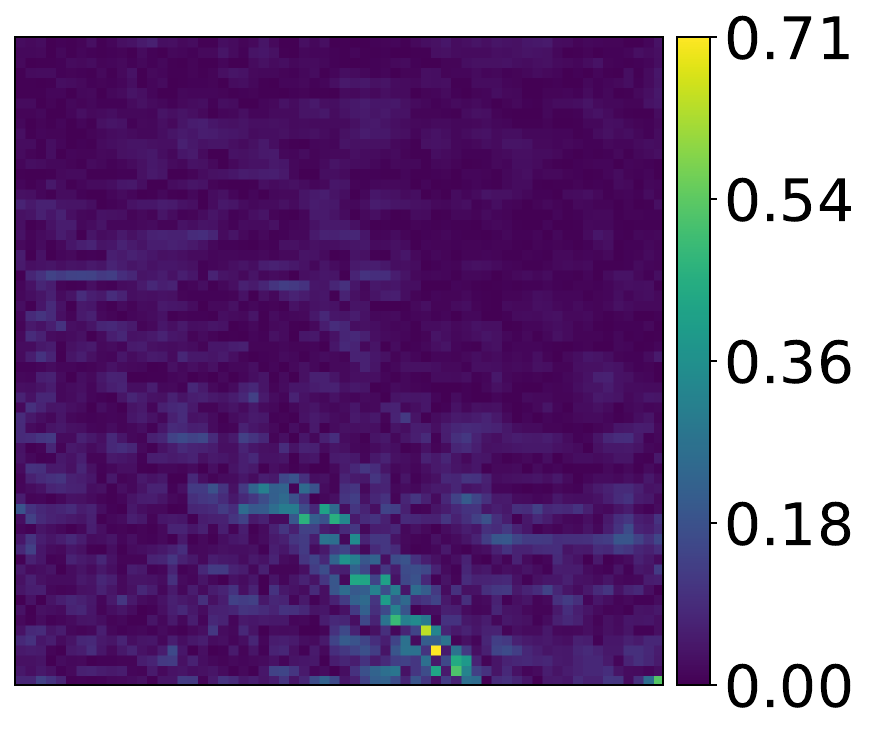}}
\qquad
\subfloat[OACAE-MLP reconstruction]{\includegraphics[width = 1in]{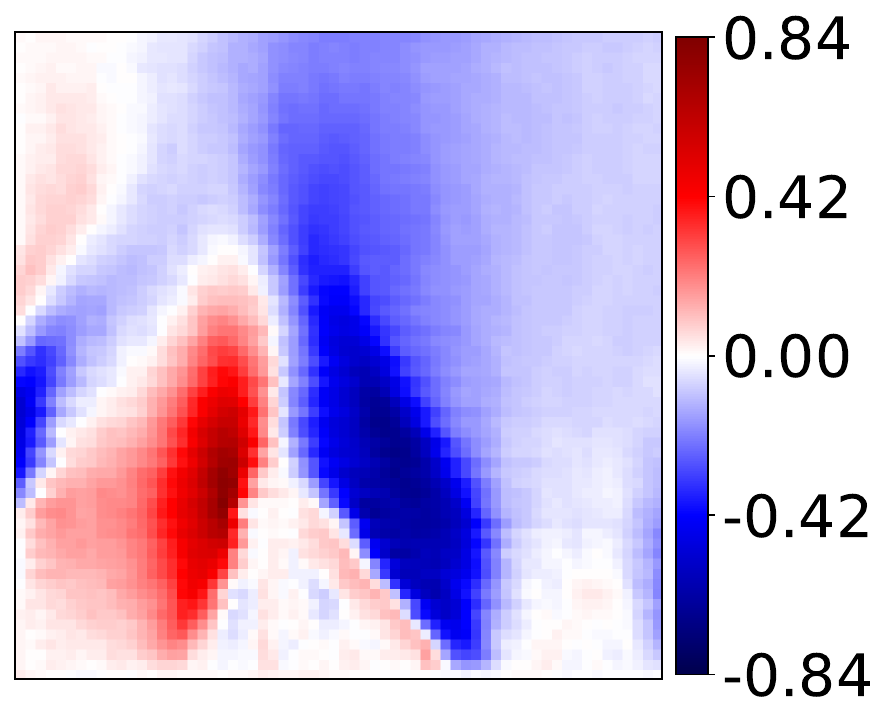}}
\qquad
\subfloat[error v OACAE-MLP]{\includegraphics[width = 1in]{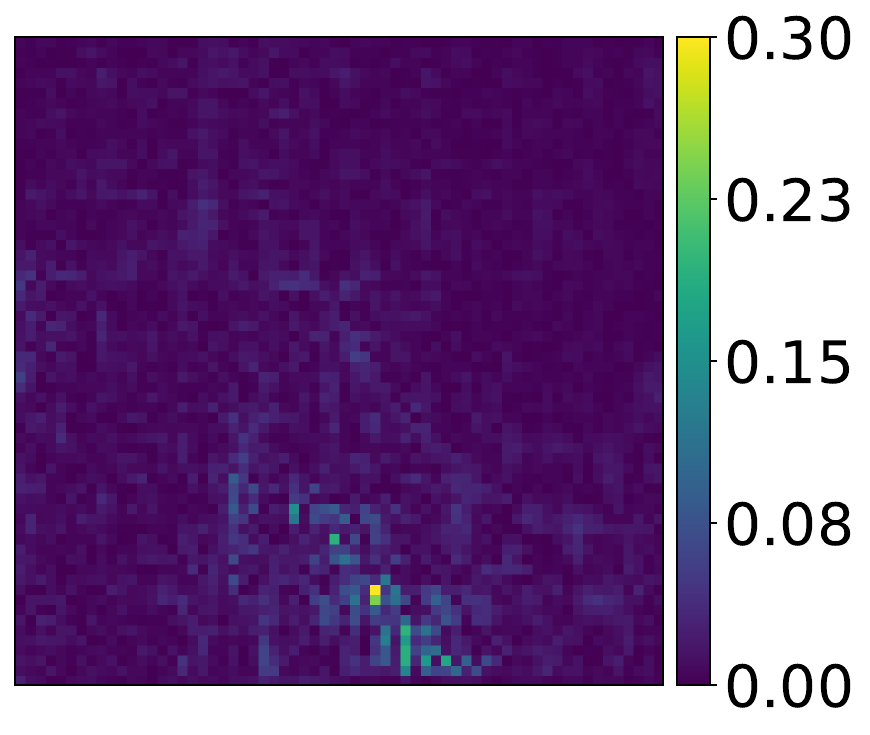}}\\
\subfloat[truth u]{\includegraphics[width = 1in]{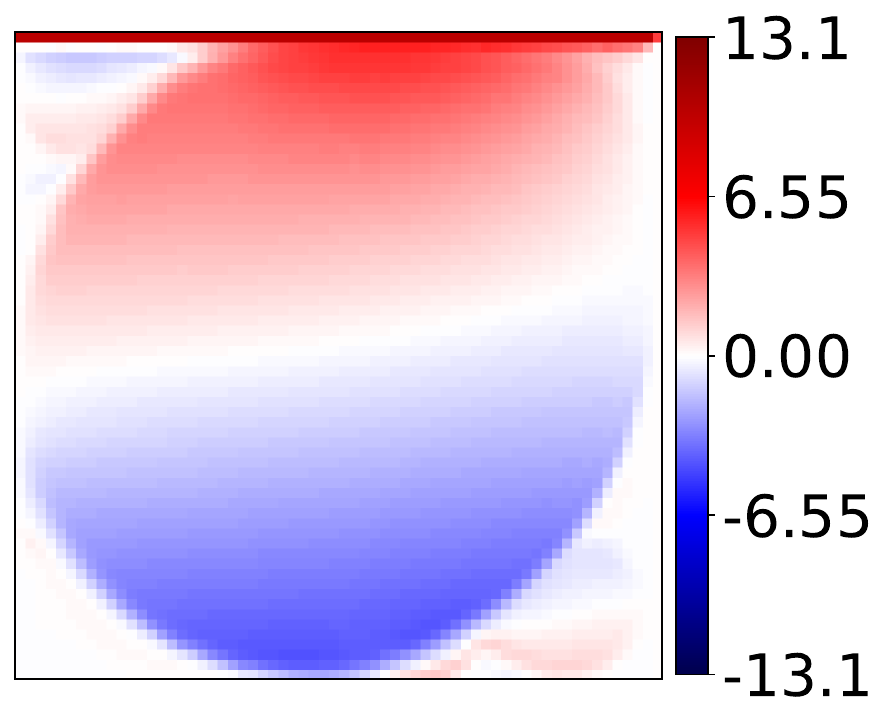}}
\qquad
\subfloat[no error]{\includegraphics[width = 1in]{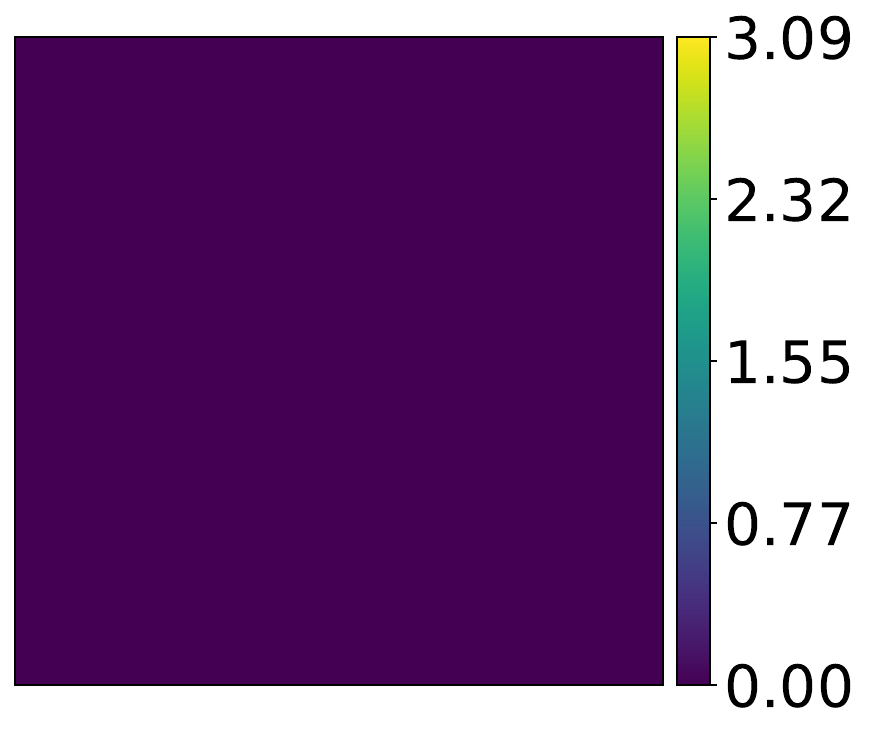}}
\qquad
\subfloat[truth v]{\includegraphics[width = 1in]{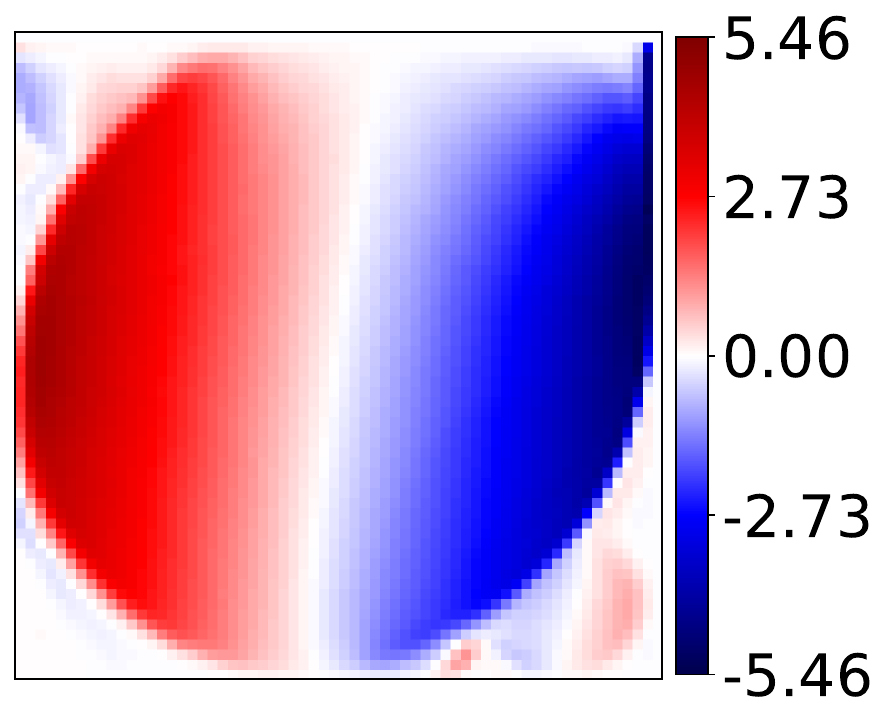}}
\qquad
\subfloat[no error]{\includegraphics[width = 1in]{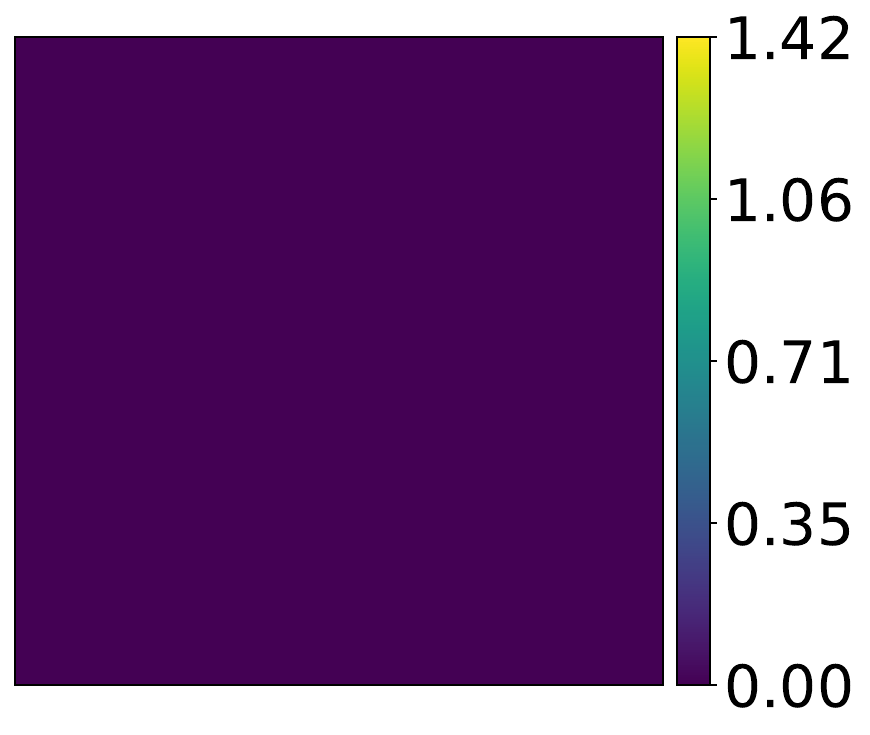}}\\
\subfloat[POD reconstruction]{\includegraphics[width = 1in]{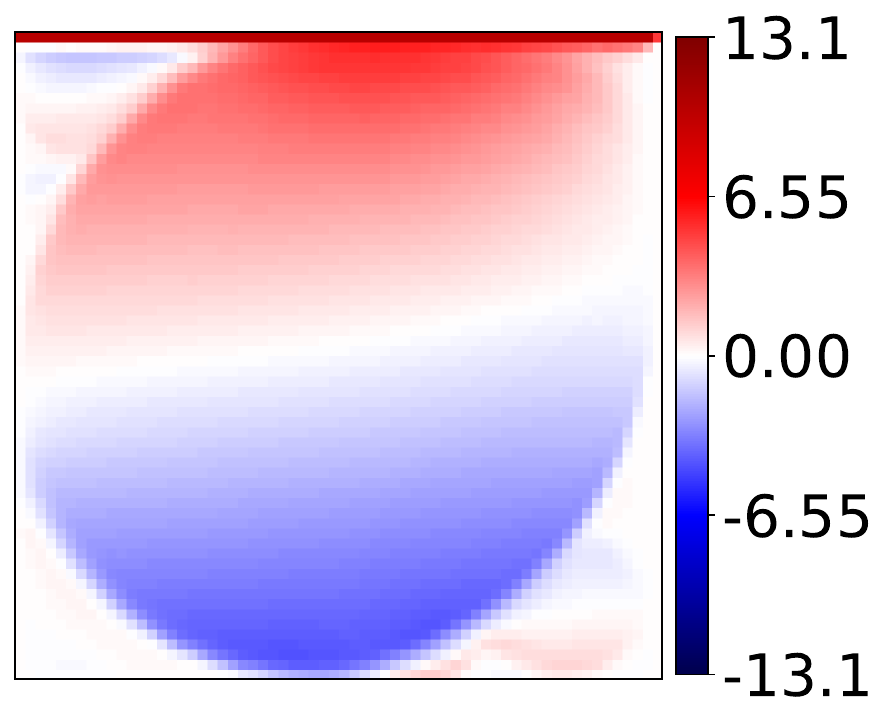}}
\qquad
\subfloat[error u POD]{\includegraphics[width = 1in]{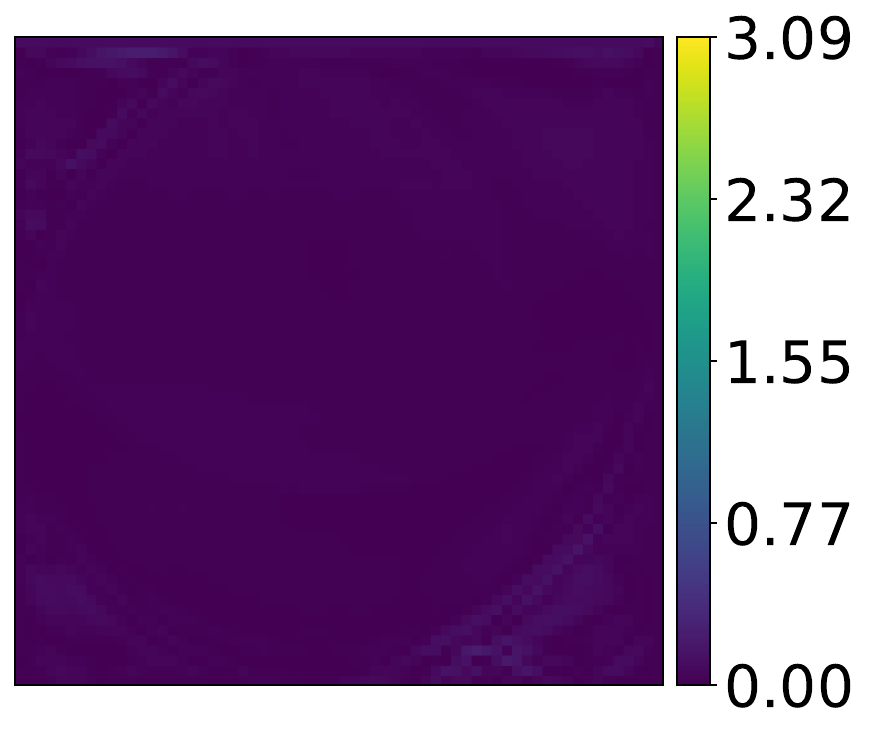}}
\qquad
\subfloat[POD reconstruction]{\includegraphics[width = 1in]{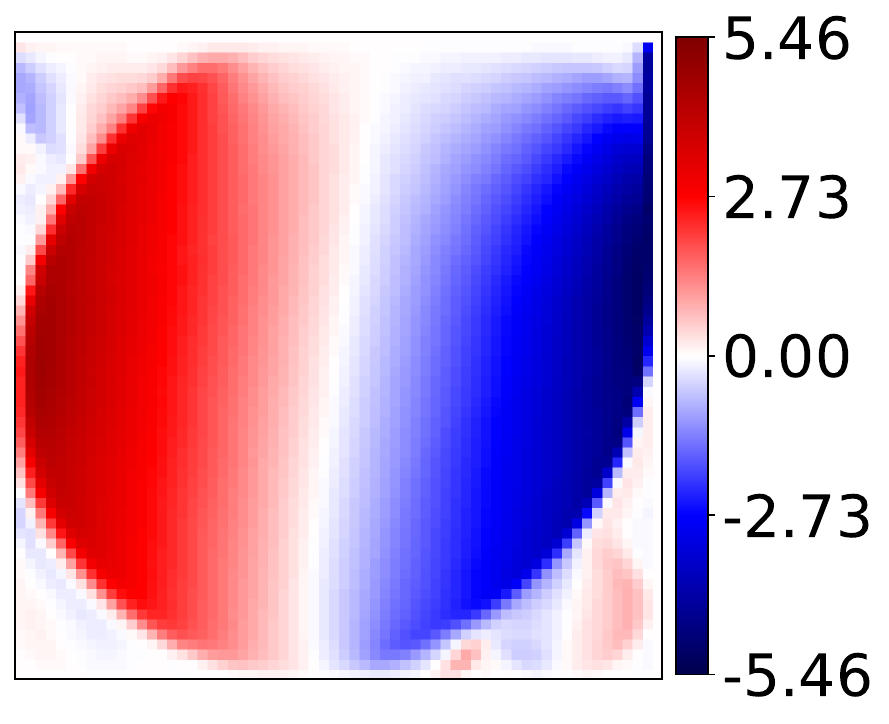}}
\qquad
\subfloat[error v POD]{\includegraphics[width = 1in]{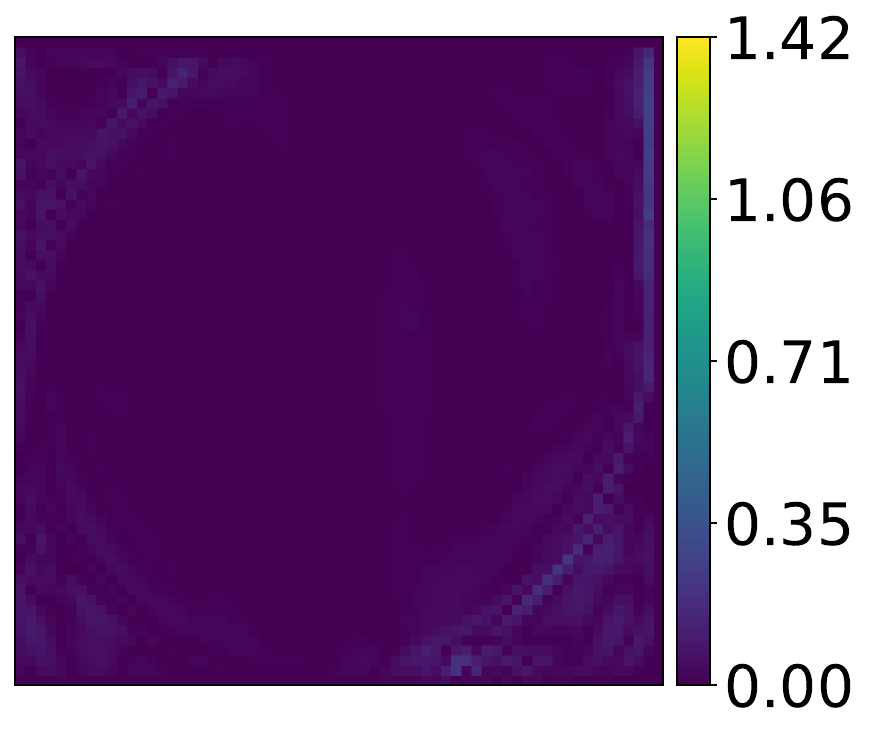}}\\
\subfloat[CAE-MLP reconstruction]{\includegraphics[width = 1in]{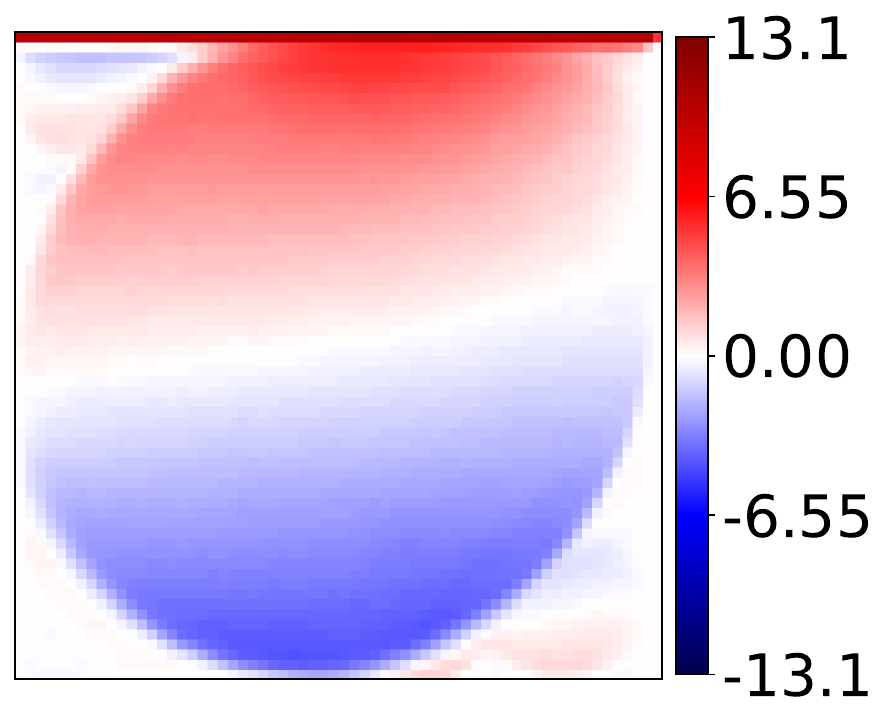}}
\qquad
\subfloat[error u CAE-MLP]{\includegraphics[width = 1in]{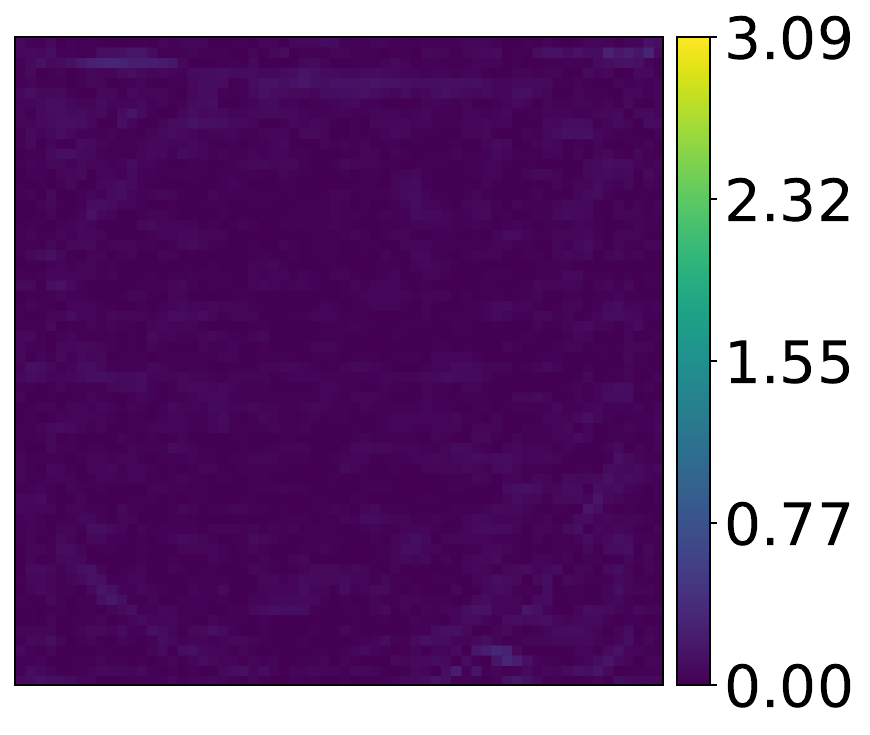}}
\qquad
\subfloat[CAE-MLP reconstruction]{\includegraphics[width = 1in]{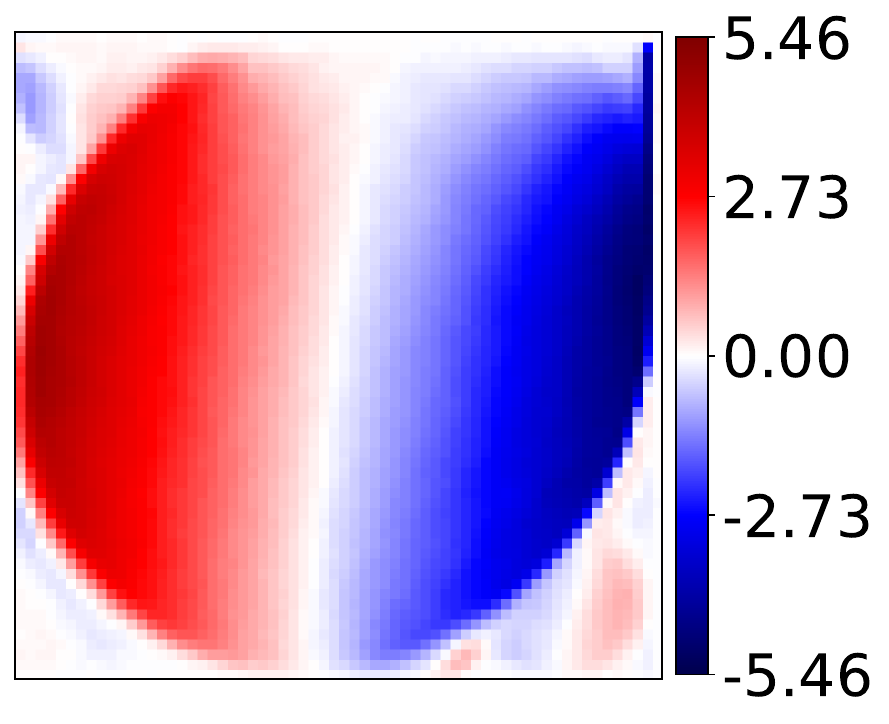}}
\qquad
\subfloat[error v CAE-MLP]{\includegraphics[width = 1in]{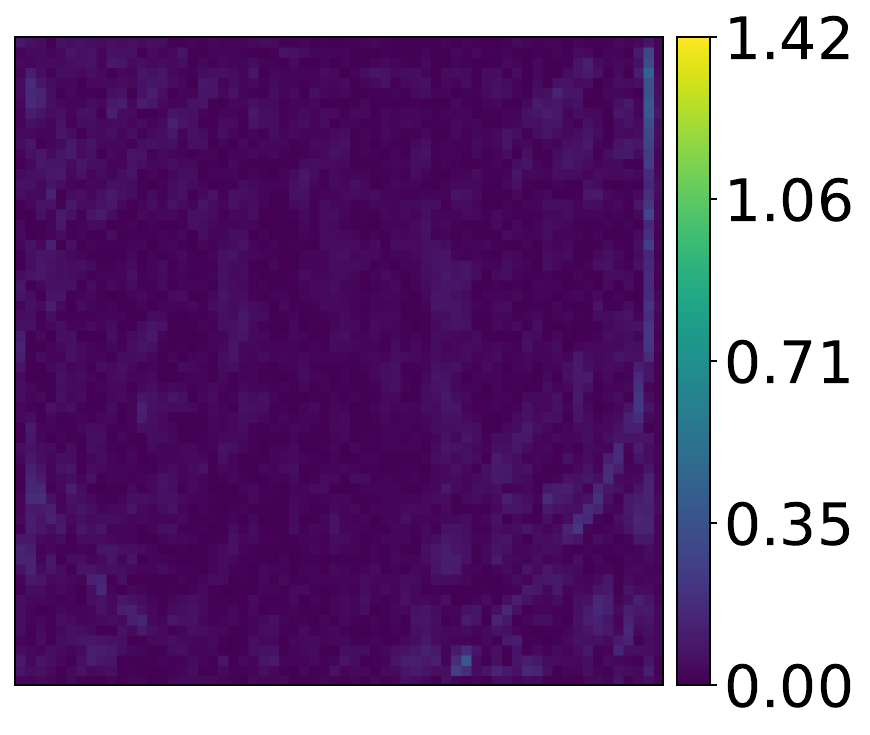}}\\
\subfloat[OACAE-MLP reconstruction]{\includegraphics[width = 1in]{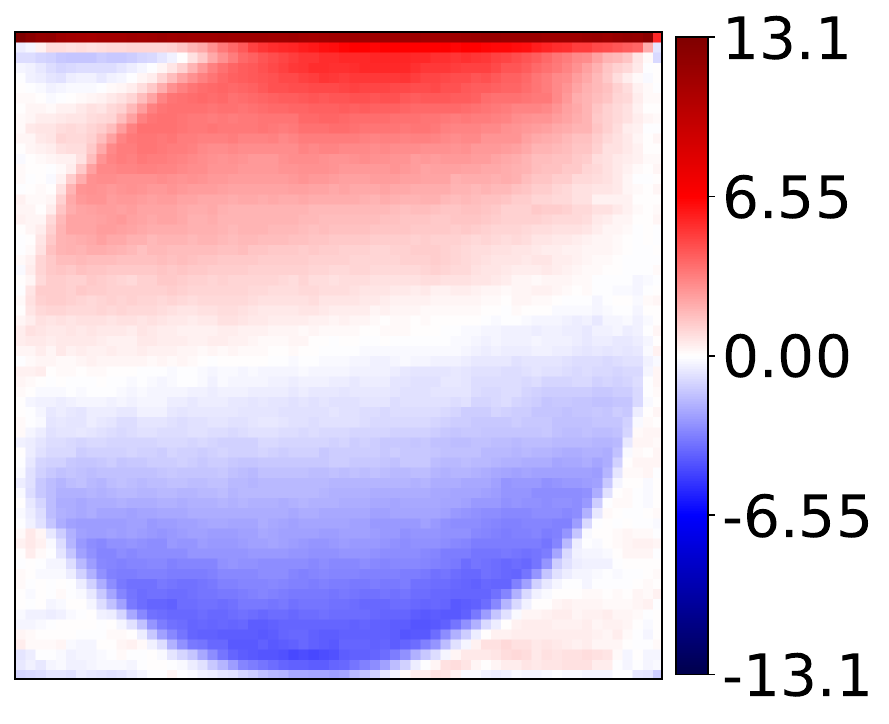}}
\qquad
\subfloat[error u OACAE-MLP]{\includegraphics[width = 1in]{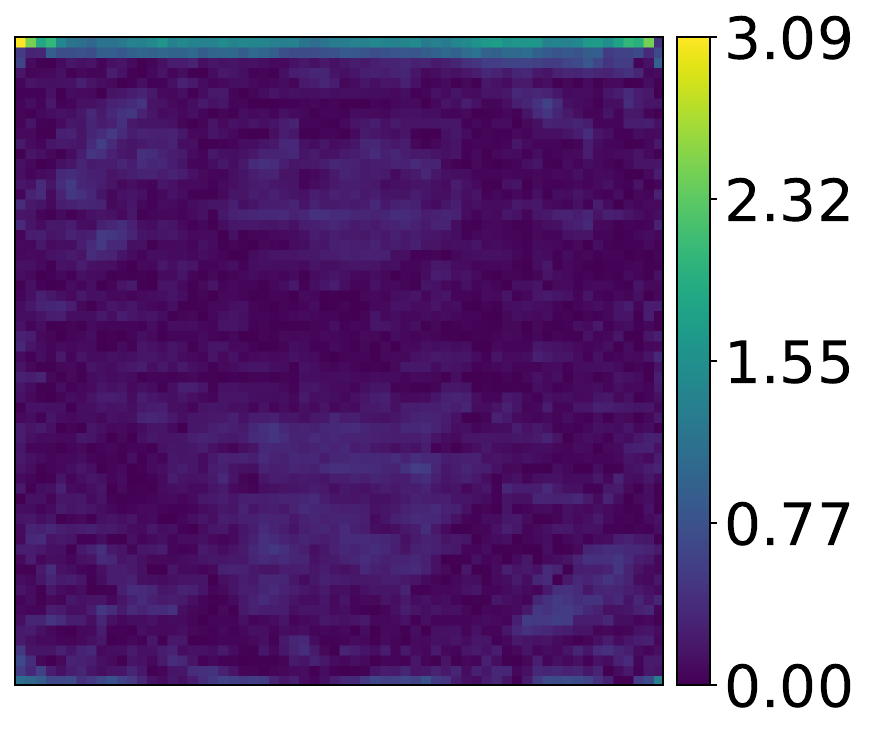}}
\qquad
\subfloat[OACAE-MLP reconstruction]{\includegraphics[width = 1in]{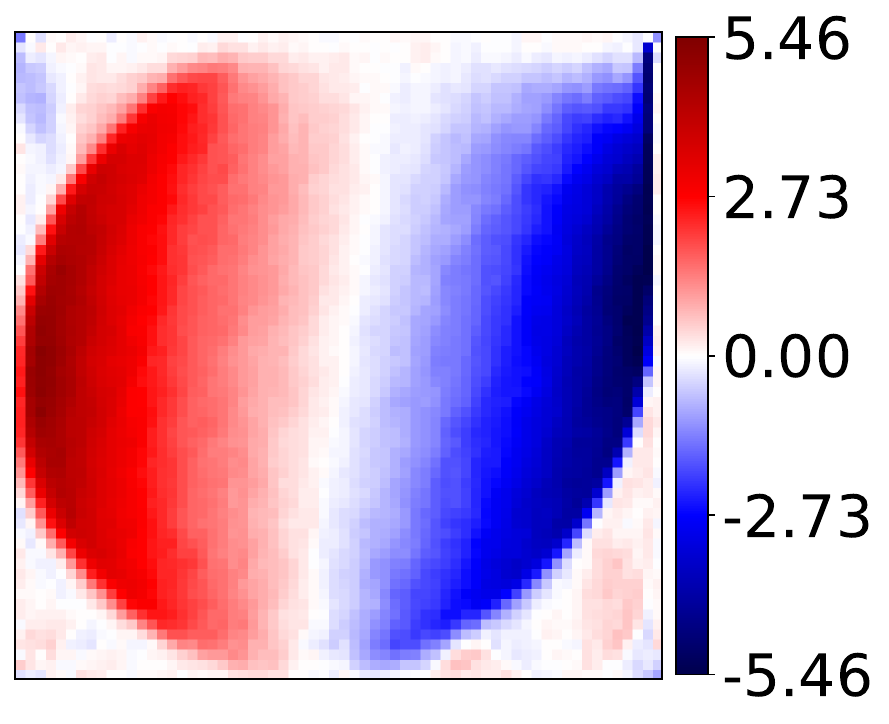}}
\qquad
\subfloat[error v OACAE-MLP]{\includegraphics[width = 1in]{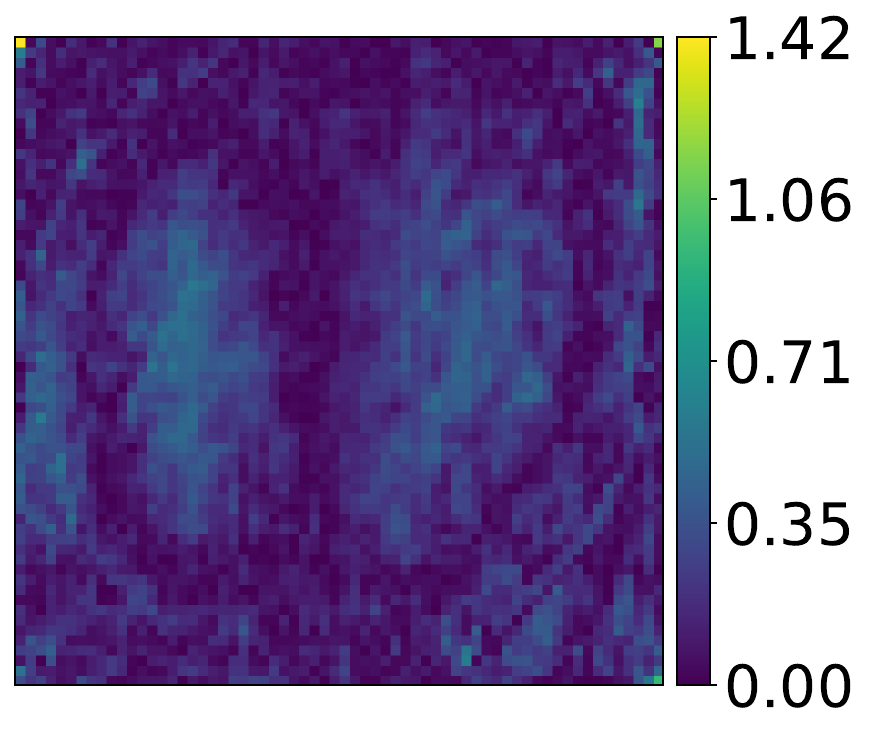}}\\
   \caption{Dam and Cavity reconstruction}
   \label{fig:reconstruction-dam}
\end{figure}

\section{Quantitative analysis of the physics-aware latent space}
\label{app:latent_analysis}

The reconstruction and prediction errors reported in Section \ref{sec:reconstruction} and Section \ref{sec:prediction} evaluate the accuracy of the reduced-order surrogates in the physical space. However, these metrics do not directly indicate whether the learned latent variables are physically informative. In this part, we further analyze the latent representations learned by standard CAE and OACAE.

For each parameter case, the corresponding latent codes form a temporal trajectory in the latent space. A latent representation that is favorable for inverse modeling should satisfy two desirable properties. First, latent trajectories associated with different parameter configurations should be distinguishable, so that different physical regimes are not strongly overlapped. Second, for a fixed parameter configuration, the latent trajectory should evolve smoothly in time, so that the multi-time-step surrogate mapping remains consistent for variational assimilation.

Let \(z_{i,t}\in\mathbb{R}^{d}\) denote the latent code of the \(t\)-th snapshot in the \(i\)-th parameter case, where \(i=1,\ldots,N\), \(t=1,\ldots,T\), and \(d\) is the latent dimension. For each parameter case, we define the latent centroid and the within-case variance as
\[
\mu_i = \frac{1}{T}\sum_{t=1}^{T} z_{i,t},
\qquad
\sigma_i^2 = \frac{1}{T}\sum_{t=1}^{T} \|z_{i,t}-\mu_i\|_2^2.
\]
The average within-case variance is then given by
\[
\sigma_{\mathrm{within}}^2
=
\frac{1}{N}\sum_{i=1}^{N}\sigma_i^2.
\]
This quantity measures the average dispersion of each temporal latent trajectory around its case centroid. To quantify the separation between different parameter cases, we compute the pairwise distance between case centroids,
\[
d_{ij} = \|\mu_i-\mu_j\|_2,\qquad i\neq j,
\]
and define the average between-case distance as
\[
d_{\mathrm{between}}
=
\frac{2}{N(N-1)}
\sum_{1\leq i<j\leq N} d_{ij}.
\]
Since the absolute scale of the latent space may differ between CAE and OACAE, we use the normalized case-separability ratio
\[
R_{\mathrm{sep}}
=
\frac{d_{\mathrm{between}}}{\sqrt{\sigma_{\mathrm{within}}^2}}.
\]
A larger value of \(R_{\mathrm{sep}}\) indicates that different parameter cases are more clearly separated relative to the temporal dispersion within each case.

In addition to case-level separability, we quantify the temporal regularity of the latent trajectories. Because the latent coordinates may have different scales, each latent dimension is first standardized over the whole dataset. Denoting the standardized latent code by \(\tilde{z}_{i,t}\), we define the first-order temporal smoothness indicator
\[
S_i^{(1)}
=
\frac{1}{T-1}
\sum_{t=1}^{T-1}
\|\tilde{z}_{i,t+1}-\tilde{z}_{i,t}\|_2,
\]
and the second-order temporal smoothness indicator
\[
S_i^{(2)}
=
\frac{1}{T-2}
\sum_{t=2}^{T-1}
\|\tilde{z}_{i,t+1}-2\tilde{z}_{i,t}+\tilde{z}_{i,t-1}\|_2.
\]
The global smoothness indicators are obtained by averaging over all parameter cases:
\[
S^{(1)} = \frac{1}{N}\sum_{i=1}^{N}S_i^{(1)},
\qquad
S^{(2)} = \frac{1}{N}\sum_{i=1}^{N}S_i^{(2)}.
\]
Smaller values of \(S^{(1)}\) and \(S^{(2)}\) indicate smoother temporal evolution in the latent space.

Table~\ref{tab:latent_train} reports the latent-space metrics on the training set. Compared with CAE, OACAE yields a smaller within-case variance and a substantially larger normalized separability ratio. In particular, \(R_{\mathrm{sep}}\) increases from \(1.501653\) for CAE to \(3.474383\) for OACAE. The first- and second-order temporal smoothness indicators are also reduced, indicating that the OACAE latent trajectories are more regular in time.

\begin{table}[htbp]
\centering
\caption{Comparison of latent representations between CAE and OACAE on the training dataset.}
\label{tab:latent_train}
\begin{tabular}{lccccc}
\hline
Model & \(S^{(1)}\downarrow\) & \(S^{(2)}\downarrow\) & \(\sigma_{\mathrm{within}}^2\) & \(d_{\mathrm{between}}\) & \(R_{\mathrm{sep}}\uparrow\) \\
\hline
CAE   & \(0.322924 \pm 0.073712\) & \(0.199233 \pm 0.066340\) & 6.648507 & 3.871967 & 1.501653 \\
OACAE & \(0.145713 \pm 0.033335\) & \(0.104804 \pm 0.051789\) & 0.703090 & 2.913287 & 3.474383 \\
\hline
\end{tabular}
\end{table}

\begin{figure}[H]
    \centering
    \begin{subfigure}[t]{0.48\textwidth}
        \centering
        \includegraphics[width=\textwidth]{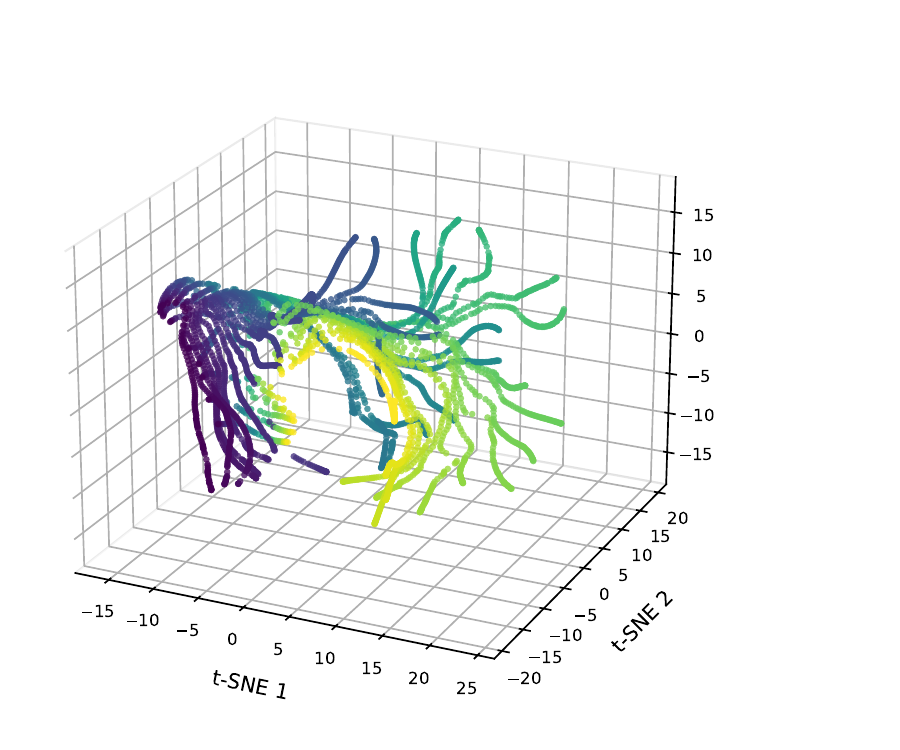}
        \caption{CAE}
    \end{subfigure}
    \hspace{0.0005\textwidth}
    \begin{subfigure}[t]{0.48\textwidth}
        \centering
        \includegraphics[width=\textwidth]{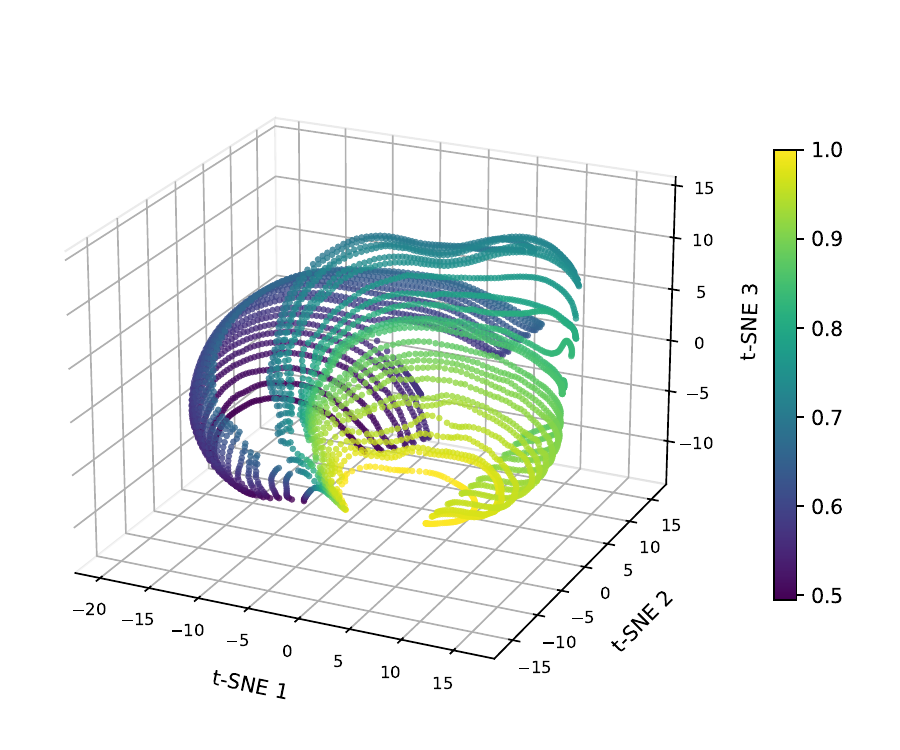}
        \caption{OACAE}
    \end{subfigure}
    \caption{Visualization of latent manifolds of dam flow evolutions for different boundary conditions (inlet velocity).}
    \label{fig:latent_tsne}
\end{figure}

Figure~\ref{fig:latent_tsne} provides a complementary visualization of the latent manifolds using t-SNE projections. The CAE latent trajectories exhibit stronger overlap and less organized case-level structure, whereas the OACAE latent trajectories are more structured according to parameter-induced dynamical regimes. These visual and quantitative results support the interpretation that observable supervision improves the parameter consistency of the learned latent representation. 

\section{Sensitivity analyses}  
\label{app: Sensitivity analyses}

\subsection{Sensitivity to training loss weights}
\label{app:loss_weight_sensitivity}

We first examine the sensitivity of the proposed AE-MLP variational DA framework to the training
loss weights \(\alpha\) and \(\beta\). 
We fix \(\beta=1\) and vary
\[
\alpha \in \{0,0.01,0.05,0.1,0.2,0.5\},
\]
and fix \(\alpha=0.05\) and vary
\[
\beta \in \{0.5,1,2\}.
\]
For each setting, we evaluate the autoencoder reconstruction error, the AE-MLP prediction
error, and the full-observation 3D-Var inverse-calibration error.

\begin{figure}[htbp]
    \centering
    \includegraphics[width=0.96\textwidth]{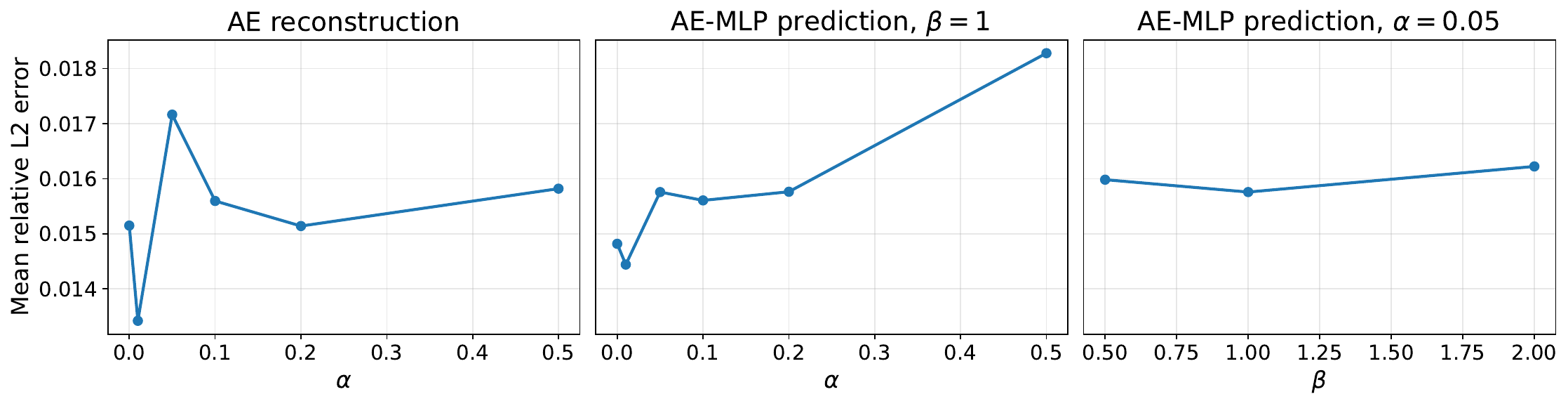}
    \caption{Sensitivity analysis of $\alpha$ and $\beta$ for forward modeling.}
    \label{fig:sensitivity_alpha_beta_forward}
\end{figure}

\begin{figure}[htbp]
    \centering
    \includegraphics[width=0.95\textwidth]{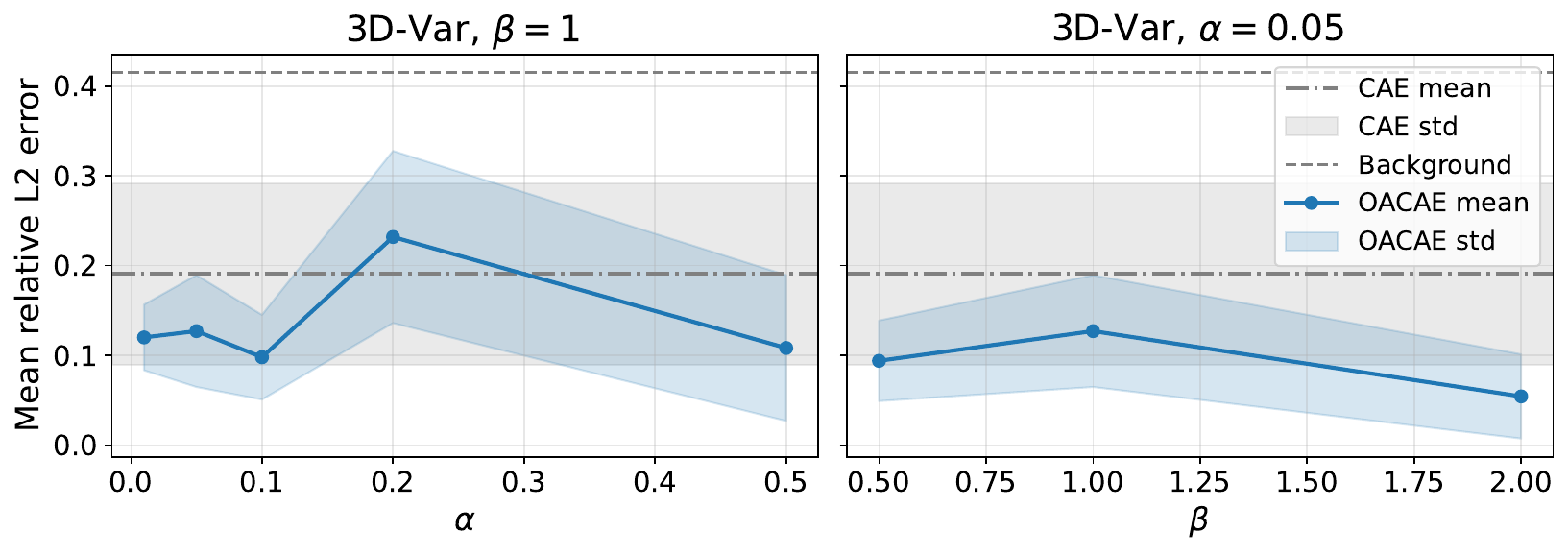}
    \caption{Sensitivity analysis of \(\alpha\) and \(\beta\) for inverse modeling.}
    \label{fig:app_inverse_weight_sensitivity}
\end{figure}

\subsection{Sensitivity to latent dimension}
\label{app:latent_dimension_sensitivity}

We further study the sensitivity of the proposed framework to the latent-space dimension. 
Both CAE-MLP and OACAE-MLP are evaluated using latent dimensions
\[
q \in \{2,6,13,32,64\}.
\]
For all tested latent dimensions, the weighting parameters are fixed at values of \(\alpha=0.05\) and \(\beta=1\). The comparison includes the autoencoder reconstruction error,
the AE-MLP prediction error, and the full-observation 3D-Var inverse-calibration error.

\begin{figure}[htbp]
    \centering
    \includegraphics[width=0.96\textwidth]{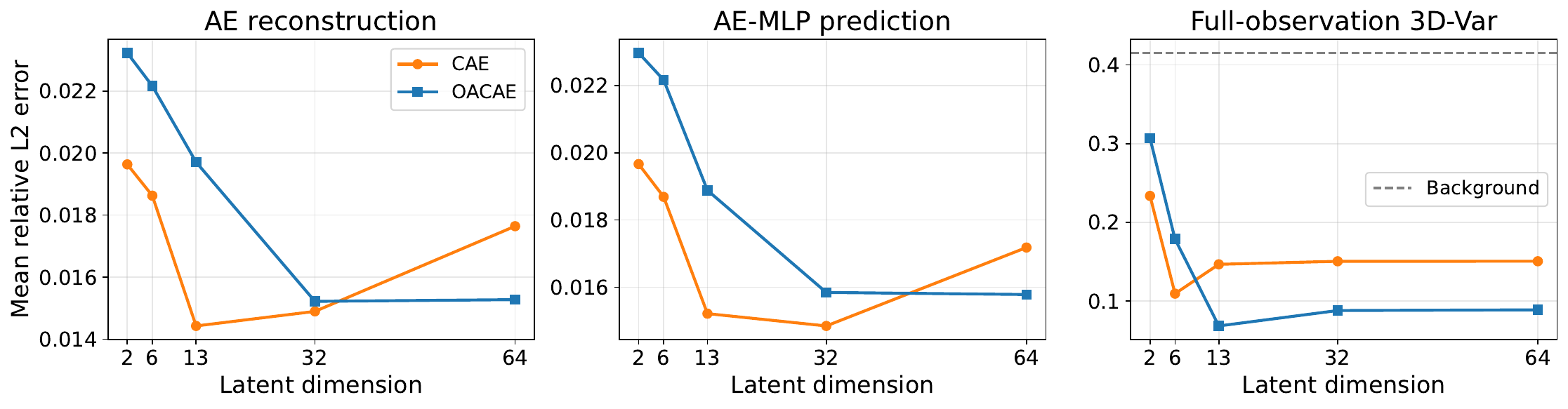}
    \caption{Sensitivity analysis of AE latent dimension.}
    \label{fig:sensitivity_latent_dim}
\end{figure}

As shown in Figure~\ref{fig:sensitivity_latent_dim}, very small latent dimensions,
such as \(q=2\) and \(q=6\), lead to larger errors because the latent space does not have
sufficient capacity to represent the dominant flow structures and the parameter-relevant
information. Increasing the latent dimension improves both reconstruction and prediction
performance, but the improvement saturates around \(q=32\). For the inverse 3D-Var calibration task, all tested latent dimensions substantially reduce the
parameter error compared with the background estimate. The OACAE-MLP framework shows
a clear improvement from \(q=2\) to \(q=13\), and then remains stable for \(q=32\) and
\(q=64\). As a practical heuristic, the POD energy spectrum can be used as a conservative reference
for selecting a preliminary latent dimension, since POD provides the optimal linear low-rank
approximation of the snapshot data in the mean-square sense. In addition, if the solution
manifold is mainly parameterized by \(p_{\mathrm{eff}}\) effective variables, a simple manifold-based
heuristic suggests testing latent dimensions around \(p_{\mathrm{eff}}\) and \(2p_{\mathrm{eff}}+1\)\cite{Franco_2022}. In the
present datasets, the effective number of generating variables is approximately
\(p_{\mathrm{eff}}=6\), which motivates the tested dimensions \(q=6\) and \(q=13\), together with
larger values \(q=32\) and \(q=64\). This argument is used only as a heuristic reference rather
than as a strict optimality criterion. The final choice is based on the empirical sensitivity
results for reconstruction, prediction, and inverse calibration.

\subsection{Sensitivity to variational optimization learning rate}
\label{app:learning_rate}

\begin{figure}[htbp]
    \centering

    \begin{subfigure}[t]{0.48\textwidth}
        \centering
        \includegraphics[width=\linewidth]{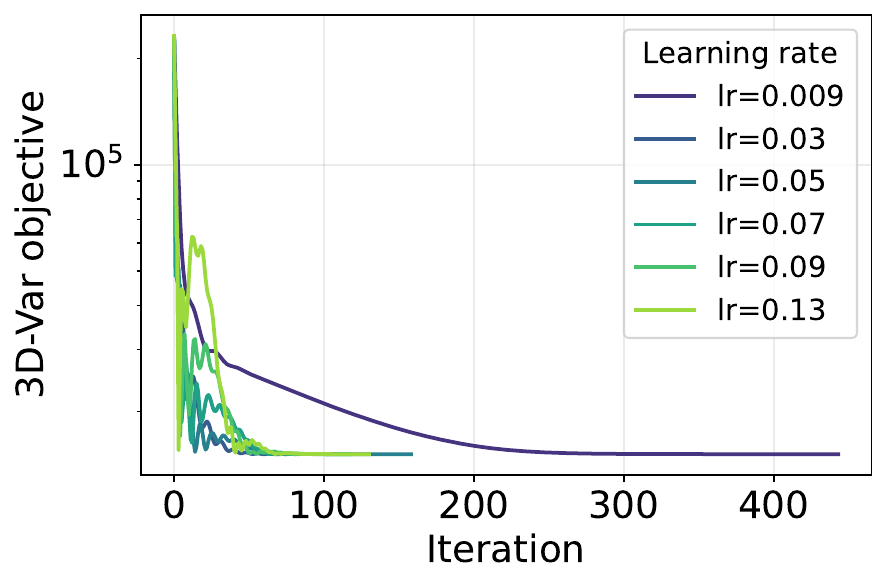}
        \caption{CAE 3dvar history fixing observation varying learning rate}
        \label{fig:cae-single-lr}
    \end{subfigure}
    \hfill
    \begin{subfigure}[t]{0.48\textwidth}
        \centering
        \includegraphics[width=\linewidth]{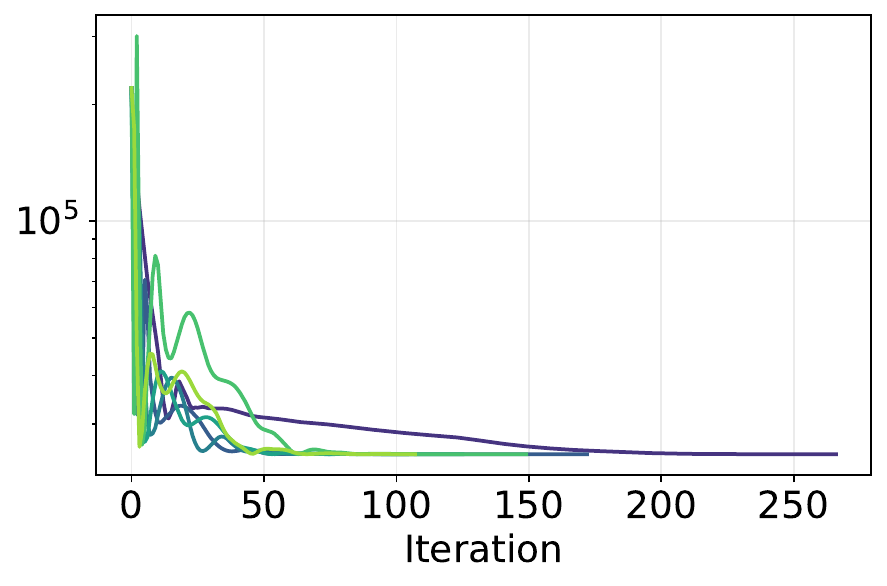}
        \caption{OACAE 3dvar history fixing observation varying learning rate}
        \label{fig:oacae-single-lr}
    \end{subfigure}

    \vspace{0.05em}
    
    \begin{subfigure}[t]{0.48\textwidth}
        \centering
        \includegraphics[width=\linewidth]{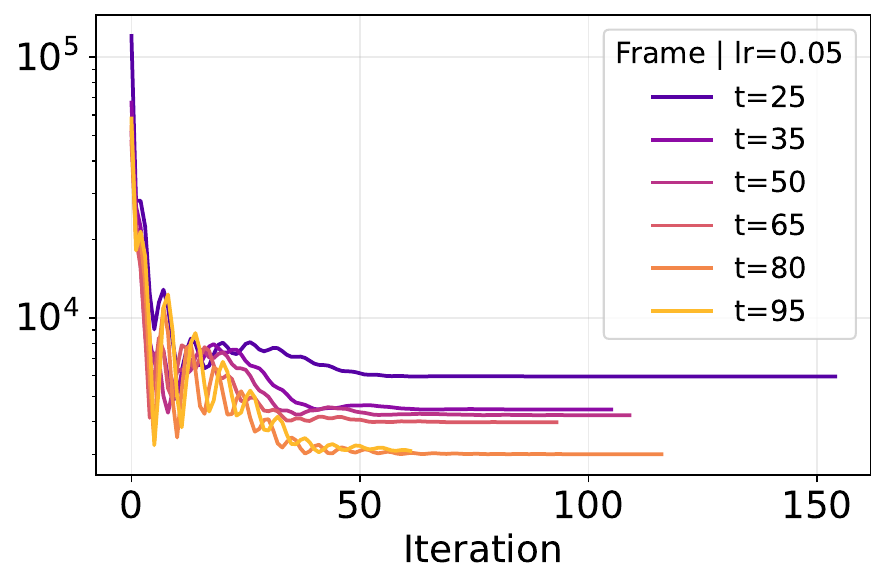}
        \caption{CAE 3dvar history fixing learning rate varying observations}
        \label{fig:cae-fixed-lr}
    \end{subfigure}
    \hfill
    \begin{subfigure}[t]{0.48\textwidth}
        \centering
        \includegraphics[width=\linewidth]{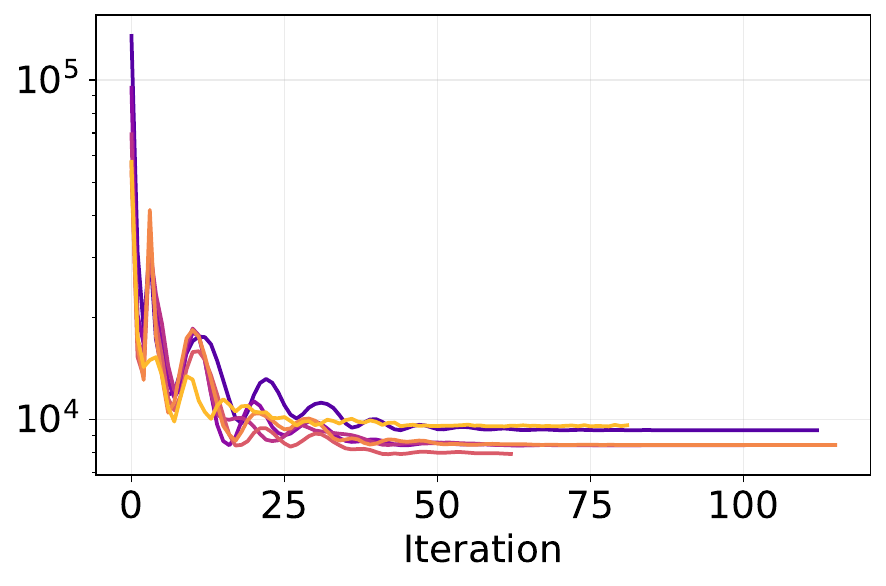}
        \caption{OACAE 3dvar history fixing learning rate varying observations}
        \label{fig:oacae-fixed-lr}
    \end{subfigure}

    \caption{Gradient-based optimization objective histories.}
    \label{Figures/fig:gradient-optimization-histories}
\end{figure}

We also investigate the sensitivity of the 3D-Var optimization to the learning rate used in the gradient-based minimization. 
We performed a
learning-rate sweep over
\[
\eta \in \{0.009,\,0.03,\,0.05,\,0.07,\,0.09,\,0.13\},
\]
for both CAE-MLP and OACAE-MLP under the same assimilation setting. The corresponding objective histories are presented in
Figures~\ref{Figures/fig:gradient-optimization-histories}(a) and
\ref{Figures/fig:gradient-optimization-histories}(b). For both models, all tested learning rates eventually approach similar objective plateaus for the representative observation, indicating that the converged objective is relatively insensitive to the learning rate within the tested range. 
However, the convergence speed and transient behavior depend on the optimizer setting. The smallest learning rate gives smooth but slow convergence, whereas larger learning rates accelerate the initial decrease but may introduce stronger transient oscillations or occasional overshooting. Figures~\ref{Figures/fig:gradient-optimization-histories}(c) and
\ref{Figures/fig:gradient-optimization-histories}(d) further examine convergence at the selected learning rate $\eta=0.05$ for observations taken at different time instances. Although the initial objective values and final plateaus vary with 
the observation, the OACAE-MLP trajectories are more tightly clustered, while the CAE-MLP trajectories show a wider spread in final objective values. This suggests more uniform 
convergence behavior for OACAE-MLP across different observations.

\begin{figure}[htbp]
    \centering
    \begin{subfigure}[t]{0.48\textwidth}
        \centering
        \includegraphics[width=\textwidth]{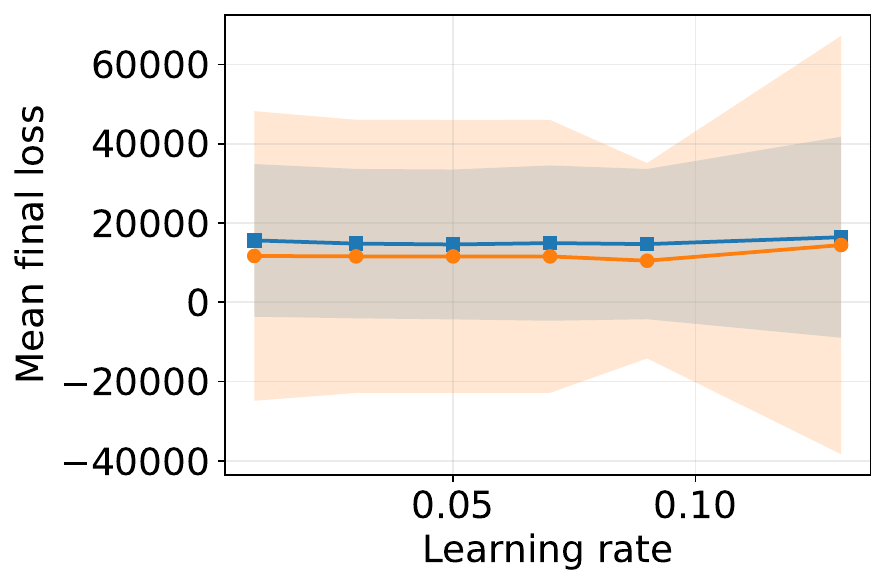}
        \caption{Mean final objective values varying learning rate}
    \end{subfigure}
    \hspace{0.0005\textwidth}
    \begin{subfigure}[t]{0.48\textwidth}
        \centering
        \includegraphics[width=\textwidth]{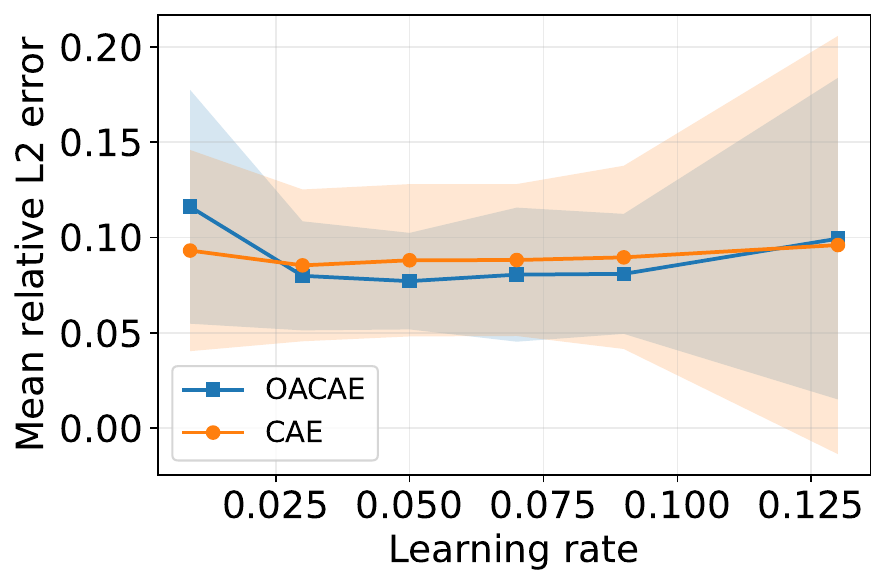}
        \caption{Parameter calibration errors varying learning rate}
    \end{subfigure}
    \caption{Sensitivity of the 3D-Var optimization results to the learning rate.}
    \label{Figures/fig: gradient based optimization}
\end{figure}

The statistical calibration results are summarized in Figure~\ref{Figures/fig: gradient based optimization}. The mean final objective remains nearly constant for \(0.03\leq \eta_{\mathrm{opt}}\leq 0.09\). Although CAE-MLP 
can yield a lower mean final objective value, this should not be interpreted as better calibration, because CAE-MLP and OACAE-MLP define different nonlinear mappings and therefore different optimization landscapes. The parameter-recovery results show that OACAE-MLP achieves lower mean relative errors for the calibrated parameters over the tested learning-rate range. It also exhibits smaller standard deviations for most intermediate learning rates. Overall, the interval
\[
0.03 \leq \eta_{\mathrm{opt}} \leq 0.09
\]
provides the best compromise between convergence speed, parameter accuracy, and robustness.

\subsection{Sensitivity to background and observation covariance scaling}
\label{app:covariance_scaling}

The variational data-assimilation objective depends on the background covariance matrix \(\mathbf{B}\) and the observation covariance matrix \(\mathbf{R}\). These matrices control the relative weights of the background term and the observation term. To assess the sensitivity of the calibration results to this balance, we perform additional experiments by scaling \(\mathbf{B}\) and \(\mathbf{R}\). Let \(\mathbf{B}_0\) and \(\mathbf{R}_0\) denote the reference covariance matrices used in the main experiments. We consider scaled covariance matrices of the form
\[
\mathbf{B} = \lambda_\mathbf{B} \mathbf{B}_0,
\qquad
\mathbf{R} = \lambda_\mathbf{R} \mathbf{R}_0,
\]
where \(\lambda_\mathbf{B}\) and \(\lambda_\mathbf{R}\) were varied over the range
\[
\lambda_\mathbf{B}, \lambda_\mathbf{R} \in
\left\{10^{-4}, 10^{-3}, 10^{-2}, 10^{-1}, 1, 10, 10^2, 10^3, 10^4\right\}.
\] 

\begin{figure}[htbp]
    \centering
    \includegraphics[width=0.85\textwidth]{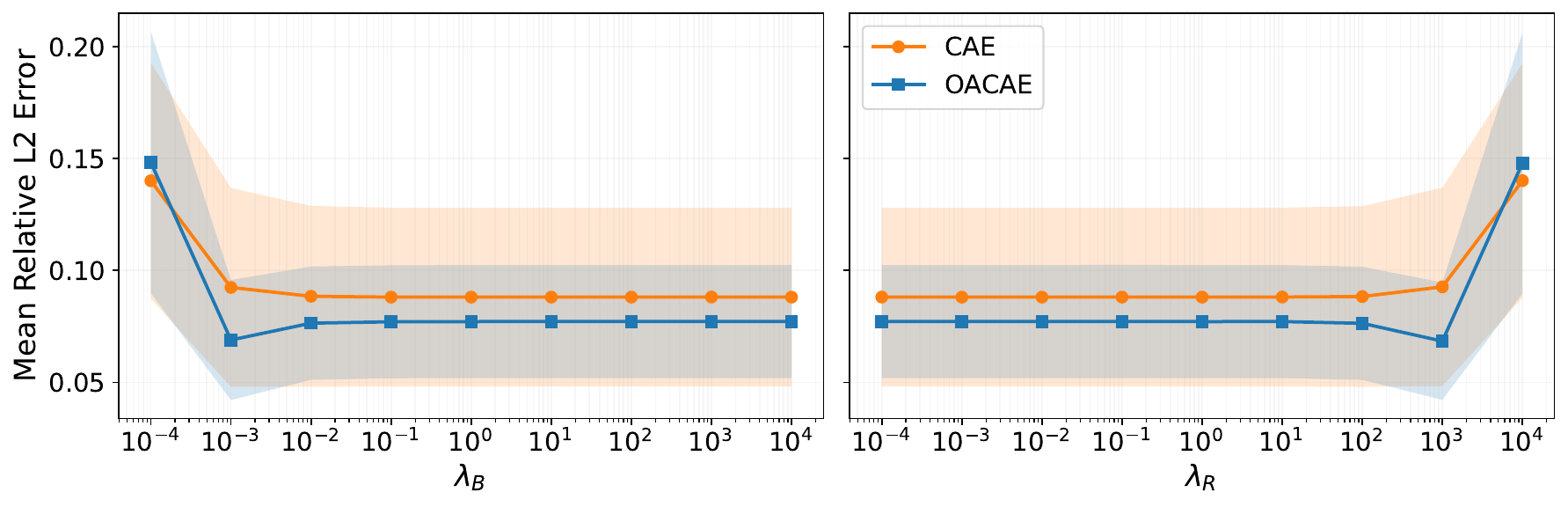}
    \caption{Sensitivity of the assimilation results to the scaling of the background covariance matrix $\mathbf{B}$ and the observation covariance matrix \(\mathbf{R}\). The left panel varies \(\mathbf{B} = \lambda_\mathbf{B} \mathbf{B}_0\) while fixing \(\mathbf{R} = \mathbf{R}_0\), and the right panel varies \(\mathbf{R} = \lambda_\mathbf{R} \mathbf{R}_0\) while fixing \(\mathbf{B} = \mathbf{B}_0\).}
    \label{fig:sensitivity_l2_BR}
\end{figure}

Figure~\ref{fig:sensitivity_l2_BR} reports the calibration performance under different covariance-scaling settings. When $\mathbf{B}$ is scaled to be very small, the background term becomes overly dominant, forcing the analysis to remain too close to the background state and leading to larger parameter errors (Figure~\ref{fig:sensitivity_l2_BR} left). Conversely, when $\mathbf{R}$ is scaled to be very large, the observation term is strongly downweighted, weakening the observational correction and also increasing the parameter error(Figure~\ref{fig:sensitivity_l2_BR} right). These results indicate that the proposed framework is stable with respect to reasonable choices of B and R and the conclusions of the main experiments are not tied to a single covariance choice.

\section{Computational efficiency}  
\label{app: Computational efficiency}

Finally, we compare the online computational efficiency of the proposed differentiable variational assimilation framework with an ensemble-based POD-GPR assimilation baseline. The purpose of this experiment is to evaluate the cost of solving the inverse problem after the offline surrogate training stage has been completed. Therefore, the reported wall-clock time corresponds to the online calibration stage only. 

\begin{figure}[H]
\centering
\includegraphics[width=0.80\textwidth]{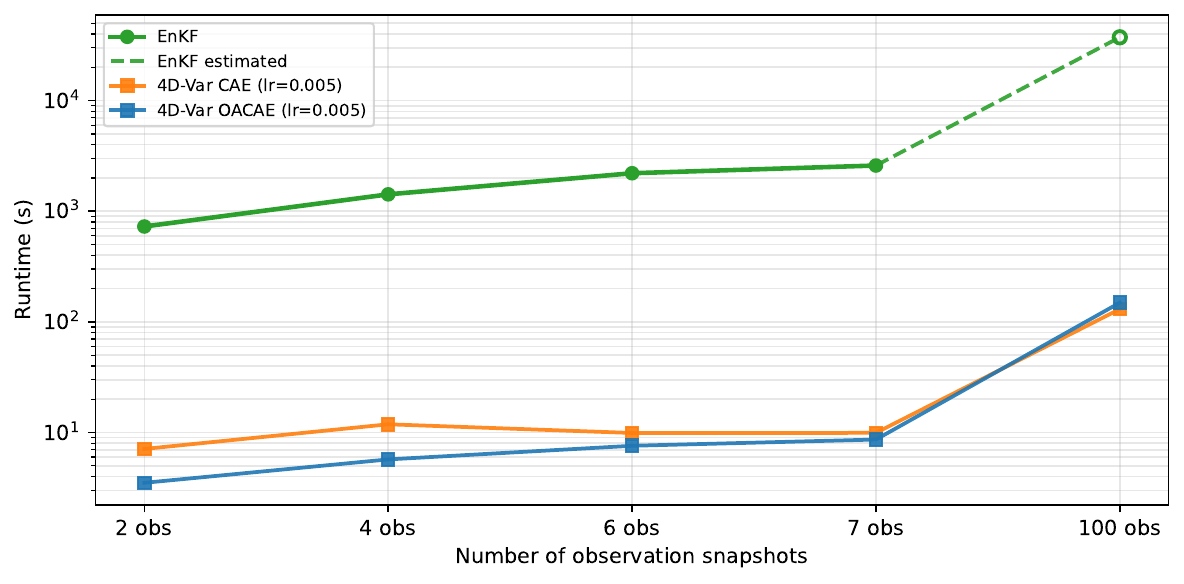}
\caption{Online runtime comparison between the EnKF-POD-GPR baseline and the differentiable 4D-Var frameworks using CAE-MLP and OACAE-MLP surrogates. The dashed EnKF curve denotes the estimated runtime for the 100-snapshot case, and the 4D-Var results are obtained with learning rate \(0.005\).}
\label{fig:runtime_comparison}
\label{fig:app_runtime}
\end{figure}

The POD-GPR ensemble-based strategy requires repeated surrogate evaluations for all ensemble members and for each assimilated time snapshot. In contrast, the AE-MLP variational framework solves the inverse problem by directly optimizing the physical control parameters through automatic differentiation and the optimization can be accelerated on a GPU. Figure~\ref{fig:app_runtime} compares the wall-clock time for different numbers of assimilated time snapshots. The results show that the proposed AE-MLP variational framework is substantially more efficient than the ensemble-based POD-GPR baseline. This advantage becomes more pronounced as the number of assimilated snapshots increases.

\end{document}